\pdfoutput=1
\documentclass[10pt,twocolumn,letterpaper]{article}

\usepackage{iccv}
\usepackage{times}
\usepackage{epsfig}
\usepackage{graphicx}
\usepackage{amsmath}
\usepackage{amssymb}
\usepackage[font=small]{caption}

\makeatletter
\g@addto@macro\normalsize{%
  \setlength\abovedisplayskip{5pt}
  \setlength\belowdisplayskip{6pt}
}
\makeatother

\newcommand{\csection}[1]{
    \section{#1}
}

\newcommand{\csubsection}[1]{
    \vspace{-0.1cm}
    \subsection{#1}
    \vspace{-0.1cm}
}

\usepackage{amsmath}
\usepackage{amssymb}    %
\usepackage{amsthm}
\usepackage{mathrsfs}   %
\usepackage{bm}
\usepackage{url}

\usepackage[dvipsnames]{xcolor}

\usepackage{enumitem}
\usepackage{subcaption}
\usepackage{comment}
\usepackage{slashed}
\usepackage{booktabs}  %
\usepackage{float}
\usepackage{microtype}

\usepackage{mathtools}  %

\usepackage[normalem]{ulem}  %

\usepackage{tabularx}
\newcolumntype{C}{>{\centering\arraybackslash}X}
\newcolumntype{R}{>{\raggedleft\arraybackslash}X}
\newcolumntype{L}{>{\raggedright\arraybackslash}X}

\usepackage{adjustbox}

\usepackage{changepage}

\usepackage{makecell}

\usepackage{advdate}

\usepackage{diagbox}

\usepackage{booktabs}%

\usepackage{multirow}

\usepackage{bbm}

\usepackage{setspace}  %

\usepackage{pifont}
\newcommand{\cmark}{\ding{51}}

\usepackage{inconsolata}

\newcommand{\fwd}[0]{\texttt{move\_forward}\xspace}
\newcommand{\turnr}[0]{\texttt{turn\_right}\xspace}
\newcommand{\turnl}[0]{\texttt{turn\_left}\xspace}
\newcommand{\stp}[0]{\texttt{stop}\xspace}

\newcommand{\figref}[1]{Fig\onedot~\ref{#1}}
\newcommand{\equref}[1]{Eq\onedot~\eqref{#1}}
\newcommand{\secref}[1]{Sec\onedot~\ref{#1}}
\newcommand{\tabref}[1]{Tab\onedot~\ref{#1}}

\usepackage{color, colortbl}
\definecolor{Gray}{gray}{0.88}

\newcommand{\beginsupplement}{
    \setcounter{table}{0}
    \renewcommand{\thetable}{S\arabic{table}}%
    \setcounter{figure}{0}
    \renewcommand{\thefigure}{S\arabic{figure}}%
    \setcounter{equation}{0}
    \renewcommand{\theequation}{S\arabic{equation}}
}

\usepackage[resetlabels]{multibib}
\newcites{supp}{References}

\usepackage[pagebackref=true,breaklinks=true,letterpaper=true,colorlinks,bookmarks=false]{hyperref}

\iccvfinalcopy %

\def\iccvPaperID{7366} %

\ificcvfinal\fi

\begin{document}

\title{The Surprising Effectiveness of Visual Odometry Techniques \\for Embodied PointGoal Navigation}

\author{
    Xiaoming Zhao$^\dagger$, Harsh Agrawal$^\ddagger$, Dhruv Batra$^{\ddagger,\mathsection}$, Alexander Schwing$^\dagger$ \and
    {\small $\prescript{\dagger}{}{}$University of Illinois, Urbana-Champaign}\and
    {\small$\prescript{\ddagger}{}{}$Georgia Institute of Technology}\and
    {\small$\prescript{\mathsection}{}{}$Facebook AI Research}\and
    {\small\url{https://xiaoming-zhao.github.io/projects/pointnav-vo/}}
}

\maketitle
\ificcvfinal\thispagestyle{empty}\fi

\begin{abstract}
   It is fundamental for personal robots to reliably navigate to a specified goal. To study this task, PointGoal navigation has been introduced in simulated Embodied AI environments. Recent advances solve this PointGoal navigation task with near-perfect accuracy (99.6\% success) in photo-realistically simulated environments, assuming noiseless egocentric vision, noiseless actuation and most importantly, \textit{perfect localization}.
   However, under realistic noise models for visual sensors and actuation, and without access to a ``GPS and Compass sensor,'' the  99.6\%-success agents for PointGoal navigation  only succeed with 0.3\%.\footnote{\url{https://eval.ai/web/challenges/challenge-page/580/leaderboard/1631} (Habitat Team).}
   In this work, we demonstrate the surprising effectiveness of visual odometry for the task of PointGoal navigation in this realistic setting, \ie,  with realistic noise models for perception and actuation and without access to GPS and Compass sensors.
   We show that integrating visual odometry techniques into
   navigation policies %
   improves the state-of-the-art  on the popular Habitat PointNav benchmark by a large margin, improving success from 64.5\% to 71.7\% while executing 6.4 times faster.
\end{abstract}

\csection{Introduction}

The ability to navigate efficiently and accurately within an indoor environment is fundamental to  personal robots and has been a focus of research in computer vision for many years~\cite{Nilsson1984ShakeyTR}.
To coalesce the community around a common framework and standard metrics, Anderson \etal~\cite{Anderson2018OnEO} proposed the task of PointGoal navigation.
In PointGoal navigation, an agent is randomly spawned in a previously unseen environment and has to navigate to a point goal specified relative to the agent's initial location and orientation,~\eg,~`Go 5m north, 3m west relative to start'.
The agent uses a discrete action space (\eg,~\fwd 0.25$m$, \turnl or \turnr $30^{\circ}$, and \stp) to navigate in the environment.
Under the assumption of noiseless egocentric vision (noise-free RGB + depth sensors), noise-free actuation (\eg,~\turnl will always turn exactly $30^{\circ}$) and perfect localization using \texttt{GPS+Compass} sensors, recent methods solve this task with near-perfect accuracy (99.6\% success)~\cite{Wijmans2020DDPPOLN}. 

\begin{figure}[t]
\vspace{-0.5cm}
\centering
\includegraphics[width=\linewidth]{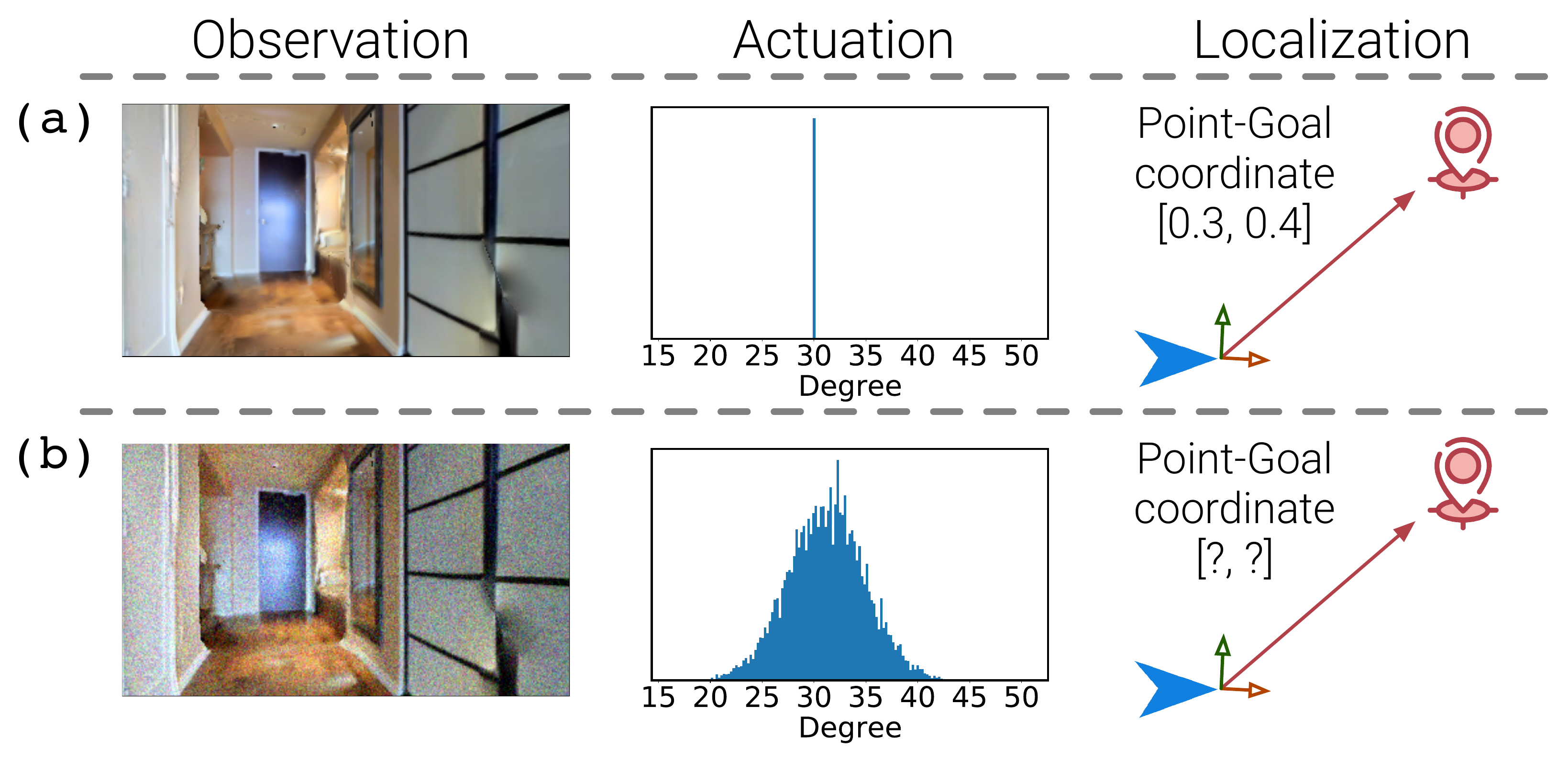}
\vspace{-0.9cm}
\caption{Noiseless (a) and noisy (b) PointGoal navigation. In the noisy setting, the agent observes: 1) sensor noises in egocentric observation; 2) actuation perturbations. The second column shows a histogram of orientation angle changes caused by a \texttt{turn\_left} action; 3) no localization information. The agent's inaccurate localization results in uncertainty about the goal location.}
\label{fig: teaser}
\vspace{-0.5cm}
\end{figure}

However, these assumptions are unrealistic. 
Note that GPS sensors typically don't yield a precise location in indoor environments. 
In addition, perception and actuation of real robots often depend heavily on
environment lighting and %
friction coefficients of surfaces. 
To study this more realistic setting, in a recent  benchmark\footnote{\url{https://aihabitat.org/challenge/2020/}}, PointGoal navigation was updated to include noisy actuation models from real robots~\cite{pyrobot2019}. 
For example, for a single \turnl action, the actual turn angle  varies significantly as shown in  column two of
\figref{fig: teaser}. %
Also, RGB   and depth noise models from~\cite{choi2015robust} were incorporated to simulate a real-world camera. 
Most importantly, as illustrated in  column three of \figref{fig: teaser}, the agent \textit{does not} have access to \texttt{GPS+Compass} data and must navigate solely based on egocentric RGB + depth (RGB-D) measurements.
Under such a more realistic setting, the performance of a policy that is near-perfect in noiseless scenarios~\cite{Wijmans2020DDPPOLN} drops drastically to 0.3\%.
Improving upon it, prior  state-of-the-art~\cite{Karkus2021cvpr} incorporates particle SLAM  into visual navigation and achieves a success rate of 64.5\% under such a realistic setting. Compared to the 99.6\% success rate on the noiseless version of the task, navigation with noisy perception and actuation as well as without localization information hence remains challenging. 
    
To better understand the challenges of navigation in this realistic setting, we study three visual odometry (VO) techniques. 
We find those VO techniques to be surprisingly effective for PointGoal navigation  in this realistic setting.
Specifically, we
1) leverage the geometric invariances of visual odometry;  
2) incorporate discretization and ensembling to safeguard against noise;
and 3) use top-down orthographic projection  of depth information as an additional signal.
For 1), we note that the estimated motion for a given pair of observations is related to the motion estimated for the permuted  observation. %
Two loss terms encourage this relation. %
For 2) we study Dropout~\cite{srivastava2014dropout} in the last two layers of the visual odometry model to safeguard against uncertainty within the egomotion prediction, following~\cite{Kendall2016ModellingUI}. 
We also find depth discretization to be effective.
For 3), we infer an egocentric top-down projection from depth information at each \textit{individual} step. We find that such a simple projection, which is \textit{local} to each step, benefits egomotion estimation. 

On the Habitat Challenge 2020 PointNav benchmark, we show that those three techniques are surprisingly effective, achieving a 71.7\% success rate and a 52.5\% SPL, which improves significantly upon the 64.5\% and 37.7\% SPL  from prior state-of-the-art (SOTA).
Moreover, using VO in a navigation policy also executes 6.4 times faster than prior SOTA.
We perform exhaustive ablations to show the efficacy of each of the three techniques and find that \emph{all the aforementioned techniques} contribute to a more accurate navigation. %

Importantly, we train this visual odometry model separately
instead of learning it online with the policy. Using the VO model as a drop-in replacement for a perfect \texttt{GPS+Compass} permits to \textbf{re-use} navigation policies that were learned with perfect localization information (\ie, with \texttt{GPS+Compass} sensor) without any expensive re-training. Note that  the visual odometry model can be trained for different environment dynamics using a static dataset of only a couple of million frames. In contrast, navigation policies are typically trained using over a billion frames collected using six-months of GPU-time~\cite{Wijmans2020DDPPOLN}.

To summarize, we study three techniques for realistic PointGoal navigation: %
1) leveraging geometric invariances via losses;  %
2) incorporating discretization and ensembling; 
3) using top-down projection of depth information. 

We show: learning such a visual odometry model %
\textit{offline} using only a couple of million frames and  directly replacing the \texttt{GPS+Compass} input of a navigation policy achieves SOTA performance on the standard PointNav benchmark.

\begin{figure*}[t]
\vspace{-0.5cm}
\centering
\hspace*{-0.in}
\includegraphics[width=\linewidth]{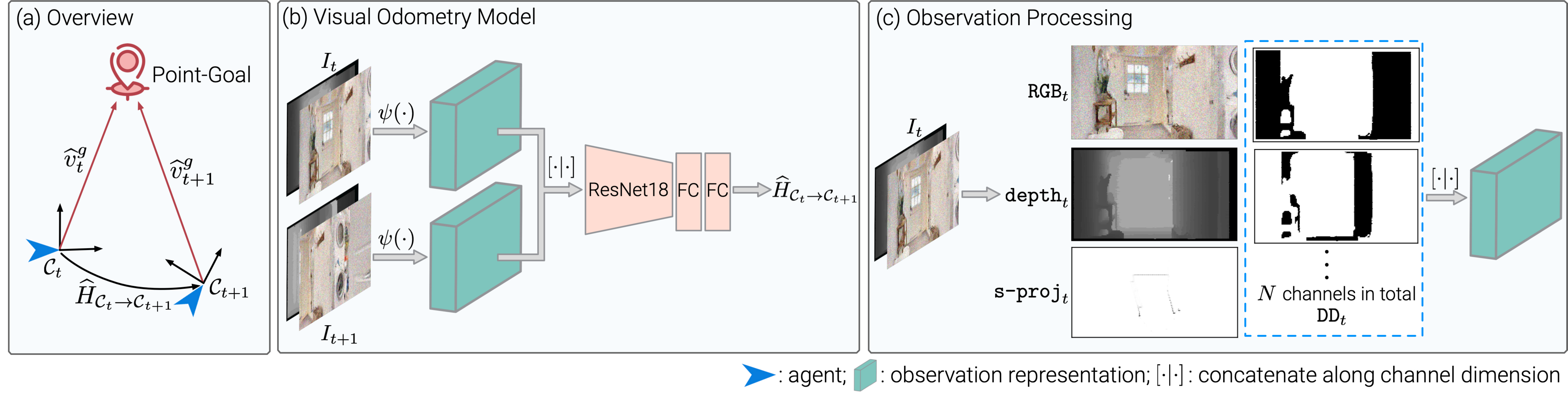}
\vspace{-1.0cm}
\captionsetup{width=1.01\linewidth}
\caption{The studied method.
\textbf{(a)} We estimate the transformation  $\widehat{H}_{\mathcal{C}_t \rightarrow \mathcal{C}_{t+1}} {\in} SE(2)$ in PointGoal navigation (\secref{sec: approach overview}).
\textbf{(b)} The visual odometry (VO) operates on two consecutive egocentric observations ($I_t$, $I_{t+1}$) and yields $\widehat{H}_{\mathcal{C}_t \rightarrow \mathcal{C}_{t+1}}$ (\secref{sec: approach training}). %
\textbf{(c)}
Illustration for $\psi(\cdot)$.
To deal with noise, besides raw $\texttt{RGB}_t$ and $\texttt{depth}_t$, we find discretization $\texttt{d-depth}_t$ (\secref{sec: robust vo}) and top-down projection $\texttt{s-proj}_t$ (\secref{sec: approach top-down map}) to help. %
}
\vspace{-0.5cm}
\label{fig: approach overview}
\end{figure*}

\csection{Related work}

\noindent\textbf{Navigation for embodied tasks.} Recently, there has been a  renewed interest in the field of Embodied AI. The community has built several indoor navigation simulators \cite{Savva2019HabitatAP, xia2020interactive, Savva2017MINOSMI, ai2thor} on top of photo-realistic scans of 3D environments \cite{ai2thor, Chang2017Matterport3DLF, Straub2019TheRD, xiazamirhe2018gibsonenv, Wu2018BuildingGA}. To test a robot's ability to perceive, navigate and interact with the environment, the community has also introduced several tasks~\cite{xia2020interactive, Batra2020ObjectNavRO, Shridhar2019ALFREDAB, Das2018EmbodiedQA, Wijmans2019EmbodiedQA, Narasimhan2020SeeingTU, Anderson2018VisionandLanguageNI, Krantz2020BeyondTN, Thomason2019VisionandDialogN, JainCVPR2019, JainECCV2020, WatkinsValls2019LearningYW, gupta2019cognitive,LiuCORL2019,LiuNEURIPS2020,LiuICML2021,LiuIROS2021} and benchmarks. 
Specifically, Batra \etal~\cite{Batra2020ObjectNavRO} introduce evaluation details for the task of \textit{Object Navigation}, requiring the agent to navigate to a given object class instead of a final point-goal.
Similarly, \textit{Room Navigation} \cite{Narasimhan2020SeeingTU} requires the agent to navigate to a given room type.
More recently, Krantz \etal~\cite{Shridhar2019ALFREDAB, Krantz2020BeyondTN, Thomason2019VisionandDialogN} extend the navigation task to utilize instructions in natural language.
VLN~\cite{Anderson2018OnEO, Krantz2020BeyondTN} and ALFRED~\cite{Shridhar2019ALFREDAB} require the agent to follow a sequence of natural language instructions in order to reach the specified goal. Thomason \etal~\cite{Thomason2019VisionandDialogN} introduce Vision-and-Dialog Navigation that requires back-and-forth communication in order to reach the desired location. 
Jain \etal~\cite{JainCVPR2019,JainECCV2020} develop FurnLift and FurnMove to study visual multi-agent navigation. 
While these tasks differ in their setup, each of them requires the agent to navigate accurately in an environment. Towards this, the agent's navigation policy assumes perfect knowledge of an agent's location and orientation (for example by using a perfect \texttt{GPS+Compass} sensor).
Recently, to alleviate this unrealistic assumption, Datta~\etal~\cite{Datta2020IntegratingEL} propose to estimate egomotion from a pair of depth maps. Like them, we also conduct egomotion estimation from visual observation. However, differently, we study components that improve robustness. As we show in \secref{sec: exp ablations}, without improving robustness to observation and actuation noise, %
the model yields inferior results.

\noindent\textbf{Camera pose estimation and visual odometry (VO).}
Camera pose estimation is related to localization estimation.
\Eg,
direct use of a convolutional neural net (CNN) to estimate relative camera pose was studied~\cite{Zamir2016Generic3R, Laskar2017CameraRB}, following the aforementioned egomotion estimation~\cite{Datta2020IntegratingEL}. These models don't usually consider robustness. 
Meanwhile, in the last few decades, a number of methods have been developed for VO~\cite{6096039, 6153423}. The pipeline typically consists of several steps from  camera calibration, feature selection and matching to motion estimation from correspondences, outlier detection, and bundle adjustment.
More recently, various deep-learning-based architectures have been proposed for VO. For instance, Wang \etal~\cite{Wang2017DeepVOTE} proposed a CNN + recurrent neural net (RNN) to estimate VO in an outdoor environment from RGB input. Because three successive frames in indoor navigation have little overlap, we find sequential training with an RNN to not help. In contrast, we use a faster ResNet-18~\cite{He2016DeepRL} architecture to learn VO from a noisy RGB-D input pair.
Wang \etal~\cite{Wang2019ImprovingLE} leverage the mathematical group property of the rigid motion to learn a VO model for outdoor navigation.
Similarly, we also utilize geometric invariance constraints as a self-supervisory signal during training. In addition, we deliberately utilize representations that make the model robust to observation noise.

To model the agent's uncertainty about its egomotion prediction, Kendall \etal~\cite{Kendall2016ModellingUI} used Dropout~\cite{srivastava2014dropout} after each convolution layer and the penultimate linear layer. At test time, their model uses 40 random samples to get a  robust estimate of the egomotion. 40 forward passes of the model at every time step is prohibitively expensive when used as input to a navigation policy. Moreover, since the input to the VO model is already noisy, adding Dropout to the CNN architecture provides little benefit. Instead, we add Dropout to the \textit{last two} layers of the model, and approximate the effect of averaging the predictions from multiple models by scaling the parameters of the last two layers. This permits robust estimation with a single forward pass. %

\csection{Approach}\label{sec: approach}

We study a simple but effective visual odometry (VO) model, suitable for Embodied AI tasks that predict egomotion from a pair of noisy RGB-D frames. This VO model, which is based solely on classical components, can be used as a drop-in replacement for a perfect \texttt{GPS+Compass} sensor in a downstream navigation task. In the following, an overview is provided before the  components are discussed.

\csubsection{Overview}\label{sec: approach overview}

The model is illustrated in \figref{fig: approach overview}. PointGoal navigation~\cite{Anderson2018OnEO} requires an agent to navigate to a point goal $\bm{v}_t^g$, which is specified relative to the agent's current location at each time step $t$. After the first move, due to noise, the agent only has an estimate $\widehat{\bm{v}}_t^g$ of the relative position. 

Based on the estimated relative coordinates $\widehat{\bm{v}}_t^g$ as well as egocentric observations $I_{\leq t}$ until time $t$, \eg, measurements from an RGB-D sensor, the agent chooses the next action towards the goal. For this, the agent computes a distribution over an action space $\mathcal{A} = \{\text{\turnl},\text{\turnr},\dots\}$, \ie, a policy $\pi(\cdot \vert \widehat{\bm{v}}_t^g, I_{\leq t})$. Upon executing action $a_t \in \mathcal{A}$, the agent's position and orientation change. This  results in a change of the agent's local coordinate system  from $\mathcal{C}_t$ to $\mathcal{C}_{t+1}$. %
Any point's location in coordinate system $\mathcal{C}_t$ can be transformed to that of coordinate system $\mathcal{C}_{t+1}$  using a transformation $H_{\mathcal{C}_t \rightarrow \mathcal{C}_{t+1}}$, which is an element of the group of rigid transformations in the 2D plane, \ie, $SE(2)$.
This assumes that the agent's motion is  planar which holds because an episode is defined on a single floor. 
Note, all techniques  can be extended easily to  $SE(3)$ if required.

However, transformation $H_{\mathcal{C}_t \rightarrow \mathcal{C}_{t+1}}$ is not available because perfect location change measurements are not accessible. Hence, we need to  estimate  $\widehat{H}_{\mathcal{C}_t \rightarrow \mathcal{C}_{t+1}}\in SE(2)$ given the agent's egocentric observations. %
Using the transformation estimate $\widehat{H}_{\mathcal{C}_t \rightarrow \mathcal{C}_{t+1}}$, the agent computes the goal's relative position at time  $t + 1$ from its prior estimate $\widehat{\bm{v}}_t^g$ via
\begin{align}
    \widehat{\bm{v}}_{t+1}^g  = \widehat{H}_{\mathcal{C}_t \rightarrow \mathcal{C}_{t+1}} \cdot \widehat{\bm{v}}_{t}^g. \label{eq: goal position change}
\end{align}
\secref{sec: se2 from vo with geo inv}  discusses how to estimate the transformation $\widehat{H}_{\mathcal{C}_t \rightarrow \mathcal{C}_{t+1}}$ from egocentric observations by using geometric invariances. %
\secref{sec: robust vo} explains  a simple way to make a  visual odometry model  robust to uncertainty in  egomotion estimates. 
Next, \secref{sec: approach top-down map}  discusses a simple method to utilize a top-down projection from egocentric observation as an additional signal.
Finally,  \secref{sec: approach training}  details training.

\csubsection{Geometric Invariances for Visual Odometry}\label{sec: se2 from vo with geo inv}

The goal is to learn a convolutional neural net (CNN) that estimates the transformation $\widehat{H}_{\mathcal{C}_t \rightarrow \mathcal{C}_{t+1}}\in SE(2)$ from a given  pair of egocentric observations $(I_t, I_{t+1})$. 
Formally, an element of $SE(2)$ is specified by  %
a translation $\widehat{\bm{\xi}}_{\mathcal{C}_t \rightarrow \mathcal{C}_{t+1}}\in\mathbb{R}^2$ in the ground plane and an angle $\widehat{\theta}_{\mathcal{C}_t \rightarrow \mathcal{C}_{t+1}} \in \mathbb{R}$, \ie,   
\begin{align}
    \widehat{H}_{\mathcal{C}_t \rightarrow \mathcal{C}_{t+1}} = \begin{bmatrix} \widehat{R}_{\mathcal{C}_t \rightarrow \mathcal{C}_{t+1}} & \widehat{\bm{\xi}}_{\mathcal{C}_t \rightarrow \mathcal{C}_{t+1}} \\ & 1 \end{bmatrix}\label{eq: def of SE(2)},
\end{align}
with $\widehat{R}_{\mathcal{C}_t \rightarrow \mathcal{C}_{t+1}} \!{=}\! \setlength\arraycolsep{2pt} \begin{bmatrix} \cos (\widehat{\theta}_{\mathcal{C}_t \rightarrow \mathcal{C}_{t+1}}) & - \sin (\widehat{\theta}_{\mathcal{C}_t \rightarrow \mathcal{C}_{t+1}}) \\ \sin (\widehat{\theta}_{\mathcal{C}_t \rightarrow \mathcal{C}_{t+1}}) & \cos (\widehat{\theta}_{\mathcal{C}_t \rightarrow \mathcal{C}_{t+1}}) \end{bmatrix} \!{\in} SO(2)$ denoting the estimated rotation matrix from the special orthogonal group.
Given this parameterization, we found $SE(2)$ estimation 
via regression to be effective when using the following loss:
$\mathcal{L}^{\text{reg}}_{\mathcal{C}_t \rightarrow \mathcal{C}_{t+1}} \triangleq$
\begin{align}
    \Vert \bm{\xi}_{\mathcal{C}_t \rightarrow \mathcal{C}_{t+1}} - \widehat{\bm{\xi}}_{\mathcal{C}_t \rightarrow \mathcal{C}_{t+1}} \Vert_2^2 + \Vert \theta_{\mathcal{C}_t \rightarrow \mathcal{C}_{t+1}} \!{-} \widehat{\theta}_{\mathcal{C}_t \rightarrow \mathcal{C}_{t+1}} \Vert_2^2. \label{eq: vo regress loss}
\end{align}
Here, $\bm{\xi}_{\mathcal{C}_t \rightarrow \mathcal{C}_{t+1}}$ and $\theta_{\mathcal{C}_t \rightarrow \mathcal{C}_{t+1}}$ are grounth-truth $SE(2)$ components while $\widehat{\bm{\xi}}_{\mathcal{C}_t \rightarrow \mathcal{C}_{t+1}}$ and $\widehat{\theta}_{\mathcal{C}_t \rightarrow \mathcal{C}_{t + 1}}$ are estimates of the  model $f_\phi$ illustrated in \figref{fig: approach overview}(b), \ie,
\begin{align}
    \left( \widehat{\bm{\xi}}_{\mathcal{C}_t \rightarrow \mathcal{C}_{t+1}}, \widehat{\theta}_{\mathcal{C}_t \rightarrow \mathcal{C}_{t + 1}} \right) = f_{\phi}\left((\psi(I_t), \psi(I_{t+1}))\right).  \label{eq: homography from VO}
\end{align}
Further, $\phi$ refers to parameters of the VO model and $\psi$ denotes a function that processes egocentric observations. The  architecture of the  model will be presented in \secref{sec: approach training}.

Note, use of the loss given in \equref{eq: vo regress loss} is common for learning the parameters of a VO model which often exhibits the structure given in \equref{eq: homography from VO},~\eg,~\cite{Wang2017DeepVOTE, Datta2020IntegratingEL}. %
However, as we show in~\secref{sec: exp ablations}, without specifically accounting for perceptual and actuation noise,  pure regression does not work well. We discuss robustness improvements next. %

Beyond regressing to  ground truth data via the loss given in \equref{eq: vo regress loss},  more information is available in a pair of observations $(I_t, I_{t+1})$. 
To see this, suppose the agent observes $(I_t, I_{t+1})$ followed by $(I_{t+1}, I_{t})$.  %
In this case we know that, in general, the agent returned  to its original location. This is more formally described via the $SE(2)$ invariance  $H_{\mathcal{C}_t \rightarrow \mathcal{C}_{t+1}} H_{\mathcal{C}_{t + 1} \rightarrow \mathcal{C}_{t}} = I_{3 \times 3}$.
Such geometric invariances are ubiquitous. To exploit them, in addition to the regression loss given in \equref{eq: vo regress loss}, we found two additional losses %
during training of a VO model to help:
\begin{align}
    \mathcal{L}^{\text{inv}}_{\mathcal{C}_t \rightarrow \mathcal{C}_{t+1}} \triangleq \mathcal{L}^{\text{inv, rot}}_{\mathcal{C}_t \rightarrow \mathcal{C}_{t+1}} + \mathcal{L}^{\text{inv, trans}}_{\mathcal{C}_t \rightarrow \mathcal{C}_{t+1}}. \label{eq: geo invariance loss}
\end{align}
$\mathcal{L}^{\text{inv, rot}}_{\mathcal{C}_t \rightarrow \mathcal{C}_{t+1}}$ and $\mathcal{L}^{\text{inv, trans}}_{\mathcal{C}_t \rightarrow \mathcal{C}_{t+1}}$ are the rotation and translation invariance loss, which are explained next. 
    
\noindent \textbf{Rotation invariance.} Intuitively, if a  rotation with angle $\theta_{\mathcal{C}_t \rightarrow \mathcal{C}_{t+1}}$  transforms coordinates in $\mathcal{C}_t$ to ones in $\mathcal{C}_{t+1}$, then the inverse coordinate transformation from $\mathcal{C}_{t+1}$ to $\mathcal{C}_{t}$ will be achieved via a rotation with angle $-\theta_{\mathcal{C}_t \rightarrow \mathcal{C}_{t+1}}$, \ie,  $\theta_{\mathcal{C}_{t+1} \rightarrow \mathcal{C}_{t}} = -\theta_{\mathcal{C}_t \rightarrow \mathcal{C}_{t+1}}$.
Consequently, a VO model which receives egocentric observations $(I_{t}, I_{t+1})$ followed by observations $(I_{t+1}, I_{t})$ should be encouraged to predict
$\widehat{\theta}_{\mathcal{C}_t \rightarrow \mathcal{C}_{t + 1}} + \widehat{\theta}_{\mathcal{C}_{t+1} \rightarrow \mathcal{C}_{t}} = 0$. 
This is achieved via the self-supervised learning loss
\begin{align}
    \mathcal{L}^{\text{inv, rot}}_{\mathcal{C}_t \rightarrow \mathcal{C}_{t+1}} \triangleq \big\Vert \widehat{\theta}_{\mathcal{C}_t \rightarrow \mathcal{C}_{t+1}} + \widehat{\theta}_{\mathcal{C}_{t+1} \rightarrow \mathcal{C}_{t}} \big\Vert_2^2.  \label{eq: geo variance loss for rotation}
\end{align}

\noindent \textbf{Translation invariance.} 
The translation invariance property is intuitively similar to the one  for rotation. 
If the transformation from  $\mathcal{C}_t$ to   $\mathcal{C}_{t+1}$ consists of pure translation $\bm{\xi}_{\mathcal{C}_t \rightarrow \mathcal{C}_{t+1}}$, then the reverse transformation from $\mathcal{C}_{t+1}$ to $\mathcal{C}_t$ is simply another translation with $\bm{\xi}_{\mathcal{C}_{t+1} \rightarrow \mathcal{C}_{t}} = - \bm{\xi}_{\mathcal{C}_t \rightarrow \mathcal{C}_{t+1}}$.
This results in the loss $\|\widehat{\bm{\xi}}_{\mathcal{C}_{t} \rightarrow \mathcal{C}_{t+1}} + \widehat{\bm{\xi}}_{\mathcal{C}_{t+1} \rightarrow \mathcal{C}_t}\|_2^2$. 
The relation is slightly more involved when the transformation consists of both rotation and translation. We obtain %
\begin{align}
    \mathcal{L}^{\text{inv, trans}}_{\mathcal{C}_t \rightarrow \mathcal{C}_{t+1}} \triangleq \big\Vert \widehat{\bm{\xi}}_{\mathcal{C}_{t} \rightarrow \mathcal{C}_{t+1}} + \widehat{R}_{\mathcal{C}_t \rightarrow \mathcal{C}_{t+1}} \cdot \widehat{\bm{\xi}}_{\mathcal{C}_{t+1} \rightarrow \mathcal{C}_t} \big\Vert_2^2. \label{eq: geo variance loss for translation}
\end{align}
We provide the formal derivation of the losses in \equref{eq: geo variance loss for rotation} and \equref{eq: geo variance loss for translation} in the appendix.

\csubsection{Robustness to Uncertainty}\label{sec: robust vo}

In addition to leveraging geometric invariances, we found it was important to further increase robustness of the model's $SE(2)$ estimation. This is important because measurements are noisy:
1) visual observations differ even if the camera position and orientation are identical because of observation noises. This makes the processing of observations brittle;
2) perturbations in actuation influence the VO model's prediction since they increase the variance of rotation and translation. For robustness we use two classical techniques: %

\noindent \textbf{Ensemble.} To improve robustness, one can train an ensemble of models.
Averaging predictions over an ensemble typically reduces variance. 
However, reinforcement learning (RL) based navigation systems need billions of samples to train a good policy~\cite{Wijmans2020DDPPOLN}. Since the policy relies on the VO model to provide the  agent's current location estimate, it is important to increase the inference speed and avoid unnecessary computations. %
Therefore, instead of ensembling multiple models, we found it helpful to train one CNN architecture while adding Dropout~\cite{srivastava2014dropout} to the last two fully-connected (FC) layers. This economically resembles the behavior of training a large number of ensembles~\cite{Baldi2013UnderstandingD, Hinton2012ImprovingNN}.
During training, Dropout randomly disables hidden units in the FC layer with a probability $p$, essentially sampling from a collection of sub-networks. During inference, every hidden unit in the FC layer is scaled with the same factor $p$ to mimic the averaging of predictions from multiple sub-networks. 

\noindent \textbf{Depth discretization.} %
In addition, we found depth discretization to yield a more robust representation of the egocentric observation of a range sensor. 
Specifically, a single-channel depth map $\texttt{depth}$ is discretized into representation $\texttt{d-depth}$ with $N$ channels using a one-hot encoding.
Given a pixel of $\texttt{depth}$ at image coordinates  $(x, y)$ we obtain the value of the $i$-th channel of $\texttt{d-depth}$ via
\begin{align}
    \texttt{d-depth}_{i} (x, y) = \mathbbm{1}\left\{ \texttt{depth}(x, y) \in [z_{i-1}, z_i) \right\}, \label{eq: discretized depth}
\end{align}
where $\mathbbm{1}\{\cdot\}$ denotes the indicator function and $\{z_{i-1}, z_i\}$ are endpoints of discretization intervals.
Intuitively,  this  increases the absolute tolerance of the depth uncertainty to $\min_{i} \frac{\vert z_i - z_{i-1} \vert}{2}$ since the same representation will be generated unless a depth entry crosses the  interval boundary. 
Empirically we find an equidistant discretization  into $N$ intervals using end-points
    $z_i = i \cdot (z_{\max} - z_{\min})/N$
to work well. 
Here, $z_{\max}$ and $z_{\min}$ are the maximum (10$m$) and minimum depth (0$m$) value respectively.

\csubsection{Top-Down Projection as Additional Signal}\label{sec: approach top-down map}

\begin{figure}[t]
\centering
\includegraphics[width=\linewidth]{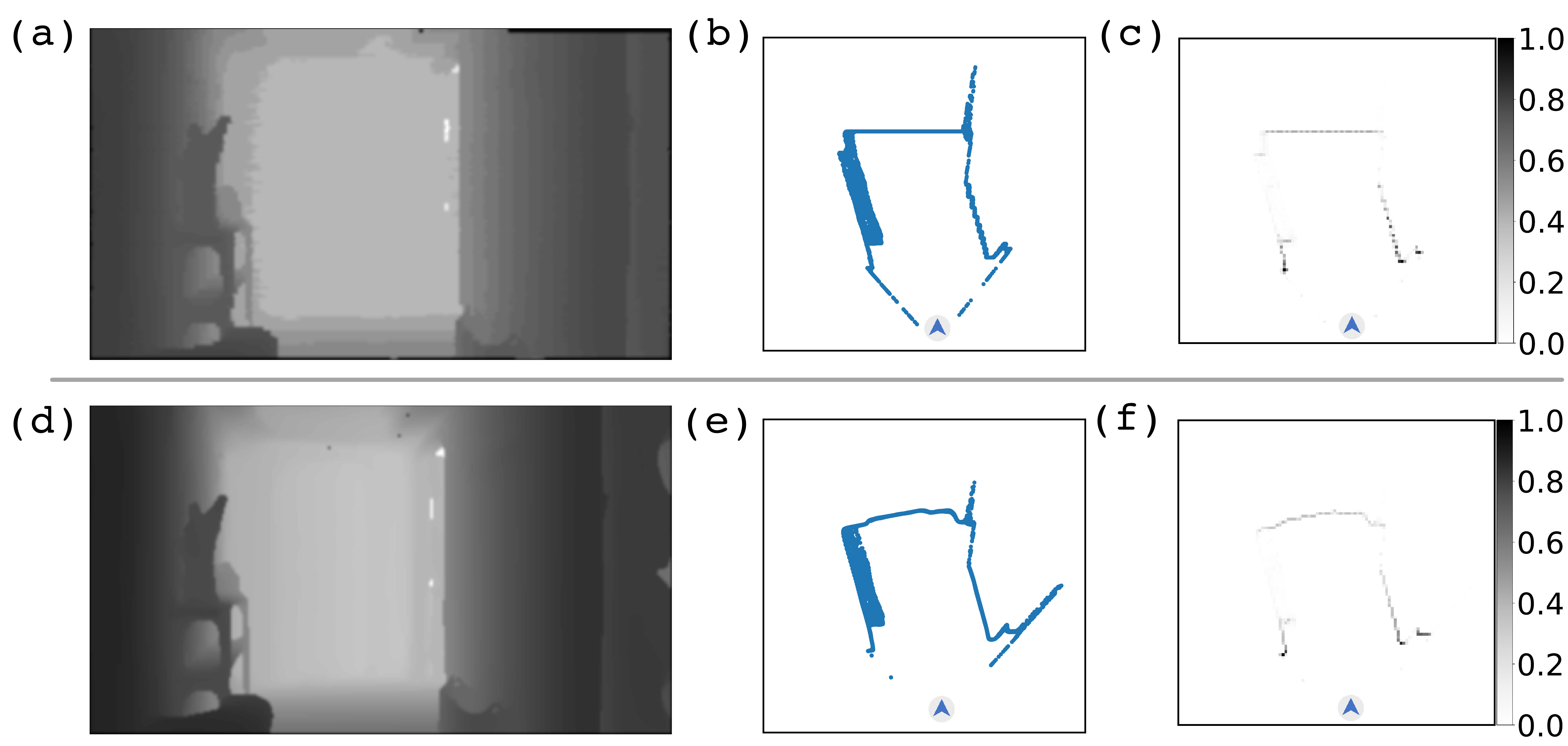}
\vspace{-0.8cm}
\caption{Steps to infer an egocentric top-down projection from depth. Top and bottom rows show inferred top-down projections from noisy and noiseless depth image at the same location. (b,e): top-down scatter plot. (c,f): the  \textit{soft} top-down projection. %
As can be seen, after processing, (c) and (f) share more similarities than (b) and (e), making the representation more robust to depth noises. 
}
\label{fig: top-down-map}
\vspace{-0.6cm}
\end{figure}

Intuitively a map should further improve model robustness. 
However, the key challenge in our  setting: noise in the depth sensor is  fairly subtle and  often hardly visible (see \figref{fig: top-down-map}(a,d)).
But once projected to a 2D layout, the noise manifests itself in gross deviations, holes, and blockages as apparent in \figref{fig: top-down-map}(b,e). %
To address this challenge we use a normalized \textit{soft} projection. 
Normalized \textit{soft} projection $\texttt{s-proj}_t$, shown in \figref{fig: top-down-map}(c,f), resembles the room layout given by the depth maps. Note that they also share more similarities than the projection given in \figref{fig: top-down-map}(b,e). 

We obtain the \textit{soft} projection by 1) mapping depth observations into 3D point clouds, 2) using a  2D top-down orthographic projection, and 3) normalizing the projection with respect to the number of points within each pixel. \textit{Soft} projections are provided as input to the end-to-end trained VO model which learns to use it appropriately.
Details of how to compute soft projections are presented in appendix.

\csubsection{VO Model Architecture, Training Details, and Integration with Navigation Policy}\label{sec: approach training}

\noindent \textbf{Model Architecture.} The visual odometry model $f_\phi$ in \equref{eq: homography from VO} employs a ResNet-18~\cite{He2016DeepRL} backbone to extract visual features. For this we first compute representations from egocentric observation as sketched in \figref{fig: approach overview}(c) via 
\begin{align}
    \psi(I_t) \triangleq (\texttt{RGB}_t, \texttt{depth}_t, \texttt{d-depth}_t, \texttt{s-proj}_t).
\end{align}
Then, we stack $(\psi(I_t), \psi(I_{t+1}))$ along the channel dimension to obtain the ResNet-18 input.
Since $\texttt{RGB}_t$, $\texttt{depth}_t$, $\texttt{d-depth}_t$ and $\texttt{s-proj}_t$ have three, one, $N$ and one channels respectively, the input to the ResNet-18 is a tensor with $(2N+10)$ channels.
To estimate $\widehat{H}_{{\cal C}_t\rightarrow{\cal C}_{t+1}}$, we use two Fully Connected (FC) layers with Dropout on top of the ResNet-18 feature extractor. These FC layers operate on  512-dimensional features %
and produce the output $\left( \hat{\xi}^x_{\mathcal{C}_t \rightarrow \mathcal{C}_{t+1}}, \hat{\xi}^z_{\mathcal{C}_t \rightarrow \mathcal{C}_{t+1}}, \widehat{\theta}_{\mathcal{C}_t \rightarrow \mathcal{C}_{t+1}} \right)$.
Here $\hat{\xi}^z_{\mathcal{C}_t \rightarrow \mathcal{C}_{t+1}}$ refers to the translation in the agent's forward direction while $\hat{\xi}^x_{\mathcal{C}_t \rightarrow \mathcal{C}_{t+1}}$ refers to the translation in the direction perpendicular to the forward motion on the ground plane. 

\noindent \textbf{VO training.} We train the visual odometry model  $f_\phi$ on a dataset $\mathcal{D}_{\text{train}} = \left\{ \left( (I_t, I_{t+1}), \bm{\xi}_{\mathcal{C}_t \rightarrow \mathcal{C}_{t+1}}, \theta_{\mathcal{C}_t \rightarrow \mathcal{C}_{t+1}} \right)  \right\} \triangleq \left\{ d_{\mathcal{C}_t \rightarrow \mathcal{C}_{t+1}} \right\}$.
Each data point consists of a pair of egocentric observations as well as ground-truth translation and rotation angle.
The model is optimized to jointly minimize the regression loss and geometric invariance loss defined in \equref{eq: vo regress loss} and \equref{eq: geo invariance loss}, \ie, we address $\min_\phi \mathcal{L}_{\text{VO}} \triangleq$
\begin{align}
     \hspace{-0.3cm}\sum\limits_{d_{\mathcal{C}_t \rightarrow \mathcal{C}_{t+1}} \in \mathcal{D}_{\text{train}}}\hspace{-0.7cm}  \left[ \lambda_{\text{reg}} \mathcal{L}^{\text{reg}}_{\mathcal{C}_t \rightarrow \mathcal{C}_{t+1}} \!\!+\!  \lambda_\text{inv}^\text{trans} \mathcal{L}^{\text{inv, trans}}_{\mathcal{C}_t \rightarrow \mathcal{C}_{t+1}} \!\!+\! \lambda_\text{inv}^\text{rot} \mathcal{L}^{\text{inv, rot}}_{\mathcal{C}_t \rightarrow \mathcal{C}_{t+1}} \right], \nonumber %
\end{align}
where $\lambda_{\text{reg}}$, $\lambda_\text{inv}^\text{trans}$ and $\lambda_\text{inv}^\text{rot}$ are user-specified hyper-parameters. We set them to $1.0$ in our experiments.
We optimize the VO model with Adam~\cite{Kingma2015AdamAM} using a learning rate of $2.5 {\times} 10^{-4}$. The dropout factor is $p = 0.2$ during training.

\noindent \textbf{Navigation policy training.}
The focus of our work is PointGoal navigation
under realistic conditions,~\ie, noisy observations and actuation as well as no access to \texttt{GPU+Compass} sensors.
In order to demonstrate that VO techniques can be a simple drop-in replacement for a ground truth \texttt{GPS+Compass} sensor, we directly use the navigation policy from~\cite{Wijmans2020DDPPOLN}. %
Specifically, the navigation policy $\pi$ consists of a 2-layer LSTM~\cite{Hochreiter1997LongSM} and uses a ResNet-18~\cite{He2016DeepRL} backbone to process the visual observations.
The policy is \emph{learned independently} of the visual odometry model and has access to perfect location data. During training, at each time step $t$, the policy $\pi$ operates on egocentric observations $I_{\leq t}$, the ground-truth point goal $\bm{v}_t^g$  as well as prior actions $a_{\leq t-1}$ 
, and computes a distribution over the action space $\mathcal{A}$. To learn the policy we use
DD-PPO~\cite{Wijmans2020DDPPOLN},  a distributed version of PPO~\cite{Schulman2017ProximalPO}. We use the same set of  hyper-parameters and reward shaping settings~\cite{Wijmans2020DDPPOLN}, %
which we discuss more in the appendix.

\noindent \textbf{Visual odometry for navigation.} During inference, at every time  $t+1$, the agent obtains an egocentric observation $I_{t+1}$. Together with the previous egocentric observation $I_t$, the VO model $f_\phi$ computes the $SE(2)$ estimate  $\widehat{H}_{\mathcal{C}_t \rightarrow \mathcal{C}_{t+1}}$ using \equref{eq: homography from VO}. Given the relative position estimate $\widehat{\bm{v}}_{t}^g$ from the previous time $t$, the agent updates the current estimate $\widehat{\bm{v}}_{t+1}^g$ via \equref{eq: goal position change} and uses it as policy input.

\vspace{-0.1cm}
\csection{Experiments}
\vspace{-0.2cm}

\begin{figure}[t!]
\vspace{-0.3cm}
\centering
\begin{subfigure}{0.49\textwidth}
\centering
    \includegraphics[width=0.9\linewidth]{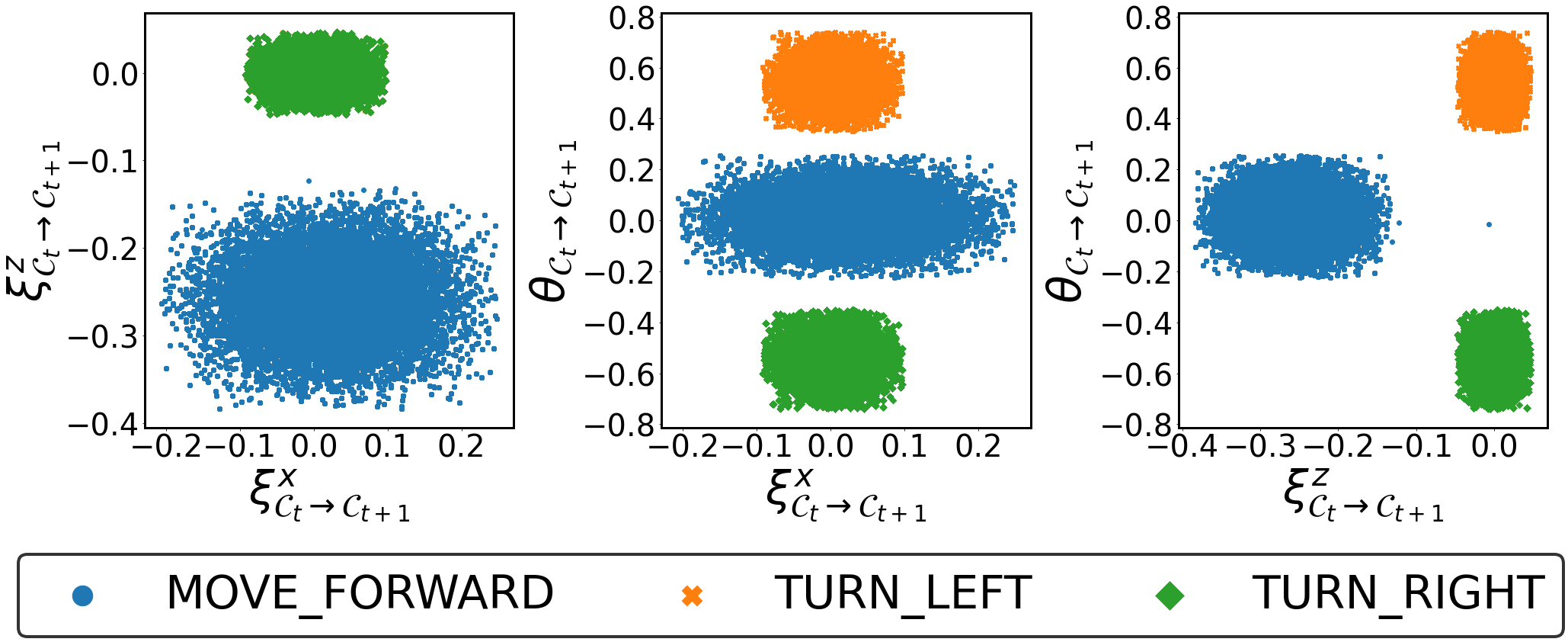}
    \captionsetup{width=0.9\linewidth}
    \caption{Plot of $\bm{\xi}_{\mathcal{C}_t \rightarrow \mathcal{C}_{t+1}}$ and $\theta_{\mathcal{C}_t \rightarrow \mathcal{C}_{t+1}}$ from data \textit{w/o} collisions.}
    \label{fig:exp vo data no collision}
\end{subfigure}%
\hspace{0.01\textwidth}
\begin{subfigure}{0.49\textwidth}
\centering
    \includegraphics[width=0.9\linewidth]{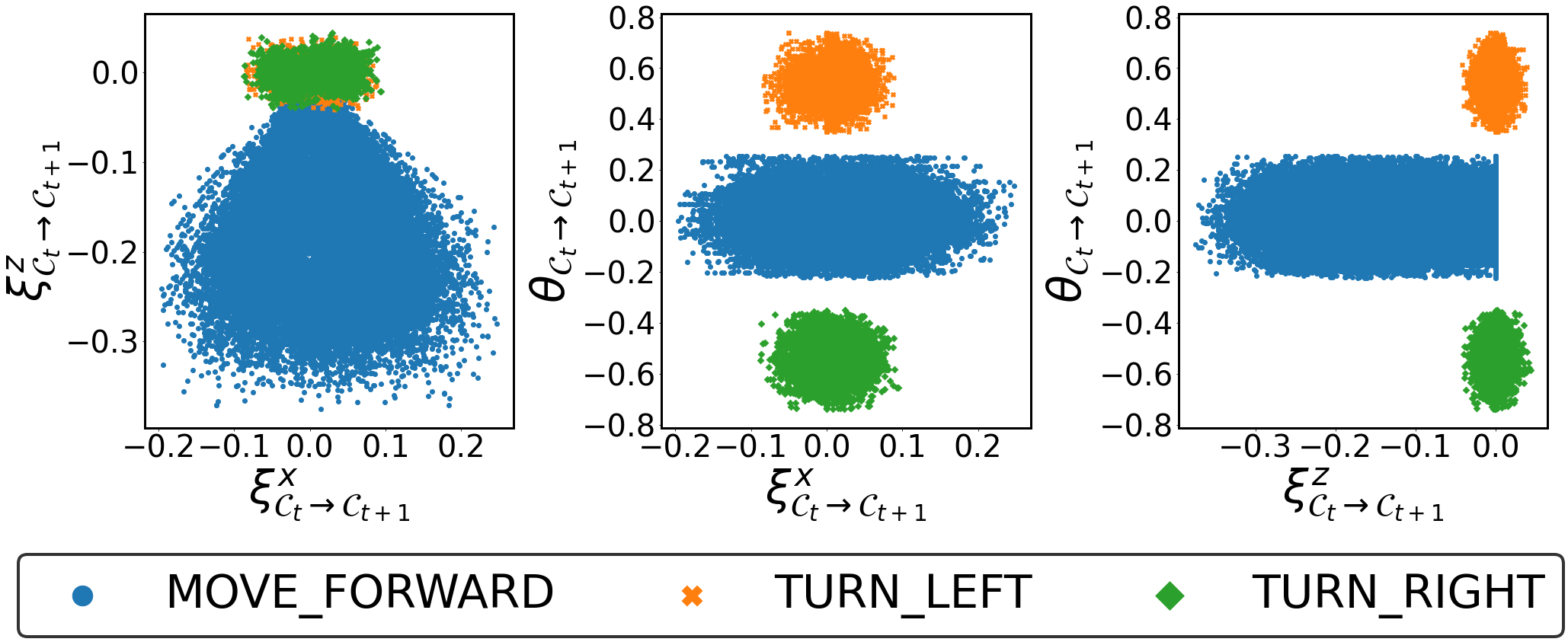}
    \captionsetup{width=0.9\linewidth}
    \caption{Plot of $\bm{\xi}_{\mathcal{C}_t \rightarrow \mathcal{C}_{t+1}}$ and $\theta_{\mathcal{C}_t \rightarrow \mathcal{C}_{t+1}}$ from data \textit{with} collisions.}
    \label{fig:exp vo data with collision}
\end{subfigure}%
\vspace{-0.3cm}
\caption{Three-drawing plot of VO training data $\mathcal{D}_{\text{train}}$ described in \secref{sec: experimental setup}. Different actions have obviously distinct $SE(2)$ distributions, which we find cannot be well-learnt with a unifed model.}
\label{fig:exp vo data}
\vspace{-0.5cm}
\end{figure}

We strive to answer the following questions:
1) to what extent does such a visual odometry (VO) model help navigation?
2) what contributes to  its performance?
We report results on the online Habitat Challenge test split in~\secref{sec: online leaderboard} and conduct ablation on the offline validation split in~\secref{sec: exp ablations}.

\vspace{-0.1cm}
\subsection{Experimental Setup}\label{sec: experimental setup}
\vspace{-0.1cm}
\noindent \textbf{Simulator specification.} All  experiments are conducted using the Habitat simulator~\cite{Savva2019HabitatAP} and we follow the Habitat PointNav Challenge~\cite{HabChalle2020} guidelines for all  studies. We summarize them here and defer  details to the appendix:

\noindent\textbf{Dataset.} We utilize the training data released as part of the Habitat Challenge. It consists of 72 scenes from the Gibson dataset~\cite{Xia2018GibsonER} with a rating of 4 or above (Gibson-$4+$). The offline validation split consists of 14 different scenes which are not part of the training dataset.

\noindent\textbf{Observations.} Similar to a LoCoBot\footnote{\url{http://www.locobot.org/}}, the agent is equipped with an RGB-D camera mounted at a height of 0.88$m$. It has a $70^\circ$ field of view and records egocentric observations of resolution $341 (\text{width}) \times 192 (\text{height})$. The visual observations incorporate a noise model %
\cite{choi2015robust}.

\noindent\textbf{Actuation.}
The action space $\mathcal{A}$ consists of four actions: \fwd which moves the agent forward by $\sim25cm$, \turnl and \turnr which rotate the agent by $\sim30^\circ$, and \texttt{stop}.
The agent exhibits actuation noise modeled after the LoCoBot robot~\cite{pyrobot2019}. 
During collisions, the `sliding' behavior that allows the agent to \textit{slide} along the obstacle instead of stopping is disabled. This more accurately mimics the movement of a real robot~\cite{Kadian2020Sim2RealPD}. \figref{fig:exp vo data} shows how actuation noise and collisions affect an agent's ground-truth translation and rotation for each action type.

\noindent \textbf{VO dataset.} To train the VO model, we create a dataset $\mathcal{D}_{\text{train}}$ of one million data points from 24,286 trajectories uniformly sampled from 72 training scenes.\footnote{Trajectories are shortest paths computed on ground-truth layout map.}
As described in \secref{sec: approach training}, each data point $d_{\mathcal{C}_t \rightarrow \mathcal{C}_{t+1}}$ consists of a pair of observations as well as ground-truth translation and rotation: $\left( (I_t, I_{t+1}), \bm{\xi}_{\mathcal{C}_t \rightarrow \mathcal{C}_{t+1}}, \theta_{\mathcal{C}_t \rightarrow \mathcal{C}_{t+1}} \right)$.
We generate data points from each scene by repeating the following three-step procedure:
1) randomly sample a starting position and orientation of the agent and a navigable PointGoal in the scene;
2) follow the shortest path to navigate from starting point to the point goal;
and 3)  randomly sample data points $d_{\mathcal{C}_t \rightarrow \mathcal{C}_{t+1}}$ along the trajectory. 
We find that due to actuation noise, the action leads to collisions approximately $11.25\%$ of the time. The distribution of the ground-truth translation and rotation in this VO dataset $\mathcal{D}_{\text{train}}$ is illustrated in \figref{fig:exp vo data}.
We observe \fwd, \turnl, and \turnr to have distinct %
distributions. This finding motivates  to train action-specific models, which is effective for this task.

\noindent \textbf{Metrics.} PointGoal Navigation is evaluated on several criteria, summarized by Anderson \etal~\cite{Anderson2018OnEO}.
An episode is considered successful ($S =1$) if the agent stops within 0.36$m$ (2$\times$ the agent radius) of the target global coordinate, otherwise the episode is marked as failed ($S = 0$).
Using the length of the shortest-path trajectory $l$ and the length of an agent's path $l_a$ for an episode, Success Weighted by Path Length (SPL) is defined as $S\frac{l}{\text{max}(l_a, l)}$.
SPL intuitively captures how closely the agent followed the shortest path and successfully completed the episode.
Distance to goal ($d_G$) captures the geodesic distance between the agent and the goal upon episode termination averaged across all episodes.
Finally, the challenge also introduced the new SoftSPL metric~\cite{Datta2020IntegratingEL}: using the starting geodesic distance to the goal $d_{\text{init}}$ and the termination geodesic distance $d_G$, SoftSPL is defined as $(1 - \frac{d_G}{d_{\text{init}}})\frac{l}{\text{max}(l_a, l)}$. It replaces the binary success $S$ with a progress indicator that measures how close the agent gets to the target global coordinate at episode termination.

\begin{table}[t]
\vspace{-0.3cm}
\renewcommand{\arraystretch}{0.9}
\begin{adjustwidth}{0.0cm}{}
\captionsetup{width=\linewidth}
\caption{Online evaluation as of 1:30 am CST, Mar.\ 17th, 2021.
$S$, SPL, and SoftSPL are reported in \%.
}
\vspace{-0.4cm}
\renewcommand\theadfont{}
\centering
{\scriptsize
\begin{tabular}{c@{\hskip 0.5em}l@{\hskip 1em}r@{\hskip 0.9em}r@{\hskip 0.9em}r@{\hskip 0.8em}r@{\hskip 0.8em}r} 
\toprule
Rank & Team & $S\uparrow$ & SPL$\uparrow$ & $d_G\downarrow$ & SoftSPL$\uparrow$ & Time ($h$)$\downarrow$  \\
\midrule
1-1 & \texttt{Ours w/ finetuning}                  & \textbf{71.7} & \textbf{52.5} & 0.802          & \textbf{66.5} & 5.83 \\
1-2  & \texttt{Ours w/o finetuning}                & \textbf{69.8} & \textbf{52.0} & 0.823          & \textbf{65.7} & 6.63 \\
\rowcolor{Gray}
2 & Karkus~\etal~\cite{Karkus2021cvpr}             & 64.5          & 37.7          & \textbf{0.697} & 52.1 & 37.50 \\
3 & Ramakrishnan~\etal~\cite{OccAnticipation}      & 29.0          & 22.0          & 2.567          & 47.3 & 11.06 \\
\rowcolor{Gray}
4 & Information Bottleneck                         & 16.3          & 12.2          & 2.075          & 56.1 & 2.73 \\
5 & Datta~\etal~\cite{Datta2020IntegratingEL}      & 15.7          & 11.9          & 2.232          & 58.6 & \textbf{2.31} \\
\rowcolor{Gray}
6 & cogmodel\_team (39)                            & 1.3           & 0.9           & 4.879          & 30.4 & 5.47 \\
7 & cso                                            & 1.2           & 0.7           & 4.632          & 24.7 & 5.57 \\
\rowcolor{Gray}
8 & UCULab                                         & 0.8           & 0.5           & 6.555          & 10.4 & 15.12 \\
9 & \texttt{Habitat Team}                          & 0.3           & 0.0           & 6.929          & 3.8 & -\\
\toprule
\end{tabular}
}
\label{tab: online eval nav metrics}
\end{adjustwidth}
\vspace{-0.6cm}
\end{table}

\begin{table*}[t]
\vspace{-0.3cm}
\renewcommand{\arraystretch}{1.}
\begin{adjustwidth}{0.0cm}{}
\captionsetup{width=\textwidth}
\caption{Evaluation on the Gibson-$4+$ validation split. %
VO prediction errors are presented in the order of $(\hat{\xi}_{\mathcal{C}_{t} \rightarrow \mathcal{C}_{t+1}}^x, \hat{\xi}_{\mathcal{C}_{t} \rightarrow \mathcal{C}_{t+1}}^z, \widehat{\theta}_{\mathcal{C}_{t} \rightarrow \mathcal{C}_{t+1}})$. Results are reported from three evaluations with different seeds.
We use D as abbreviation for depth.
$S$, SPL, and SoftSPL are reported in \%.
}
\vspace{-0.3cm}
\renewcommand\theadfont{}
\centering
{\scriptsize
\begin{tabular}{l@{\hskip 0.9em}ccc@{\hskip 0.6em}c@{\hskip 0.6em}c@{\hskip 1em}c@{\hskip 0.6em}c@{\hskip 0.6em}r@{\hskip 1em}r@{\hskip 1em}c@{\hskip 1em}r@{\hskip 1em}r@{\hskip 1em}r@{\hskip 1em}r@{\hskip 1.2em}r} 
\toprule
 & \multicolumn{8}{c}{VO} && Policy & \multirow{2}{*}{$S\uparrow$} & \multirow{2}{*}{SPL$\uparrow$} & \multirow{2}{*}{$d_G\downarrow$} & \multirow{2}{*}{SoftSPL$\uparrow$} & \multirow{2}{*}{Pred Error per Step ($e^{-2}$) $\downarrow$}  \\
\cline{2-9}
 & Visual & DD & S-Proj & Dropout & ActInfo & DataAug & GeoInv & \#param (M) && Tune & & & & &  \\
\midrule
0 & \multicolumn{7}{c}{DeepVO~\cite{Wang2017DeepVOTE}} & 100.49 & & & 50{\tiny$\pm$1} & 39{\tiny$\pm$1} & 0.93{\tiny$\pm$0.02} & 65{\tiny$\pm$0} & (2.40, 1.83, 1.62){\tiny$\pm$(0.00, 0.00, 0.01)} \\
\midrule
1 & RGB   &         &        &        &        &        &        & 3.92 && & 52{\tiny$\pm$1} & 39{\tiny$\pm$1} & 0.94{\tiny$\pm$0.01} & 64{\tiny$\pm$1} & (1.96, 1.62, 1.37){\tiny$\pm$(0.02, 0.02, 0.01)} \\ %
2 & D     &         &        &        &        &        &        & 3.92 && & 54{\tiny$\pm$2} & 40{\tiny$\pm$1} & 1.21{\tiny$\pm$0.04} & 61{\tiny$\pm$1} & (1.88, 1.53, 1.38){\tiny$\pm$(0.01, 0.02, 0.02)} \\ %
3 & RGB-D &         &        &        &        &        &        & 3.93 && & 61{\tiny$\pm$1} & 46{\tiny$\pm$1} & 1.14{\tiny$\pm$0.05} & 62{\tiny$\pm$1} & (1.72, 1.10, 1.23){\tiny$\pm$(0.04, 0.00, 0.00)} \\ %
\midrule
4 & RGB-D &         &        & \cmark &        &        &        & 3.93 && & 68{\tiny$\pm$1} & 51{\tiny$\pm$1} & 0.78{\tiny$\pm$0.03} & 66{\tiny$\pm$0} & (1.42, 0.98, 1.03){\tiny$\pm$(0.01, 0.01, 0.02)} \\   %
5 & RGB-D &         &        & \cmark (rnd10)  &&       &        & 3.93 && & 42{\tiny$\pm$1} & 31{\tiny$\pm$1} & 1.64{\tiny$\pm$0.07} & 57{\tiny$\pm$0} & (1.71, 1.35, 1.84){\tiny$\pm$(0.00, 0.01, 0.01)} \\  %
\midrule
6 & RGB-D &         &        & \cmark &        &        &        & 12.4 && & 70{\tiny$\pm$1} & 52{\tiny$\pm$1} & 0.89{\tiny$\pm$0.04} & 65{\tiny$\pm$0} & (1.39, 1.02, 1.01){\tiny$\pm$(0.01, 0.01, 0.01)} \\ %
7 & RGB-D &         &        & \cmark & Embed  &        &        & 12.4 && & 72{\tiny$\pm$0} & 53{\tiny$\pm$0} & 0.83{\tiny$\pm$0.10} & 65{\tiny$\pm$0} & (1.36, 0.89, 0.93){\tiny$\pm$(0.02, 0.01, 0.01)} \\ %
8 & RGB-D &         &        & \cmark & SepAct    &        &        & 3$\times$3.93 && & 75{\tiny$\pm$0} & 56{\tiny$\pm$0} & 0.68{\tiny$\pm$0.06} & 66{\tiny$\pm$0} & (1.24, 0.86, 0.82){\tiny$\pm$(0.00, 0.00, 0.01)} \\
\midrule
9 & RGB-D &         &        & \cmark & SepAct    & \cmark &        & 3$\times$3.93 && & 75{\tiny$\pm$2} & 56{\tiny$\pm$1} & 0.67{\tiny$\pm$0.03} & 66{\tiny$\pm$0} & (1.15, 0.85, 0.78){\tiny$\pm$(0.00, 0.00, 0.01)} \\
10 & RGB-D &        &        & \cmark & SepAct    & \cmark & \cmark & 3$\times$3.93 && & 77{\tiny$\pm$1} & 57{\tiny$\pm$0} & 0.65{\tiny$\pm$0.04} & 67{\tiny$\pm$0} & (1.13, 0.85, 0.76){\tiny$\pm$(0.01, 0.00, 0.01)} \\ %
\midrule
11 & RGB-D & 5      &        & \cmark & SepAct    & \cmark & \cmark & 3$\times$3.96 && & 74{\tiny$\pm$2} & 57{\tiny$\pm$1} & 0.70{\tiny$\pm$0.05} & 68{\tiny$\pm$0} & (1.07, 1.03, 0.69){\tiny$\pm$(0.01, 0.01, 0.01)} \\ %
12 & RGB-D & 10     &        & \cmark & SepAct    & \cmark & \cmark & 3$\times$3.96 && & 79{\tiny$\pm$1} & 60{\tiny$\pm$1} & 0.54{\tiny$\pm$0.00} & 69{\tiny$\pm$0} & (1.08, 0.90, 0.67){\tiny$\pm$(0.00, 0.00, 0.00)} \\ %
13 & RGB-D & 20     &        & \cmark & SepAct    & \cmark & \cmark & 3$\times$3.96 && & 79{\tiny$\pm$0} & 60{\tiny$\pm$0} & 0.52{\tiny$\pm$0.03} & 69{\tiny$\pm$0} & (1.06, 0.85, 0.67){\tiny$\pm$(0.00, 0.00, 0.01)} \\ %
\midrule
14 & D & 10         & \cmark & \cmark & SepAct    & \cmark & \cmark & 3$\times$3.95 && & 72{\tiny$\pm$1} & 55{\tiny$\pm$1} & 0.72{\tiny$\pm$0.01} & 68{\tiny$\pm$0} & (1.40, 0.84, 0.86){\tiny$\pm$(0.00, 0.00, 0.00)} \\ %
15 & RGB-D &        & \cmark & \cmark & SepAct    & \cmark & \cmark & 3$\times$3.93 && & 77{\tiny$\pm$1} & 59{\tiny$\pm$1} & 0.54{\tiny$\pm$0.04} & 70{\tiny$\pm$0} & (1.12, 0.91, 0.72){\tiny$\pm$(0.00, 0.00, 0.00)} \\ %
16 & RGB   & 10     & \cmark & \cmark & SepAct    & \cmark &  \cmark & 3$\times$3.96 && & 79{\tiny$\pm$1} & 61{\tiny$\pm$1} & 0.52{\tiny$\pm$0.02} & 69{\tiny$\pm$0} & (1.18, 0.78, 0.75){\tiny$\pm$(0.00, 0.00, 0.01)} \\ %
17 & RGB   &        &        & \cmark & SepAct    & \cmark &  \cmark & 3$\times$3.92 && & 59{\tiny$\pm$2} & 45{\tiny$\pm$1} & 0.74{\tiny$\pm$0.05} & 67{\tiny$\pm$0} & (2.02, 1.73, 1.15){\tiny$\pm$(0.01, 0.00, 0.01)}\\
\midrule
\rowcolor{Gray}
18 & RGB-D & 10     & \cmark & \cmark & SepAct    & \cmark & \cmark & 3$\times$3.96 && & \textbf{81}{\tiny$\pm$1} & \textbf{62}{\tiny$\pm$1} & \textbf{0.51}{\tiny$\pm$0.03} & \textbf{70}{\tiny$\pm$0} & (1.10, 0.84, 0.68){\tiny$\pm$(0.00, 0.00, 0.01)} \\  %
\midrule
\rowcolor{Gray}
19 & RGB-D & 10     & \cmark & \cmark & SepAct    & \cmark & \cmark & 3$\times$3.96 && \cmark & \textbf{82}{\tiny$\pm$1} & \textbf{63}{\tiny$\pm$1} & \textbf{0.48}{\tiny$\pm$0.00} & \textbf{71}{\tiny$\pm$0} & (1.08, 0.85, 0.65){\tiny$\pm$(0.01, 0.01, 0.00)} \\
\midrule
20 & \multicolumn{7}{c}{Ground-Truth}      && & & 97{\tiny$\pm$0} & 71{\tiny$\pm$0} & 0.42{\tiny$\pm$0.02} & 70{\tiny$\pm$0} &  \\
\toprule
\end{tabular}
}
\label{tab: offline eval combine}
\end{adjustwidth}
\vspace{-0.7cm}
\end{table*}

\csubsection{Results on the Online Leaderboard}\label{sec: online leaderboard}

\tabref{tab: online eval nav metrics} shows the results from the online leaderboard on the test-standard split\footnote{\url{https://evalai.cloudcv.org/web/challenges/challenge-page/580/leaderboard/1631}} of the Habitat Challenge PointNav Benchmark 2020 (we will call it Challenge hereafter).
The 2020 winners achieved a success of 29.0\% by integrating occupancy anticipation~\cite{OccAnticipation} into active neural SLAM~\cite{Chaplot2020LearningTE} (Rank 3 in~\tabref{tab: online eval nav metrics}).
Karkus~\etal ~\cite{Karkus2021cvpr} proposed an end-to-end particle SLAM-net to generate a global occupancy map and utilized D$^\ast$ to plan the path, pushing  SOTA to 64.5\% in Nov.\ 2020 (Rank 2 in~\tabref{tab: online eval nav metrics}).
Our approach of training a visual odometry model taking into account robustness as discussed in~\secref{sec: approach} and aforementioned action-specific design \textit{improves  SOTA to 71.7\%}.
Specifically, we evaluate the VO model quality in two settings:
1) direct integration into a pre-trained navigation policy as a drop-in module;
2) fine-tuning of a pre-trained policy \wrt the VO using a small budget.\footnote{We finetuned the policy using 14.7 million frames, instead of billions of frames required to train a policy.}
Rank 1-1 and 1-2 in~\tabref{tab: online eval nav metrics} verify that combining all of the discussed techniques achieves state-of-the-art performance on three out of four metrics, irrespective of fine-tuning.
Besides success rate, it improves SPL by $14.8$ points (from $37.7\%$ to $52.5\%$).
Regarding SoftSPL, it improves 7.9 points (from $58.6\%$ of Rank 5 to $66.5\%$).
Note, VO in the navigation policy executes evaluation 6.4 times faster than Rank 2~\cite{Karkus2021cvpr} (5.83~\vs~37.50 hours) and 1.9 times faster than Rank 3~\cite{OccAnticipation} (5.83~\vs~11.06 hours).

\vspace{-0.1cm}
\csubsection{Ablations}\label{sec: exp ablations}
\vspace{-0.1cm}
To better understand the role of each technique, we perform an extensive ablation study (Row 1 - 19) in \tabref{tab: offline eval combine}.
Specifically, we ablate over all combinations of: 
1) visual sensors (RGB and/or depth); 
2) geometric invariance learning discussed in~\secref{sec: se2 from vo with geo inv}; 
3) %
dropout and depth discretization detailed in~\secref{sec: robust vo};
4) %
soft egocentric projection described in \secref{sec: approach top-down map}; 
5) use of action-specific models  mentioned in~\secref{sec: experimental setup}.
Note, the VO is a \emph{drop-in replacement}  in a pretrained navigation policy in Row 1 - 18 (no fine-tuning).

Evaluation is conducted on 994 episodes from 14 validation scenes, each of which provides 71 episodes. We abbreviate the discretized depth $\texttt{d-depth}$ defined in \equref{eq: discretized depth} via \textit{DD}  and use \textit{S-Proj} to indicate use of the top-down projection discussed in \secref{sec: approach top-down map}. In addition to the aforementioned metrics, we also report the VO prediction absolute error per navigation step for $\hat{\xi}^x_{\mathcal{C}_t \rightarrow \mathcal{C}_{t+1}}$, $\hat{\xi}^z_{\mathcal{C}_t \rightarrow \mathcal{C}_{t+1}}$, and $\widehat{\theta}_{\mathcal{C}_t \rightarrow \mathcal{C}_{t+1}}$, discussed in \secref{sec: approach training}.

Note, prior work showed that without \texttt{GPS+Compass} sensor, the policy achieves 0 SPL after 100-million-frame training and 15\% SPL after 2.5-billion-frame training~\cite{Wijmans2020DDPPOLN}.\footnote{Note, \cite{Wijmans2020DDPPOLN} do not train with observation and actuation noise, 15\% SPL is hence an upper bound.}
In contrast, when evaluated with
perfect \texttt{GPS+Compass} sensors under noisy observations and actuations (Row 19 in~\tabref{tab: offline eval combine}), the policy  obtains 71\% SPL with 97\% success rate.
We now discuss to what extent each of the techniques detailed in~\secref{sec: approach} and \secref{sec: experimental setup} shrinks this gap.

\noindent \textbf{Both RGB and Depth observations help visual odometry.} Row 1 - 3 study the role of visual modalities for visual odometry. We find that the RGB-D model (Row 3) has lower per-step prediction error and higher navigation success rate compared to RGB-only (Row 1) and depth-only (Row 2) VO models.
This finding overturns the accepted conventional wisdom in this sub-field~\cite{Wijmans2020DDPPOLN, Datta2020IntegratingEL} that RGB models overfit and  depth-only models outperform RGB-D models.
We find that both RGB and depth observations are important for training a visual odometry model. 
We hypothesize that RGB enables better feature matching between frames.
In addition, this result highlights the advantage of separately training VO model and navigation policy as they capture different features of the input observations.

\noindent \textbf{Adding Dropout in the VO model learns a more robust egomotion estimator.}
We find significant performance improvements when using Dropout to economically mimic an ensemble for more robust egomotion prediction.
Empirical results demonstrate the effectiveness of this design
as success rate and SPL improve 7 and 5 points respectively (Row 3~\vs 4 in \tabref{tab: offline eval combine}).
To demonstrate the advantage of a single forward pass over multiple ones during inference, we conduct additional experiments (Row 5). We  randomly select hidden units with ratio $p$ at test time and average results of 10 forward passes.
Apart from the apparent  inferior results (success 42\%~\vs~68\% for Row 5~\vs~4), the VO model's throughput drastically decreases from 118.8 FPS (frames per second) for Row 4 to 8.45 FPS for Row 5.

\noindent \textbf{Learning action-specific models  helps.}
As mentioned in \secref{sec: experimental setup}, action-specific model design (SepAct) improves the navigation's success rate from 68\% (\tabref{tab: offline eval combine} Row 4) to 75\% (Row 8) while improving other metrics as well. Furthermore, SepAct increases the accuracy of VO prediction for all three components. To validate that this improvement is due to SepAct and not from an increased parameter count, we add two more ablations (Row 6 and 7):
1) in Row 6, a VO model was trained with $3\times$ more parameters (12.4M) than the single-action model (3.93M) by increasing the ResNet-18 layer width  twofold. Note, we observed that wider models work better than deeper ones for PointGoal navigation. Comparing Row 8 to Row 6, we can see that simply adding more parameters performs worse in success rate (75\% to 70\%), SPL (56\% to 52\%) as well as VO prediction;
2) in Row 7, instead of training separate models, we exposed a \textit{unified} model to action information via an action embedding. Performance increases from Row 6 to Row 7 on success rate (70\% to 72\%), SPL (52\% to 53\%) and VO prediction, establishing that action information is important for such a task. However, the worse results compared to Row 8 (success and SPL both drop 3 points) confirm the effectiveness of  SepAct.

\noindent \textbf{Encouraging geometric invariance in the egomotion predictions is helpful.}
As discussed in~\secref{sec: se2 from vo with geo inv}, the VO model can benefit from exploiting the geometric invariance properties.
Row 8~\vs Row 10 in~\tabref{tab: offline eval combine} confirms the effectiveness of this technique: success rate and SPL improves two and one points respectively.
To verify that this improvement indeed stems from the self-supervised signal instead of data augmentation, we conduct an ablation with a simple data augmentation for invertible actions like \turnl and \turnr.
Specifically, when training the VO model for \turnl, apart from using the original pair of frames collected for \turnl, we also utilize the frames collected for the \turnr action by reversing the pair of observations and computing the corresponding ground-truth $SE(2)$. 
Similar processing is applied when training the VO model for \turnr. We do not apply data augmentation to \fwd since there do not exist situations where agents move backward. 
\tabref{tab: offline eval combine} shows that sole data augmentation does not help navigation performance (success and SPL remain the same across \tabref{tab: offline eval combine} Row 9 \vs~Row 8).

\noindent \textbf{Depth discretization and top-down projection account for more satisfactory results.} As shown in \secref{sec: robust vo}, we add depth discretization \texttt{d-depth} to obtain a more robust egomotion estimation.
Indeed, use of \texttt{d-depth} increases success rate from 77\% to 79\% and SPL from 57\% to 60\% (\tabref{tab: offline eval combine} Row 10 \vs~Row 12).
To understand whether the performance is robust to the number of \texttt{d-depth}'s channels, we ablate over 5, 10, and 20 channels in Row 11 - 13. The results verify that coarse discretization harms the navigation performance (Row 11 \vs~Row 12). However, when the granularity increases (20 channels instead of 10), the gains from adding more channels are not significant (Row 12 \vs~Row 13).
Meanwhile, use of the soft projection discussed in \secref{sec: approach top-down map} benefits PointGoal navigation  improving success and SPL  by two points (Row 12 \vs~Row 18 in \tabref{tab: offline eval combine}). 

\noindent \textbf{Every representation feature is indispensable for VO.} 
To verify that \emph{every input feature is required}, we conduct  ablations by removing each feature (RGB, D, DD, S-Proj) from the VO model. Specifically, if we ignore the RGB representation, success drops from 81\% to 72\% (Row 14~\vs 18 in~\tabref{tab: offline eval combine}).
Trends are similar for depth (success drops two points from Row 16~\vs 18), depth discretization (success drops 4 points from Row 15~\vs 18), and egocentric top-down projection (Row 12~\vs 18).
Moreover, we train our VO without any depth-related parts,~\ie,~depth, DD, and S-Proj (Row 17).
Row 17~\vs~18 again verifies the importance of depth.
Note, the difference between Row 1 and Row 17 is that Row 17 uses Dropout, SepAct, DataAug, and GeoInv.
the 7-point success rate improvement validates those technique's usefulness (Row 1~\vs~Row 17).

\noindent \textbf{Tuning RL policy with VO further improves performance.} The VO model's efficiency (36 FPS for Row 18 in \tabref{tab: offline eval combine} on a 3.10GHz Intel Xeon Gold 6254 CPU and an Nvidia GeForce RTX 2080 Ti GPU)
permits fine-tuning of the RL policy with respect to the VO module. %
In~\tabref{tab: offline eval combine}'s Row 19, we observe overall best performance across  all criteria after tuning the RL policy with only 14.7 million frames, which is much more affordable than billions of frames~\cite{Wijmans2020DDPPOLN}.

\noindent \textbf{Comparison to other VO methods.}
We further compare to DeepVO~\cite{Wang2017DeepVOTE}, a supervised RNN-based VO, on PointGoal Navigation. Please see the appendix for implementation details. We train DeepVO on our collected dataset. We found DeepVO to fall short of the  simplest VO model as success rate drops from our 52\% to 50\% (Row 0 \vs 1 in~\tabref{tab: offline eval combine}). We hypothesize that the RNN does not perform well due to little overlap between consecutive frames. %

\begin{figure}[t!]
\centering
\begin{minipage}[t]{.25\linewidth}
    \subcaptionbox{SPL 85\%.\label{fig: exp qualitative a}}
    {\includegraphics[width=\linewidth]{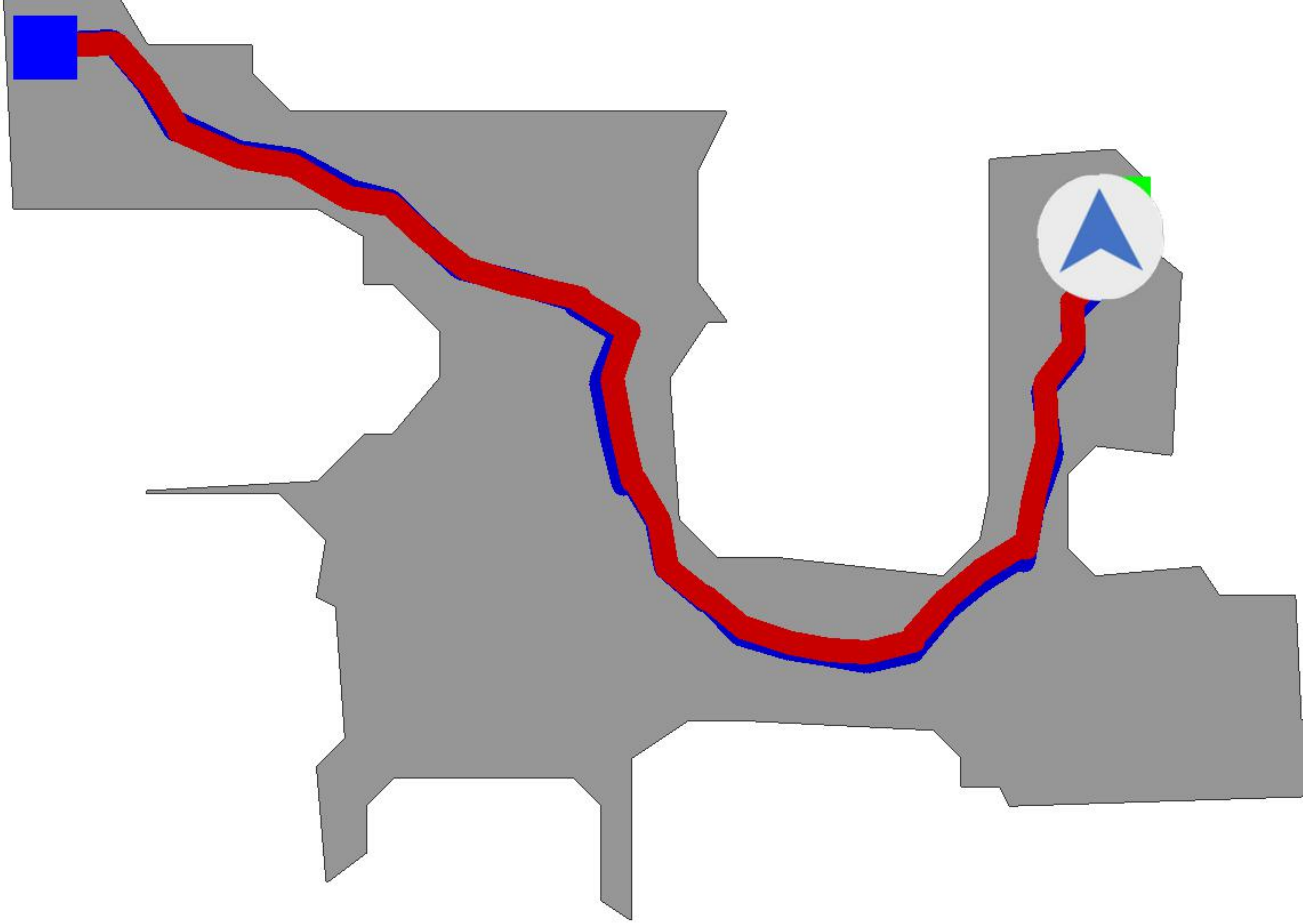}}%
\end{minipage}%
\hfill
\begin{minipage}[t]{.32\linewidth}
    \subcaptionbox{SPL 80\%.\label{fig: exp qualitative b}}
    {\includegraphics[width=\linewidth]{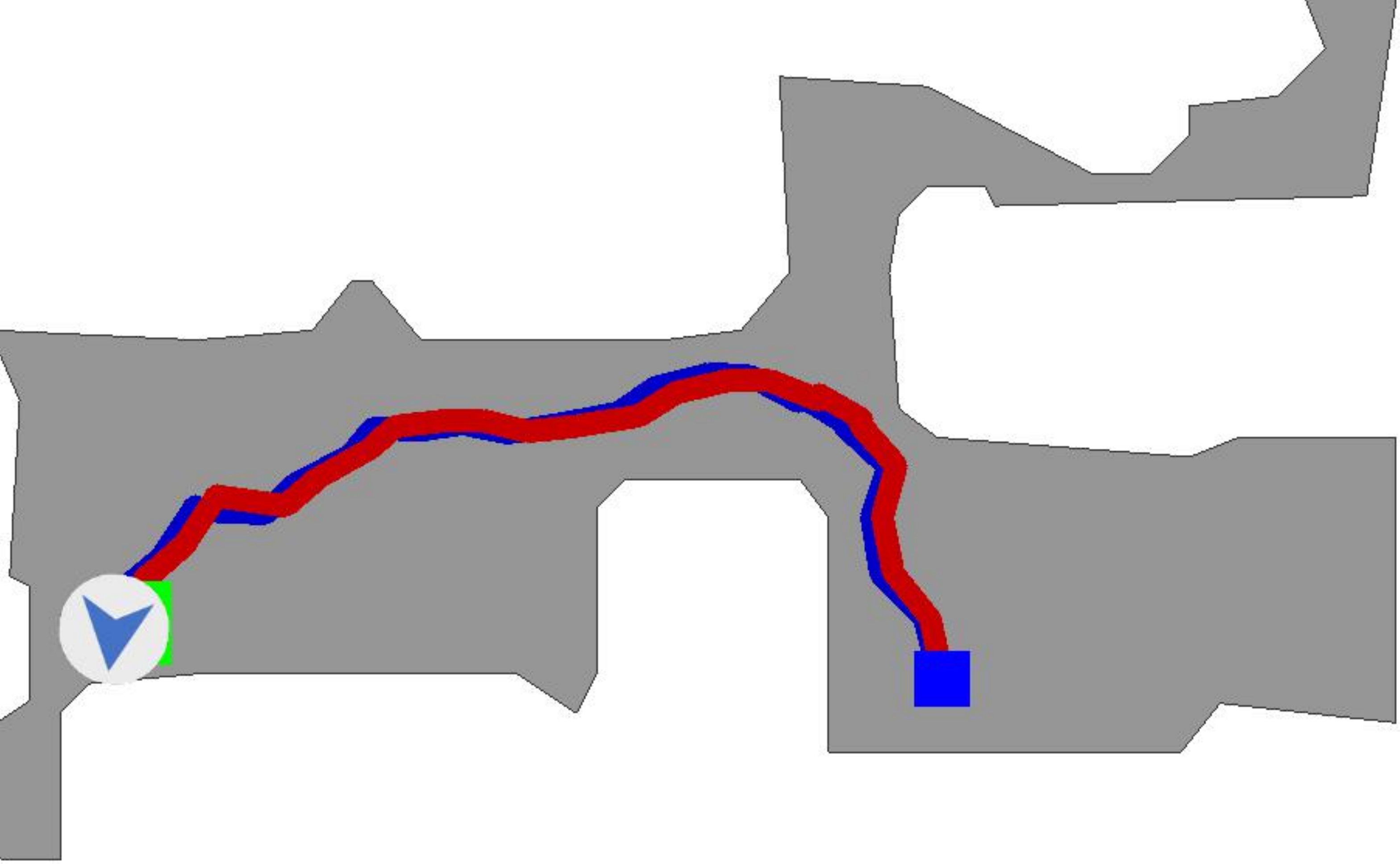}}%
\end{minipage}%
\hfill
\begin{minipage}[t]{.32\linewidth}
    \subcaptionbox{SPL 85\%.\label{fig: exp qualitative c}}
    {\includegraphics[width=\linewidth]{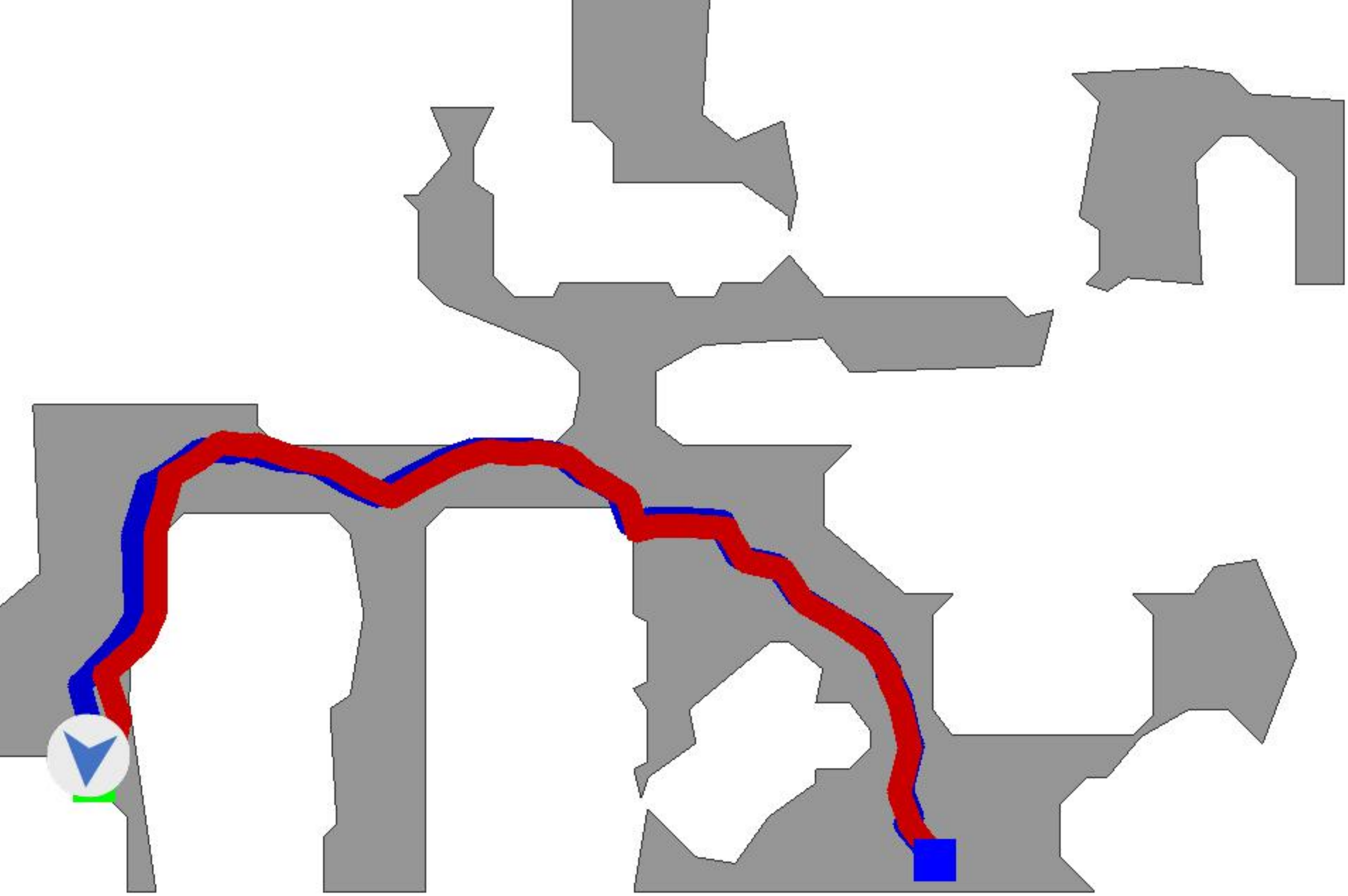}}%
\end{minipage}%
\vspace{-0.3cm}
\caption{Qualitative results. Agent is asked to navigate from {\color{blue}blue square} to {\color{ForestGreen}green square}. {\color{blue}Blue curve} is the actual path the agent takes while {\color{red}red curve} is based on the agent's estimate of its location from the VO model by integrating over $SE(2)$ estimation of each step.
}
\vspace{-0.5cm}
\label{fig: exp qualitative}
\end{figure}

\csubsection{Qualitative Results}\label{sec: exp qualitative}
\figref{fig: exp qualitative} shows several successful trajectories that overlay the ground-truth top-down map. We show that integrating VO techniques into a navigation policy permits to accurately guide the agent towards the point goal. For example, in \figref{fig: exp qualitative c}, the VO model is able to precisely estimate $SE(2)$ around corners and in case of collisions.  More examples and failure cases are available in the appendix.

\vspace{-0.2cm}
\csection{Conclusion}
\vspace{-0.1cm}
To conclude, we find classical visual odometry techniques to be surprisingly effective and yield a very strong baseline for Embodied PointGoal Navigation in a realistic setting  (noisy actuation and perception; no localization sensor). %
\\
\textbf{Acknowledgements:} This work is supported in part by NSF under Grant \#1718221, 2008387, 2045586, MRI \#1725729, and NIFA  2020-67021-32799, UIUC, Samsung, Amazon, 3M, and Cisco Systems Inc.\ (Gift Award CG 1377144 - thanks for access to Arcetri).

{\small
\bibliographystyle{ieee_fullname}
\bibliography{references}

\begin{thebibliography}{10}\itemsep=-1pt

\bibitem{HabChalle2020}
{\em Habitat Challenge 2020}, 2020.

\bibitem{Anderson2018OnEO}
Peter Anderson, Angel~X. Chang, Devendra~Singh Chaplot, Alexey Dosovitskiy,
  Saurabh Gupta, Vladlen Koltun, Jana Kosecka, Jitendra Malik, Roozbeh
  Mottaghi, Manolis Savva, and Amir~Roshan Zamir.
\newblock On evaluation of embodied navigation agents.
\newblock {\em ArXiv}, 2018.

\bibitem{Anderson2018VisionandLanguageNI}
Peter Anderson, Qi Wu, Damien Teney, Jake Bruce, Mark Johnson, Niko
  S{\"u}nderhauf, Ian~D. Reid, Stephen Gould, and Anton van~den Hengel.
\newblock Vision-and-language navigation: Interpreting visually-grounded
  navigation instructions in real environments.
\newblock {\em CVPR}, 2018.

\bibitem{Baldi2013UnderstandingD}
Pierre Baldi and Peter Sadowski.
\newblock Understanding dropout.
\newblock {\em NIPS}, 2013.

\bibitem{Batra2020ObjectNavRO}
Dhruv Batra, Aaron Gokaslan, Aniruddha Kembhavi, Oleksandr Maksymets, Roozbeh
  Mottaghi, Manolis Savva, Alexander Toshev, and Erik Wijmans.
\newblock {ObjectNav Revisited: On Evaluation of Embodied Agents Navigating to
  Objects}.
\newblock {\em ArXiv}, 2020.

\bibitem{Chang2017Matterport3DLF}
Angel~X. Chang, Angela Dai, Thomas~A. Funkhouser, Maciej Halber, Matthias
  Nie{\ss}ner, Manolis Savva, Shuran Song, Andy Zeng, and Yinda Zhang.
\newblock {Matterport3D}: Learning from rgb-d data in indoor environments.
\newblock {\em 3DV}, 2017.

\bibitem{Chaplot2020LearningTE}
Devendra~Singh Chaplot, Dhiraj Gandhi, Saurabh Gupta, Abhinav Gupta, and R.
  Salakhutdinov.
\newblock {Learning to Explore using Active Neural SLAM}.
\newblock {\em ICLR}, 2020.

\bibitem{Chen2020ASO}
C. Chen and \etal.
\newblock A survey on deep learning for localization and mapping: Towards the
  age of spatial machine intelligence.
\newblock {\em arxiv/2006.12567}, 2020.

\bibitem{choi2015robust}
Sungjoon Choi, Qian-Yi Zhou, and Vladlen Koltun.
\newblock Robust reconstruction of indoor scenes.
\newblock {\em CVPR}, 2015.

\bibitem{Das2018EmbodiedQA}
Abhishek Das, Samyak Datta, Georgia Gkioxari, Stefan Lee, Devi Parikh, and
  Dhruv Batra.
\newblock Embodied question answering.
\newblock {\em CVPR}, 2018.

\bibitem{Datta2020IntegratingEL}
Samyak Datta, Oleksandr Maksymets, Judy Hoffman, Stefan Lee, Dhruv Batra, and
  Devi Parikh.
\newblock {Integrating Egocentric Localization for More Realistic Point-Goal
  Navigation Agents}.
\newblock {\em CoRL}, 2020.

\bibitem{DeTone2018SuperPointSI}
Daniel DeTone and \etal.
\newblock Superpoint: Self-supervised interest point detection and description.
\newblock {\em CVPRW}, 2018.

\bibitem{Duggal2019DeepPrunerLE}
Shivam Duggal, Shenlong Wang, W. Ma, R. Hu, and R. Urtasun.
\newblock {DeepPruner: Learning Efficient Stereo Matching via Differentiable
  PatchMatch}.
\newblock {\em ICCV}, 2019.

\bibitem{6153423}
F. {Fraundorfer} and D. {Scaramuzza}.
\newblock Visual odometry : Part ii: Matching, robustness, optimization, and
  applications.
\newblock {\em IEEE Robotics Automation Magazine}, 19(2), 2012.

\bibitem{Geiger2012AreWR}
Andreas Geiger, Philip Lenz, and R. Urtasun.
\newblock {Are We Ready for Autonomous Driving? The KITTI Vision Benchmark
  Suite}.
\newblock {\em CVPR}, 2012.

\bibitem{gupta2019cognitive}
Saurabh Gupta, Varun Tolani, James Davidson, Sergey Levine, Rahul Sukthankar,
  and Jitendra Malik.
\newblock Cognitive mapping and planning for visual navigation.
\newblock {\em IJCV}, 2019.

\bibitem{He2016DeepRL}
Kaiming He, Xiangyu Zhang, Shaoqing Ren, and Jian Sun.
\newblock Deep residual learning for image recognition.
\newblock {\em CVPR}, 2016.

\bibitem{Hinton2012ImprovingNN}
Geoffrey~E. Hinton, Nitish Srivastava, A. Krizhevsky, Ilya Sutskever, and R.
  Salakhutdinov.
\newblock {Improving Neural Networks by Preventing Co-Adaptation of Feature
  Detectors}.
\newblock {\em ArXiv}, 2012.

\bibitem{Hochreiter1997LongSM}
Sepp Hochreiter and J{\"u}rgen Schmidhuber.
\newblock Long short-term memory.
\newblock {\em Neural Computation}, 1997.

\bibitem{Ioffe2015BatchNA}
Sergey Ioffe and Christian Szegedy.
\newblock Batch normalization: Accelerating deep network training by reducing
  internal covariate shift.
\newblock {\em ArXiv}, 2015.

\bibitem{JainECCV2020}
U. Jain$^\ast$, L. Weihs$^\ast$, E. Kolve, A. Farhadi, S. Lazebnik, A.
  Kembhavi, and A.~G. Schwing.
\newblock {A Cordial Sync: Going Beyond Marginal Policies For Multi-Agent
  Embodied Tasks}.
\newblock {\em ECCV}, 2020.

\bibitem{JainCVPR2019}
U. Jain$^\ast$, L. Weihs$^\ast$, E. Kolve, M. Rastegrari, S. Lazebnik, A.
  Farhadi, A.~G. Schwing, and A. Kembhavi.
\newblock {Two Body Problem: Collaborative Visual Task Completion}.
\newblock {\em CVPR}, 2019.

\bibitem{Kadian2020Sim2RealPD}
Abhishek Kadian, Joanne Truong, Aaron Gokaslan, Alexander Clegg, Erik Wijmans,
  Stefan Lee, M. Savva, S. Chernova, and Dhruv Batra.
\newblock {Sim2Real Predictivity: Does Evaluation in Simulation Predict
  Real-World Performance?}
\newblock {\em IROS}, 2020.

\bibitem{Karkus2021cvpr}
Peter Karkus, Shaojun Cai, and David Hsu.
\newblock {Particle SLAM-Net for Visual Navigation}.
\newblock {\em CVPR}, 2021.

\bibitem{Kendall2016ModellingUI}
Alex Kendall and Roberto Cipolla.
\newblock Modelling uncertainty in deep learning for camera relocalization.
\newblock {\em ICRA}, 2016.

\bibitem{Kingma2015AdamAM}
Diederik~P. Kingma and Jimmy Ba.
\newblock Adam: A method for stochastic optimization.
\newblock {\em ArXiv}, 2015.

\bibitem{ai2thor}
Eric Kolve, Roozbeh Mottaghi, Winson Han, Eli VanderBilt, Luca Weihs, Alvaro
  Herrasti, Daniel Gordon, Yuke Zhu, Abhinav Gupta, and Ali Farhadi.
\newblock {AI2-THOR: An Interactive 3D Environment for Visual AI}.
\newblock {\em ArXiv}, 2017.

\bibitem{Krantz2020BeyondTN}
Jacob Krantz, Erik Wijmans, Arjun Majumdar, Dhruv Batra, and Stefan Lee.
\newblock Beyond the nav-graph: Vision-and-language navigation in continuous
  environments.
\newblock {\em ECCV}, 2020.

\bibitem{Laga2020ASO}
Hamid Laga, Laurent~Valentin Jospin, Farid Boussa{\"i}d, and M. Bennamoun.
\newblock A survey on deep learning techniques for stereo-based depth
  estimation.
\newblock {\em IEEE Transactions on Pattern Analysis and Machine Intelligence},
  2020.

\bibitem{Laskar2017CameraRB}
Zakaria Laskar, Iaroslav Melekhov, Surya Kalia, and Juho Kannala.
\newblock {Camera Relocalization by Computing Pairwise Relative Poses Using
  Convolutional Neural Network}.
\newblock {\em ICCV Workshop}, 2017.

\bibitem{LiuICML2021}
I.-J. Liu, U. Jain, R. Yeh, and A.~G. Schwing.
\newblock {Cooperative Exploration for Multi-Agent Deep Reinforcement
  Learning}.
\newblock In {\em Proc. ICML}, 2021.

\bibitem{LiuIROS2021}
I.-J. Liu, Z. Ren, R. Yeh, and A.~G. Schwing.
\newblock {Semantic Tracklets: An Object-Centric Representation for Visual
  Multi-Agent Reinforcement Learning}.
\newblock In {\em Proc. IROS}, 2021.

\bibitem{LiuNEURIPS2020}
I.-J. Liu, R. Yeh, and A.~G. Schwing.
\newblock {High-Throughput Synchronous Deep RL}.
\newblock In {\em Proc. NeurIPS}, 2020.

\bibitem{LiuCORL2019}
I.-J. Liu$^\ast$, R. Yeh$^\ast$, and A.~G. Schwing.
\newblock {PIC: Permutation Invariant Critic for Multi-Agent Deep Reinforcement
  Learning}.
\newblock In {\em Proc. CORL}, 2019.
\newblock $^\ast$ equal contribution.

\bibitem{pyrobot2019}
Adithyavairavan Murali, Tao Chen, Kalyan~Vasudev Alwala, Dhiraj Gandhi, Lerrel
  Pinto, Saurabh Gupta, and Abhinav Gupta.
\newblock {PyRobot}: An open-source robotics framework for research and
  benchmarking.
\newblock {\em ArXiv}, 2019.

\bibitem{Narasimhan2020SeeingTU}
Medhini Narasimhan, Erik Wijmans, Xinlei Chen, Trevor Darrell, Dhruv Batra,
  Devi Parikh, and Amanpreet Singh.
\newblock Seeing the un-scene: Learning amodal semantic maps for room
  navigation.
\newblock {\em ECCV}, 2020.

\bibitem{Nilsson1984ShakeyTR}
N. Nilsson.
\newblock {Shakey the Robot}.
\newblock 1984.

\bibitem{OccAnticipation}
Santhosh~K. Ramakrishnan, Z. Al-Halah, and K. Grauman.
\newblock {Occupancy Anticipation for Efficient Exploration and Navigation}.
\newblock {\em ECCV}, 2020.

\bibitem{Sarlin2020SuperGlueLF}
Paul-Edouard Sarlin and \etal.
\newblock Superglue: Learning feature matching with graph neural networks.
\newblock {\em CVPR}, 2020.

\bibitem{Savva2017MINOSMI}
Manolis Savva, Angel~X. Chang, Alexey Dosovitskiy, Thomas~A. Funkhouser, and
  Vladlen Koltun.
\newblock {MINOS}: Multimodal indoor simulator for navigation in complex
  environments.
\newblock {\em ArXiv}, 2017.

\bibitem{Savva2019HabitatAP}
Manolis Savva, Abhishek Kadian, Oleksandr Maksymets, Yili Zhao, Erik Wijmans,
  Bhavana Jain, Julian Straub, Jia Liu, Vladlen Koltun, Jitendra Malik, Devi
  Parikh, and Dhruv Batra.
\newblock Habitat: A platform for embodied ai research.
\newblock {\em ICCV}, 2019.

\bibitem{6096039}
D. {Scaramuzza} and F. {Fraundorfer}.
\newblock Visual odometry [tutorial].
\newblock {\em IEEE Robotics Automation Magazine}, 18(4), 2011.

\bibitem{Schulman2016HighDimensionalCC}
John Schulman, Philipp Moritz, Sergey Levine, Michael~I. Jordan, and Pieter
  Abbeel.
\newblock High-dimensional continuous control using generalized advantage
  estimation.
\newblock {\em ArXiv}, 2016.

\bibitem{Schulman2017ProximalPO}
John Schulman, Filip Wolski, Prafulla Dhariwal, Alec Radford, and Oleg Klimov.
\newblock Proximal policy optimization algorithms.
\newblock {\em ArXiv}, 2017.

\bibitem{Shridhar2019ALFREDAB}
Mohit Shridhar, Jesse Thomason, Daniel Gordon, Yonatan Bisk, Winson Han,
  Roozbeh Mottaghi, Luke Zettlemoyer, and Dieter Fox.
\newblock Alfred: A benchmark for interpreting grounded instructions for
  everyday tasks.
\newblock {\em ArXiv}, 2019.

\bibitem{srivastava2014dropout}
Nitish Srivastava, Geoffrey Hinton, Alex Krizhevsky, Ilya Sutskever, and Ruslan
  Salakhutdinov.
\newblock Dropout: a simple way to prevent neural networks from overfitting.
\newblock {\em JMLR}, 2014.

\bibitem{Straub2019TheRD}
Julian Straub, Thomas Whelan, Lingni Ma, Yufan Chen, Erik Wijmans, Simon Green,
  Jakob~J. Engel, Raul Mur-Artal, Carl Ren, Shobhit Verma, Anton Clarkson,
  Mingfei Yan, Brian Budge, Yajie Yan, Xiaqing Pan, June Yon, Yuyang Zou,
  Kimberly Leon, Nigel Carter, Jesus Briales, Tyler Gillingham, Elias Mueggler,
  Luis Pesqueira, Manolis Savva, Dhruv Batra, Hauke~M. Strasdat, Renzo~De
  Nardi, Michael Goesele, Steven Lovegrove, and Richard~A. Newcombe.
\newblock {The Replica Dataset: A Digital Replica of Indoor Spaces}.
\newblock {\em ArXiv}, 2019.

\bibitem{Thomason2019VisionandDialogN}
Jesse Thomason, Michael Murray, Maya Cakmak, and Luke Zettlemoyer.
\newblock Vision-and-dialog navigation.
\newblock {\em ArXiv}, 2019.

\bibitem{Wang2017DeepVOTE}
Sen Wang, Ronald Clark, Hongkai Wen, and Agathoniki Trigoni.
\newblock {DeepVO}: Towards end-to-end visual odometry with deep recurrent
  convolutional neural networks.
\newblock {\em ICRA}, 2017.

\bibitem{Wang2019ImprovingLE}
Xiangwei Wang, Daniel Maturana, Shichao Yang, Wenshan Wang, Qijun Chen, and
  Sebastian~A. Scherer.
\newblock Improving learning-based ego-motion estimation with
  homomorphism-based losses and drift correction.
\newblock {\em IROS}, 2019.

\bibitem{WatkinsValls2019LearningYW}
David Watkins-Valls, Jingxi Xu, Nicholas~R. Waytowich, and Peter~K. Allen.
\newblock Learning your way without map or compass: Panoramic target driven
  visual navigation.
\newblock {\em ArXiv}, 2019.

\bibitem{Wijmans2019EmbodiedQA}
Erik Wijmans, Samyak Datta, Oleksandr Maksymets, Abhishek Das, Georgia
  Gkioxari, Stefan Lee, Irfan Essa, Devi Parikh, and Dhruv Batra.
\newblock Embodied question answering in photorealistic environments with point
  cloud perception.
\newblock {\em CVPR}, 2019.

\bibitem{Wijmans2020DDPPOLN}
Erik Wijmans, Abhishek Kadian, Ari~S. Morcos, Stefan Lee, Irfan Essa, D.
  Parikh, Manolis Savva, and Dhruv Batra.
\newblock {DD-PPO}: Learning near-perfect pointgoal navigators from 2.5 billion
  frames.
\newblock {\em ICLR}, 2020.

\bibitem{Wu2018GroupN}
Yuxin Wu and Kaiming He.
\newblock {Group Normalization}.
\newblock {\em ECCV}, 2018.

\bibitem{Wu2018BuildingGA}
Yi Wu, Yuxin Wu, Georgia Gkioxari, and Yuandong Tian.
\newblock Building generalizable agents with a realistic and rich 3d
  environment.
\newblock {\em ArXiv}, 2018.

\bibitem{xiazamirhe2018gibsonenv}
Fei Xia, Amir R.~Zamir, Zhi-Yang He, Alexander Sax, Jitendra Malik, and Silvio
  Savarese.
\newblock Gibson {Env}: real-world perception for embodied agents.
\newblock {\em CVPR}, 2018.

\bibitem{xia2020interactive}
Fei Xia, William~B Shen, Chengshu Li, Priya Kasimbeg, Micael~Edmond Tchapmi,
  Alexander Toshev, Roberto Mart{\'\i}n-Mart{\'\i}n, and Silvio Savarese.
\newblock Interactive gibson benchmark: A benchmark for interactive navigation
  in cluttered environments.
\newblock {\em IEEE Robotics and Automation Letters}, 2020.

\bibitem{Xia2018GibsonER}
Fei Xia, Amir~Roshan Zamir, Zhi-Yang He, Alexander Sax, Jitendra Malik, and
  Silvio Savarese.
\newblock Gibson env: Real-world perception for embodied agents.
\newblock {\em CVPR}, 2018.

\bibitem{Zamir2016Generic3R}
A. Zamir, T. Wekel, Pulkit Agrawal, Colin Wei, Jitendra Malik, and S. Savarese.
\newblock {Generic 3D Representation via Pose Estimation and Matching}.
\newblock {\em ECCV}, 2016.

\end{thebibliography}
}

\beginsupplement
\appendix

\twocolumn[
\section*{\Large\centering Appendix: \\The Surprising Effectiveness of Visual Odometry Techniques \\for Embodied PointGoal Navigation}\vspace{0.2cm}]

\renewcommand{\thesection}{\Alph{section}}

This appendix is structured as follows:

\begin{itemize}
    \item \secref{supp: sec: geometric loss derivation} provides the formal derivation of the geometric invariance loss described in \secref{sec: se2 from vo with geo inv}.
    \item \secref{supp: sec: ego map} describes the technical details to generate the egocentric top-down projection discussed in \secref{sec: approach top-down map}.
    \item \secref{supp: sec: nav train details} describes the navigation policy's architecture and hyperparameters used for training.
    \item \secref{supp: sec: vo train} gives details about the visual odometry model's training and inference. 
    \item \secref{supp: sec: kitti} states implementation details of DeepVO as well as our model's performance on KITTI.
    \item \secref{supp: sec: vo from depth} demonstrates that we cannot accurately estimate relative pose from depth due to sensor's noises.
    \item \secref{supp: sec: more qualitative} provides more qualitative results to demonstrate the performance of our  model.
\end{itemize}

\section{Formal Derivation for Geometric Invariance Loss}\label{supp: sec: geometric loss derivation}

Recall that we predict $\widehat{H}_{\mathcal{C}_t \rightarrow \mathcal{C}_{t+1}} \in SE(2)$ from two consecutive egocentric observations $(I_t, I_{t+1})$. Intuitively, invariance is obtained when observing $(I_t, I_{t+1})$ followed by $(I_{t+1}, I_{t})$. 
Due to the invertibility of  transformations between coordinate systems $\mathcal{C}_t$ and $\mathcal{C}_{t+1}$, we have the following relation between ground-truth transformations: 
\begin{align}
    H_{\mathcal{C}_t \rightarrow \mathcal{C}_{t+1}} H_{\mathcal{C}_{t + 1} \rightarrow \mathcal{C}_{t}} = I_{3 \times 3}, \label{supp: eq: se2 invertibility}
\end{align}
where $ I_{3 \times 3}$ is the three-dimensional identity matrix.

Meanwhile, an element from $SE(2)$ is defined as follows:
\begin{align}
    H_{\mathcal{C}_t \rightarrow \mathcal{C}_{t+1}}
    &\triangleq \begin{bmatrix} R_{\mathcal{C}_t \rightarrow \mathcal{C}_{t+1}} & \bm{\xi}_{\mathcal{C}_t \rightarrow \mathcal{C}_{t+1}} \\  & 1 \end{bmatrix} \nonumber \\
    \text{where~~}  R_{\mathcal{C}_t \rightarrow \mathcal{C}_{t+1}} &= \begin{bmatrix} \cos (\theta_{\mathcal{C}_t \rightarrow \mathcal{C}_{t+1}}) & - \sin (\theta_{\mathcal{C}_t \rightarrow \mathcal{C}_{t+1}}) \\ \sin (\theta_{\mathcal{C}_t \rightarrow \mathcal{C}_{t+1}}) & \cos (\theta_{\mathcal{C}_t \rightarrow \mathcal{C}_{t+1}}) \end{bmatrix}. \label{supp: eq: se2 form}
\end{align}
Note, the rotation matrix can be computed via $R_{\mathcal{C}_t \rightarrow \mathcal{C}_{t+1}} = \exp\left( \texttt{alg}(\theta_{\mathcal{C}_t \rightarrow \mathcal{C}_{t+1}}) \right)$, \ie,  by applying the exponential map $\exp$ on $\texttt{alg}: \mathbb{R} \mapsto \mathbb{R}^{2 \times 2}$, the function that maps an angle from $\mathbb{R}$ to an element of the Lie algebra $\mathfrak{so}(2)$, namely $\texttt{alg}(\theta) = \theta \begin{bmatrix} 0 & - 1 \\ 1 & 0 \end{bmatrix}$. When replacing the rotation matrix in \equref{supp: eq: se2 form} with this representation and expanding the relation given in \equref{supp: eq: se2 invertibility}, we obtain:
\begin{align}
    & \begin{bmatrix} \exp\left( \texttt{alg}(\theta_{\mathcal{C}_t \rightarrow \mathcal{C}_{t+1}}) \right) & \bm{\xi}_{\mathcal{C}_{t} \rightarrow \mathcal{C}_{t+1}} \\ & 1\end{bmatrix} \nonumber \\
    &\quad \cdot \begin{bmatrix} \exp\left( \texttt{alg}(\theta_{\mathcal{C}_{t+1} \rightarrow \mathcal{C}_t}) \right) & \bm{\xi}_{\mathcal{C}_{t+1} \rightarrow \mathcal{C}_t} \\ & 1 \end{bmatrix} = I_{3 \times 3}.
\end{align}
After multiplying out the left-hand side we obtain the following system of equations:
\begin{align}
      \begin{cases}
          \exp\left( \texttt{alg}(\theta_{\mathcal{C}_t \rightarrow \mathcal{C}_{t+1}} + \theta_{\mathcal{C}_{t+1} \rightarrow \mathcal{C}_{t}}) \right) = I_{2 \times 2} \\
          \exp\left( \texttt{alg}( \theta_{\mathcal{C}_t \rightarrow \mathcal{C}_{t+1}}) \right) \cdot \bm{\xi}_{\mathcal{C}_{t+1} \rightarrow \mathcal{C}_t} + \bm{\xi}_{\mathcal{C}_{t} \rightarrow \mathcal{C}_{t+1}} = \bm{0}
      \end{cases} .
\end{align}
Upon simplification, this results in
\begin{align}
      \begin{cases}
          \theta_{\mathcal{C}_t \rightarrow \mathcal{C}_{t+1}} + \theta_{\mathcal{C}_{t+1} \rightarrow \mathcal{C}_{t}} = 0 \\
          \exp\left( \texttt{alg}( \theta_{\mathcal{C}_t \rightarrow \mathcal{C}_{t+1}}) \right) \cdot \bm{\xi}_{\mathcal{C}_{t+1} \rightarrow \mathcal{C}_t} + \bm{\xi}_{\mathcal{C}_{t} \rightarrow \mathcal{C}_{t+1}} = \bm{0}
      \end{cases} , \label{supp: eq: geo invariance deduction}
\end{align}
which were used in \equref{eq: geo variance loss for rotation} and \equref{eq: geo variance loss for translation} of the main manuscript to encourage the geometric invariance via the two losses:
\begin{align}
    \mathcal{L}^{\text{inv, rot}}_{\mathcal{C}_t \rightarrow \mathcal{C}_{t+1}} &\triangleq \big\Vert \widehat{\theta}_{\mathcal{C}_t \rightarrow \mathcal{C}_{t+1}} + \widehat{\theta}_{\mathcal{C}_{t+1} \rightarrow \mathcal{C}_{t}} \big\Vert_2^2.  \label{supp: eq: geo variance loss for rotation} \\
    \mathcal{L}^{\text{inv, trans}}_{\mathcal{C}_t \rightarrow \mathcal{C}_{t+1}} &\triangleq \big\Vert \exp\left( \texttt{alg}( \widehat{\theta}_{\mathcal{C}_t \rightarrow \mathcal{C}_{t+1}}) \right) \cdot \widehat{\bm{\xi}}_{\mathcal{C}_{t+1} \rightarrow \mathcal{C}_t} + \widehat{\bm{\xi}}_{\mathcal{C}_{t} \rightarrow \mathcal{C}_{t+1}} \big\Vert_2^2 \nonumber \\
    &= \big\Vert \widehat{R}_{\mathcal{C}_t \rightarrow \mathcal{C}_{t+1}} \cdot \widehat{\bm{\xi}}_{\mathcal{C}_{t+1} \rightarrow \mathcal{C}_t} + \widehat{\bm{\xi}}_{\mathcal{C}_{t} \rightarrow \mathcal{C}_{t+1}} \big\Vert_2^2. \label{supp: eq: geo variance loss for translation}
\end{align}
This concludes the derivation. 

\section{Technical Details for Generating Egocentric Top-Down Projection}\label{supp: sec: ego map} 

We describe details on how to compute the egocentric top-down projection discussed in \secref{sec: approach top-down map}. 

\noindent\textbf{From depth map to 3D point.} Given a pixel of the depth map $\texttt{depth}$ at image coordinates $(u, v)$\footnote{We follow common practice and let $+U$ point downward while $+V$ points to the right.}, we  obtain the 3D point's Cartesian coordinates in the camera coordinate system from:
\begin{align}
    (x, y, z)^T &= h(u, v, \texttt{depth}(u, v)) \nonumber \\
                &= (K^{-1} \cdot (u + 0.5, v + 0.5, 1)^T) \cdot \texttt{depth}(u, v) \label{supp eq: point cloud from depth},
\end{align}
where $h(\cdot, \cdot, \cdot)$ represents the function for generating 3D coordinates\footnote{Following common practice, $+X$ points to the right, $+Y$ points  upward and $+Z$ points  backward.}. Here $K \in \mathbb{R}^{3\times 3}$ is the intrinsic matrix assumed to be known and $\texttt{depth}(u, v)$ denotes the $z$-buffer value at $(u, v)$. Note $u + 0.5$ and $v + 0.5$ are used to compute the 3D point from the center of the pixel and $(u + 0.5, v + 0.5, 1)$ is the homogeneous coordinate. Further, $z = \texttt{depth}(u, v)$.

\noindent\textbf{Computing bounding box for point clouds.} After generating 3D point clouds, we  obtain a bounding box for those 3D points. Specifically, 1) for the Cartesian $Z$ axis, we have  $z_{\text{min}}$ and $z_{\text{max}}$. They refer to the minimum and the maximum depth values, which are specified by the sensor; 2) for the Cartesian $X$ axis, we have $x_{\text{min}}$ and $x_{\text{max}}$ which come from leftmost/rightmost pixels in depth observation \texttt{depth}.
These values will be utilized to compute pixel coordinates in the next step.

\noindent\textbf{Computing 2D pixel coordinates of top-down projection.}
As mentioned in \secref{sec: approach overview}, we assume that the agent's motion is planar. Therefore, we ignore coordinates in the direction perpendicular to the plane. Concretely, we use coordinates  $(x, z)$. Therefore, we obtain pixel coordinates in top-down projection for such a point as $(\texttt{row}, \texttt{col})$, where $\texttt{row} = \lfloor H \cdot \frac{z - z_{\text{min}}}{z_{\text{max}} - z_{\text{min}}} \rfloor$ and $\texttt{col} = \lfloor W \cdot \frac{x - x_{\text{min}}}{x_{\text{max}} - x_{\text{min}}} \rfloor$, where $H \times W$ represents the top-down projection's resolution.

\noindent\textbf{Generating \textit{soft} top-down projection.}
1) For every pixel in $\texttt{depth}$, we  repeat the aforementioned steps to compute the corresponding pixel coordinates $(\texttt{row}, \texttt{col})$ in the top-down projection. 
2) We count the number of points which fall into each $(\texttt{row}, \texttt{col})$ cell. A \textit{soft} egocentric top-down projection $\texttt{s-proj}$ is obtained by normalizing the count to the range of $[0, 1]$.

\section{Navigative Policy Training Details}\label{supp: sec: nav train details}

In \tabref{supp: tab: hyperparameter}, we provide training details of the navigation policy used in our experiments. We explain the strucutre of our policy in the following paragraphs.

\noindent\textbf{Visual encoder.} We use ResNet-18~\cite{He2016DeepRL} as our visual feature extractor to process an egocentric observation of size $341 (\text{width}) \times 192 (\text{height})$. Following~\cite{Wijmans2020DDPPOLN}, we replace every BatchNorm~\cite{Ioffe2015BatchNA} layer with GroupNorm~\cite{Wu2018GroupN} to deal with  highly-correlated trajectories in on-policy RL and massively distributed training. A \texttt{2x2-AvgPool} layer is added before ResNet-18 so that the effective resolution is $170 \times 96$. ResNet-18 produces a $256 \times 6 \times 3$ feature map, which is converted to a $114 \times 6 \times 3$ feature map through a \texttt{3x3-Conv} layer.

\noindent\textbf{Point-Goal encoder.} At each time step $t$, the agent receives the point-goal's relative position $\bm{v}_t^g$  or $\widehat{\bm{v}}_t^g$ in polar coordinate form. Similar to~\cite{Wijmans2020DDPPOLN}, we convert the polar coordinates into a magnitude and a unit vector to alleviate the discontinuity at the $x$-axis in polar coordinates. A subsequent fully-connected layer transforms it into a 32-dimensional representation.

\noindent\textbf{Navigation Policy.} The 2-layer LSTM in the  navigation policy takes three inputs: 
1) a 512-dimensional vector of egocentric observations, which is obtained by flattening the $114 \times 6 \times 3$ feature map from the visual encoder into a 2052-dimensional vector and then feeding it into a fully-connected layer; 
2) a 32-dimensional output of the goal encoder; 
3) a 32-dimensional embedding of the previous action (or the start-token when beginning a new episode). The output of the 2-layer LSTM is fed into a fully-connected layer, obtaining a distribution
over the action space and an estimate of the value function.

\begin{table}[!ht]
\renewcommand{\arraystretch}{1.0}
\begin{adjustwidth}{0.0cm}{}
\captionsetup{width=\textwidth}
\caption{Hyperparameters.}
\centering
\begin{tabular}{lr}
\toprule
\textbf{Hyperparameter} & \textbf{Value} \\
\midrule
\multicolumn{2}{c}{\textit{PPO (DD-PPO)}} \\
\midrule
Clip parameter~\cite{Schulman2017ProximalPO} & 0.2 \\
Rollout timesteps & 128 \\
Epochs & 2 \\
Value loss coefficient & 0.5 \\
Discount factor ($\gamma$) & 0.99 \\
GAE parameter ($\lambda$)~\cite{Schulman2016HighDimensionalCC}  &  0.95 \\
Normalize advantage & False \\
Preemption threshold~\cite{Wijmans2020DDPPOLN} & 0.6 \\

\midrule
\multicolumn{2}{c}{\textit{Training}} \\
\midrule
Optimizer & Adam~\cite{Kingma2015AdamAM} \\
$(\beta_1, \beta_2)$ for Adam & $(0.9, 0.999)$ \\
Learning rate & $2.5e^{-4}$ \\
Gradient clip norm & 0.2 \\
\toprule
\end{tabular}
\label{supp: tab: hyperparameter}
\end{adjustwidth}
\end{table}

\section{VO Model Training and Inference Details}\label{supp: sec: vo train}

\subsection{Environment Details}\label{supp: sec: vo train: env}

Consistent with \cite{Savva2019HabitatAP}, in \tabref{supp: tab: gibson scene split}, we show the inventory of all scenes from Gibson~\cite{xiazamirhe2018gibsonenv} that were used in our experiments. Each of them is rated with quality level 4 or above as described in~\cite{Savva2019HabitatAP}. From the 72 scenes of the \texttt{train} split, we create a  training dataset $\mathcal{D}$ with one million data points as described in \secref{sec: experimental setup}. Similarly, a validation dataset $\mathcal{D}_{\text{val}}$ with 50,000 data points is generated from 14 scenes of the  \texttt{val} split.

\begin{table*}[!ht]
\renewcommand{\arraystretch}{0.9}
\begin{adjustwidth}{0.0cm}{}
\captionsetup{width=\textwidth}
\caption{Gibson-$4+$ scene split.}
\begin{tabularx}{\textwidth}{lL}
\toprule
Split & Scenes \\
\midrule
\texttt{Train} & Adrian, Applewold, Bolton, Cooperstown, Goffs, Hominy, Mobridge, Nuevo, Quantico, Roxboro, Silas, Stanleyville, Albertville, Arkansaw, Bowlus, Crandon, Hainesburg, Kerrtown, Monson, Oyens, Rancocas, Sanctuary, Sodaville, Stilwell, Anaheim, Avonia, Brevort, Delton, Hambleton, Maryhill, Mosinee, Parole, Reyno, Sasakwa, Soldier, Stokes, Andover, Azusa, Capistrano, Dryville, Haxtun, Mesic, Nemacolin, Pettigrew, Roane, Sawpit, Spencerville, Sumas, Angiola, Ballou, Colebrook, Dunmor, Hillsdale, Micanopy, Nicut, Placida, Roeville, Seward, Spotswood, Superior, Annawan, Beach, Convoy, Eagerville, Hometown, Mifflintown, Nimmons, Pleasant, Rosser, Shelbiana, Springhill, Woonsocket \\
\midrule
\texttt{Val} & Cantwell, Denmark, Eastville, Edgemere, Elmira, Eudora, Greigsville, Mosquito, Pablo, Ribera, Sands, Scioto, Sisters, Swormville \\
\toprule
\end{tabularx}
\label{supp: tab: gibson scene split}
\end{adjustwidth}
\end{table*}

\subsection{VO Dataset Statistics}\label{supp: sec: vo train: vo data stats}

\tabref{supp: tab: vo dataset stats} summarizes the statistics of our visual odometry (VO) training dataset $\mathcal{D}$. 
As mentioned in~\secref{sec: experimental setup}, since our training data is sampled from shortest-path trajectories, the ratio of  actions roughly represents the percentage of  actions that appeared in actual navigation tasks.

\tabref{supp: tab: vo dataset stats} provides another reason to use a separate model per action (SepAct) in a visual odometry model. Since the dataset is imbalanced with respect to the type of action, a unified model across all actions needs to deal with imbalanced training data. 
Empirically, we find that a unified model  overfits for \texttt{turn\_left} and \texttt{turn\_right} while the performance of \texttt{move\_forward} has not converged yet. The SepAct design overcomes this issue. More discussion is presented in \secref{supp: sec: vo train: vo eval}.

\begin{figure*}[t]
\centering
\begin{subfigure}{\textwidth}
\centering
    \includegraphics[width=\linewidth]{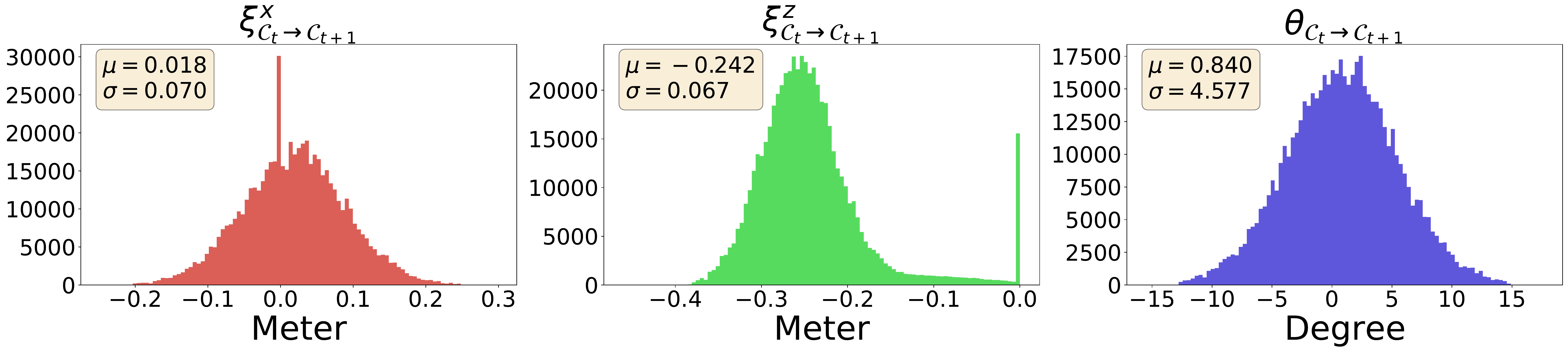}
    \captionsetup{width=0.9\linewidth}
    \caption{Action \texttt{move\_forward}.}
    \label{supp fig: vo act dist forward}
\end{subfigure}%
\hfill
\begin{subfigure}{\textwidth}
\centering
    \includegraphics[width=\linewidth]{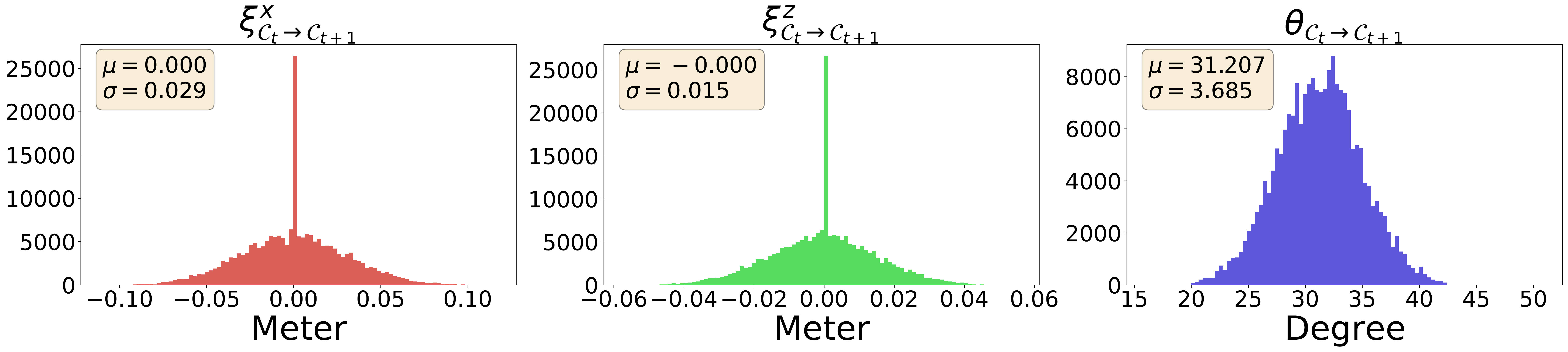}
    \captionsetup{width=0.9\linewidth}
    \caption{Action \texttt{turn\_left}.}
    \label{supp fig: vo act dist left}
\end{subfigure}%
\hfill
\begin{subfigure}{\textwidth}
\centering
    \includegraphics[width=\linewidth]{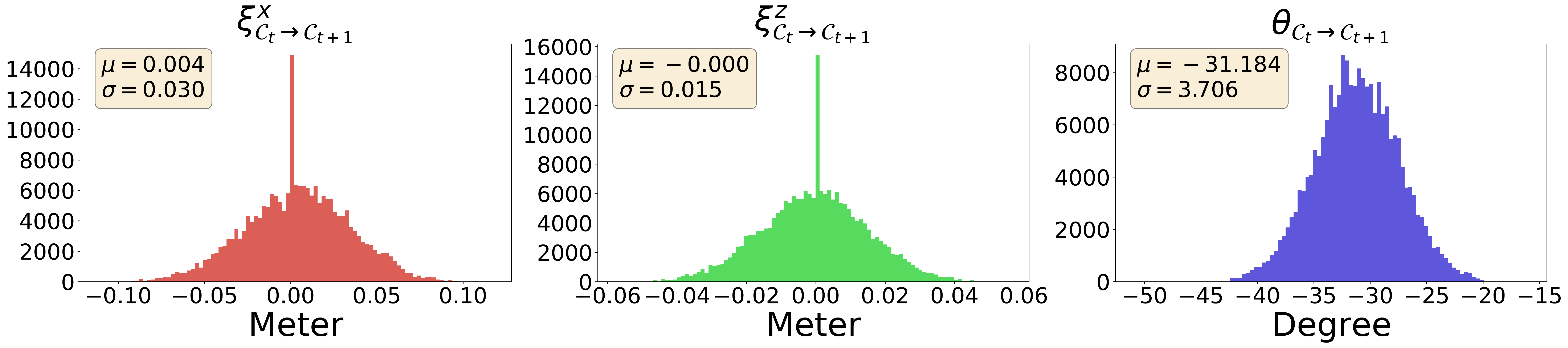}
    \captionsetup{width=0.9\linewidth}
    \caption{Action \texttt{turn\_right}.}
    \label{supp fig: vo act dist right}
\end{subfigure}%
\caption{Translation and rotation distribution histogram of each action in our VO training dataset. Because the simulator aligns the forward direction with the negative direction of the axis, most of the   $\xi^z_{\mathcal{C}_t \rightarrow \mathcal{C}_{t+1}}$ values for \texttt{move\_forward} are negative. 
}
\label{supp fig: vo act dist}
\end{figure*}

In \figref{supp fig: vo act dist}, we illustrate the distribution of translation and rotation caused by each action. We  note that for each of the actions, the distribution of the  translation changes has a peak around $0m$, which is caused by the agent getting stuck when encountering collisions.

\begin{table*}[t]
\renewcommand{\arraystretch}{1.0}
\begin{adjustwidth}{0.0cm}{}
\captionsetup{width=\textwidth}
\caption{Visual odometry training dataset statistics.}
{\small
\begin{tabularx}{\textwidth}{l|RRR|R}
\toprule
\backslashbox{Category}{Action} & \texttt{move\_forward} & \texttt{turn\_left} & \texttt{turn\_right} & Total \\
\midrule
Non-collided & 503,890 (87.90\%) & 186,291 (87.32\%) & 197,318 (92.49\%) & 887,499 (88.75\%) \\
Collided     & 69,342 (12.10\%)  & 27,143 (12.68\%)  & 16,016 (7.51\%)   & 112,501 (11.25\%) \\
\midrule
Total        & 573,232 & 213,434 & 213,334 & 1,000,000 \\
\toprule
\end{tabularx}
}
\label{supp: tab: vo dataset stats}
\end{adjustwidth}
\end{table*}

\subsection{Qualitative Examples from $\mathcal{D}$}

\begin{figure*}[t]
\centering
\begin{subfigure}{0.49\textwidth}
\centering
    \includegraphics[width=\linewidth]{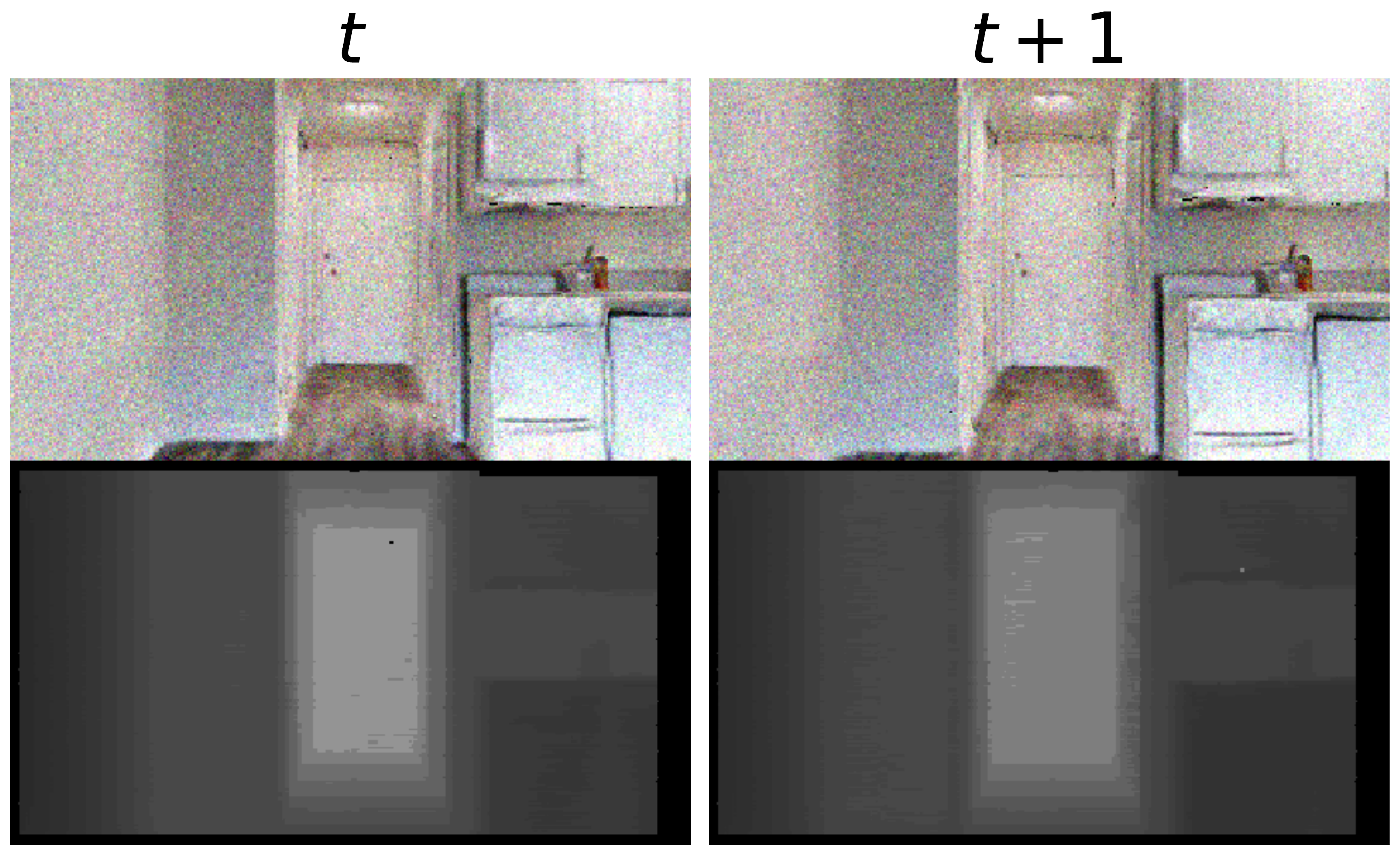}
    \captionsetup{width=\linewidth}
    \caption{\texttt{move\_forward}, no collision, $(0.05, 0.19, -0.26)$.}
    \label{supp fig: vo qualitative forward no collision}
\end{subfigure}%
\hfill
\begin{subfigure}{0.49\textwidth}
\centering
    \includegraphics[width=\linewidth]{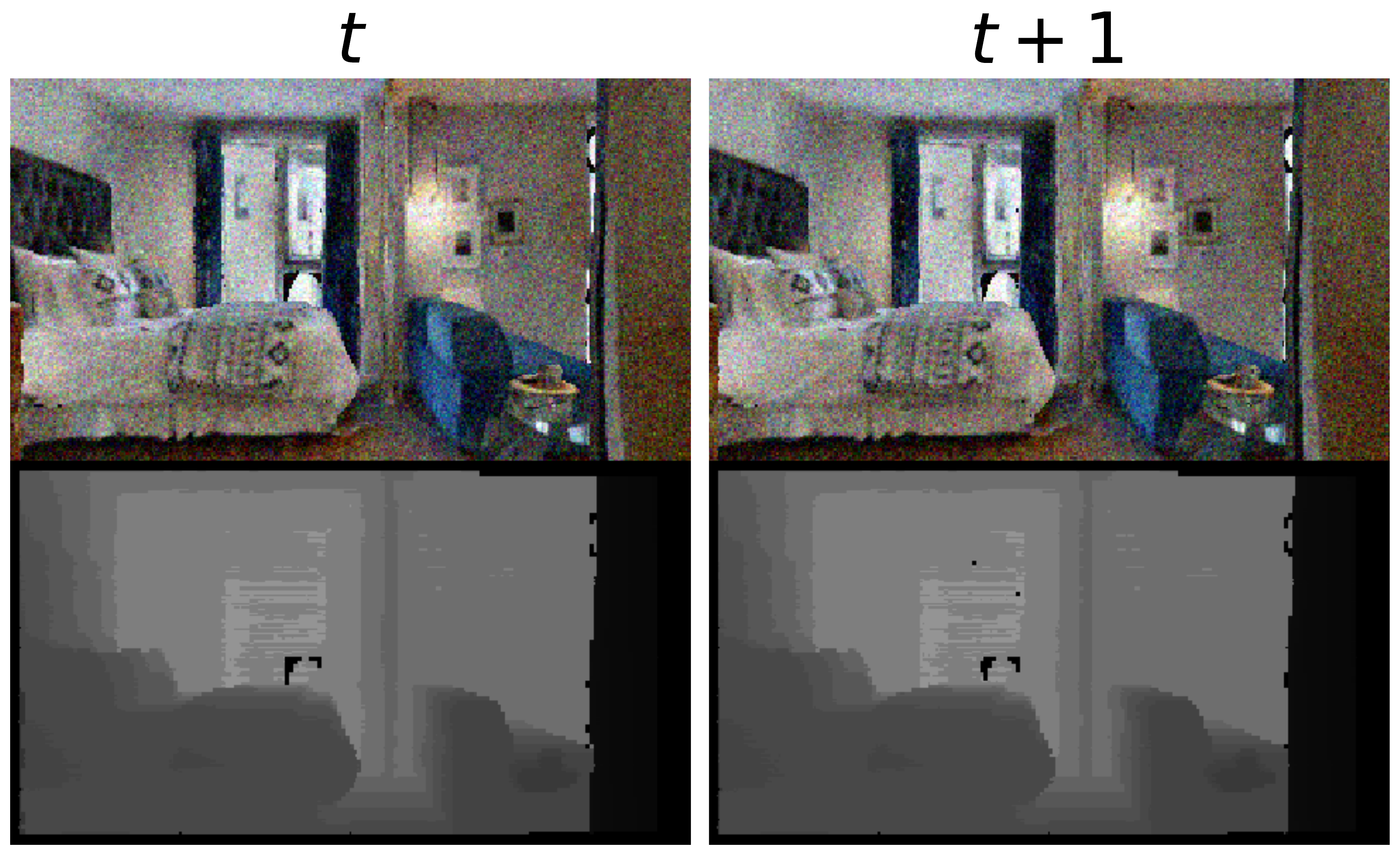}
    \captionsetup{width=\linewidth}
    \caption{\texttt{move\_forward}, with collision, $(0.00, 0.00, -0.20)$.}
    \label{supp fig: vo qualitative forward collision}
\end{subfigure}%
\hfill
\centering
\begin{subfigure}{0.49\textwidth}
\centering
    \includegraphics[width=\linewidth]{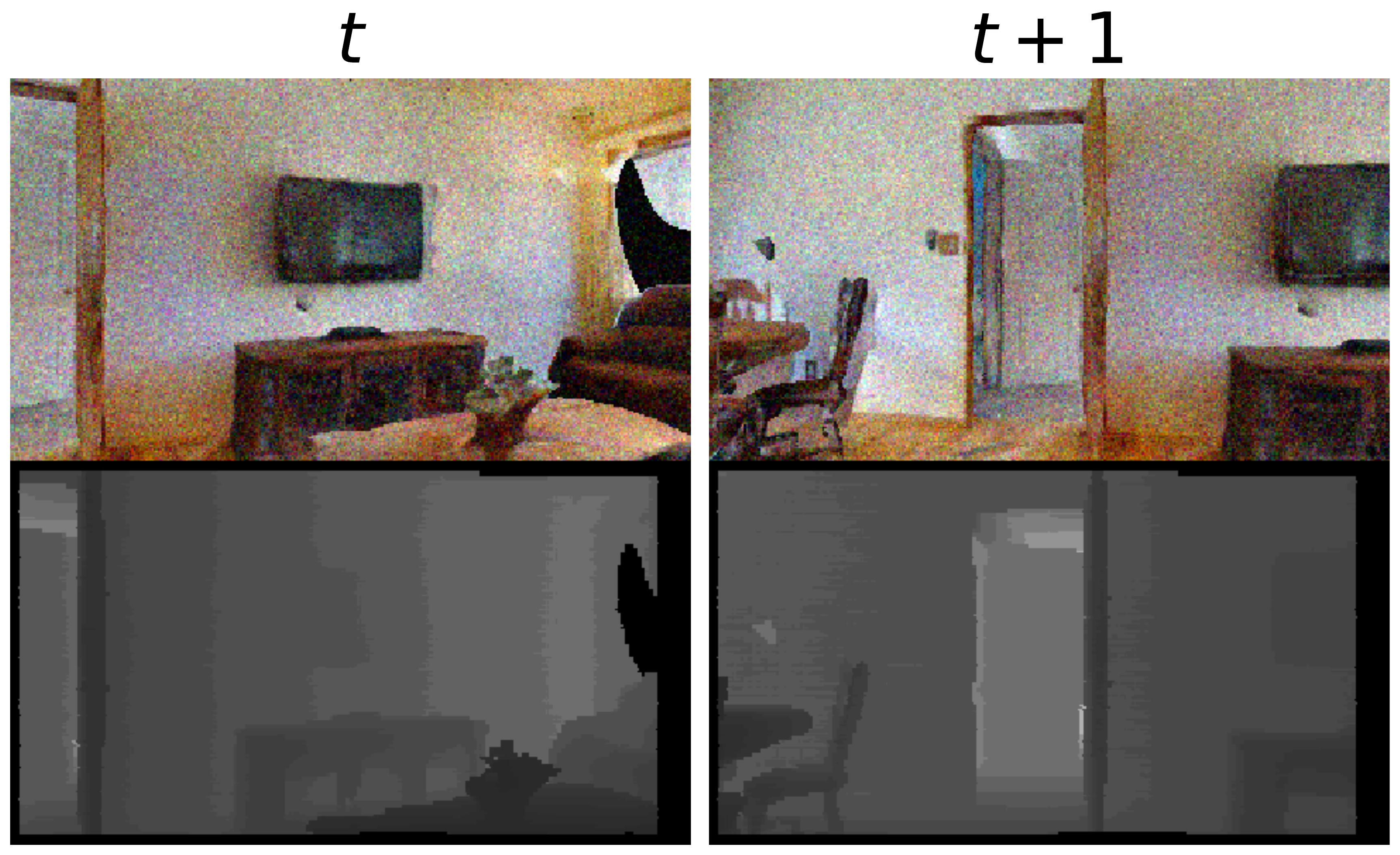}
    \captionsetup{width=\linewidth}
    \caption{\texttt{turn\_left}, no collision, $(-0.04, -0.01, 32.8)$.}
    \label{supp fig: vo qualitative left no collision}
\end{subfigure}%
\hfill
\begin{subfigure}{0.49\textwidth}
\centering
    \includegraphics[width=\linewidth]{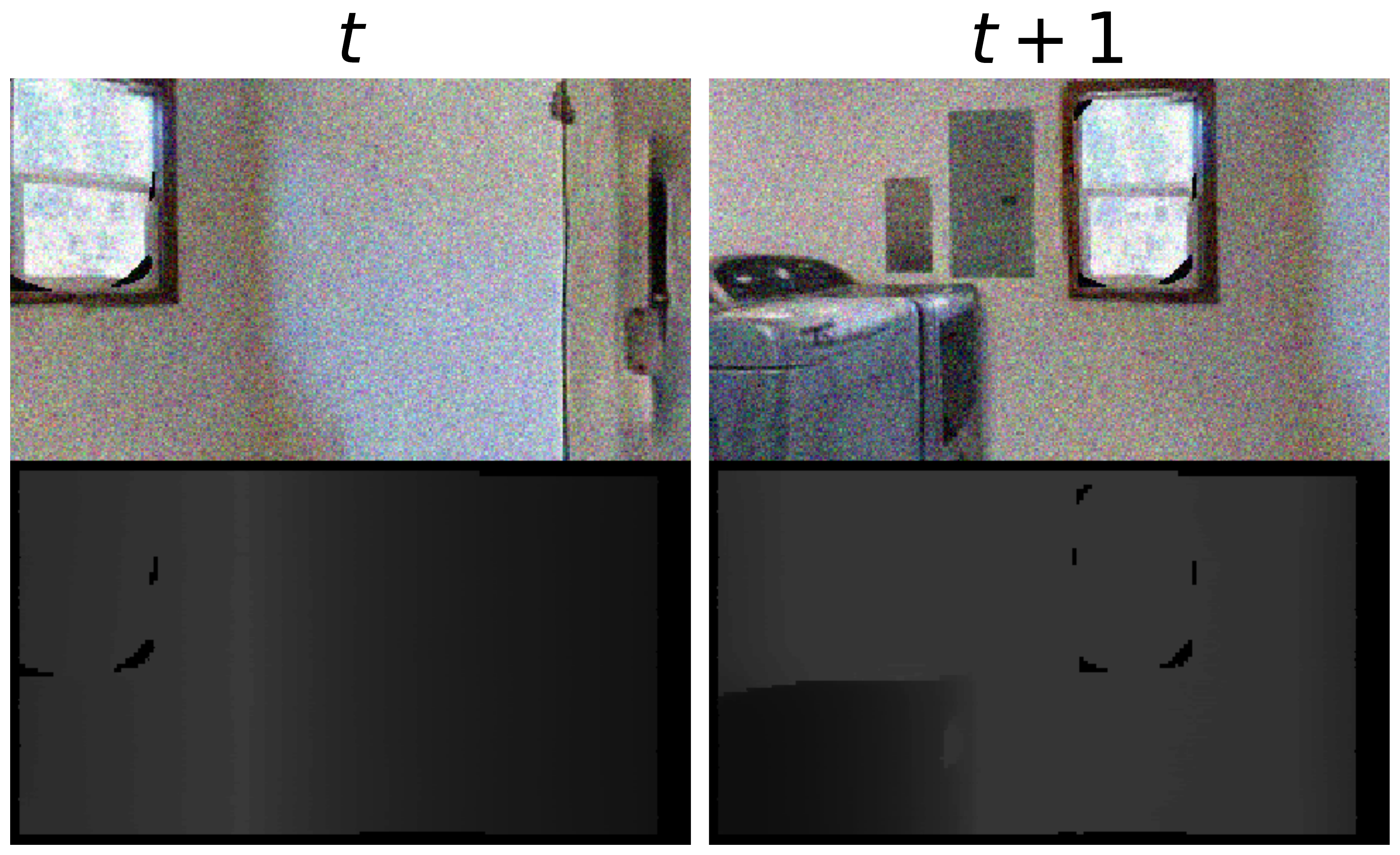}
    \captionsetup{width=\linewidth}
    \caption{\texttt{turn\_left}, with collision, $(0.00, 0.00, 38.6)$.}
    \label{supp fig: vo qualitative left collision}
\end{subfigure}%
\hfill
\centering
\begin{subfigure}{0.49\textwidth}
\centering
    \includegraphics[width=\linewidth]{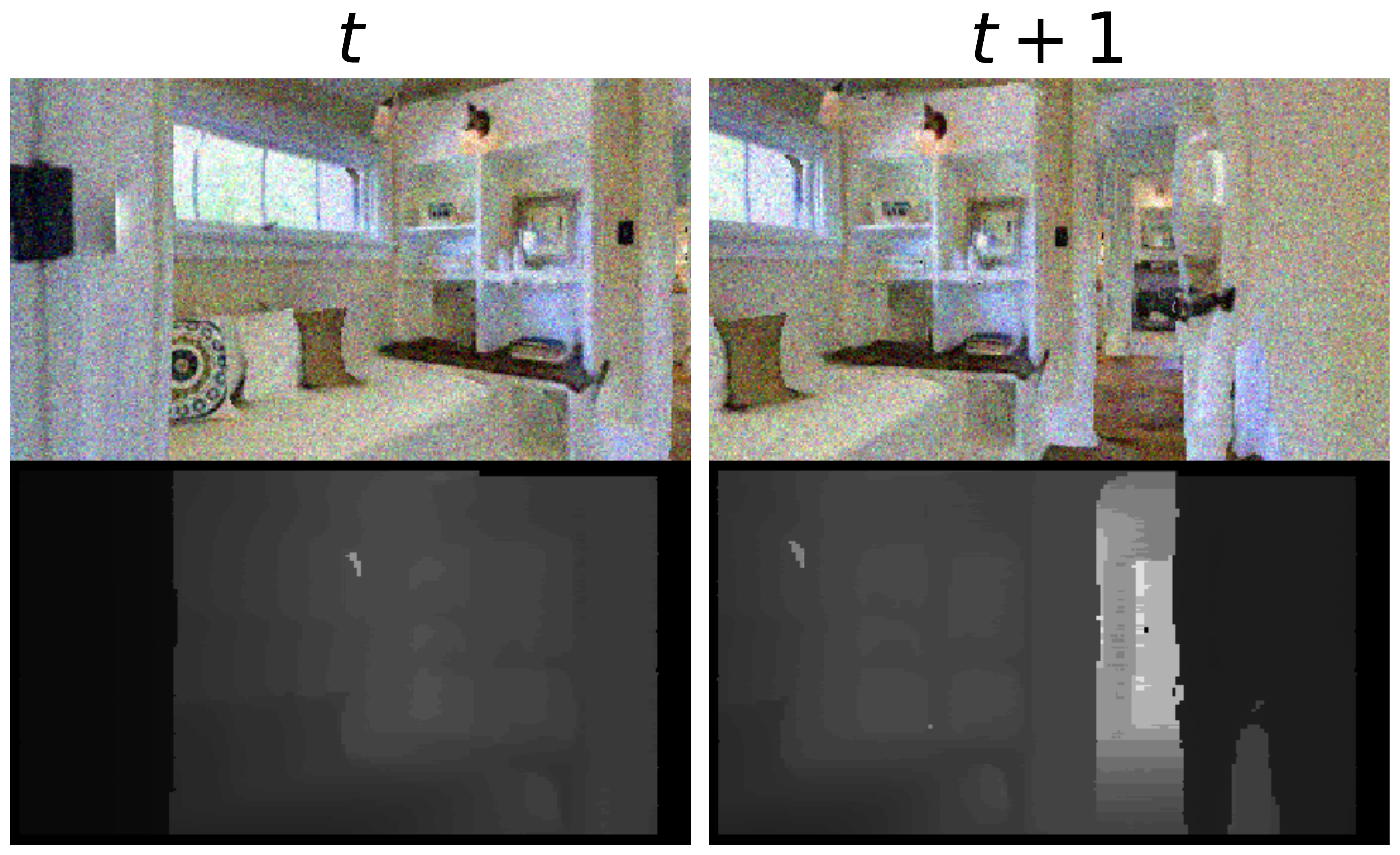}
    \captionsetup{width=\linewidth}
    \caption{\texttt{turn\_right}, no collision, $(-0.02, 0.00, -28.4)$.}
    \label{supp fig: vo qualitative right no collision}
\end{subfigure}%
\hfill
\begin{subfigure}{0.49\textwidth}
\centering
    \includegraphics[width=\linewidth]{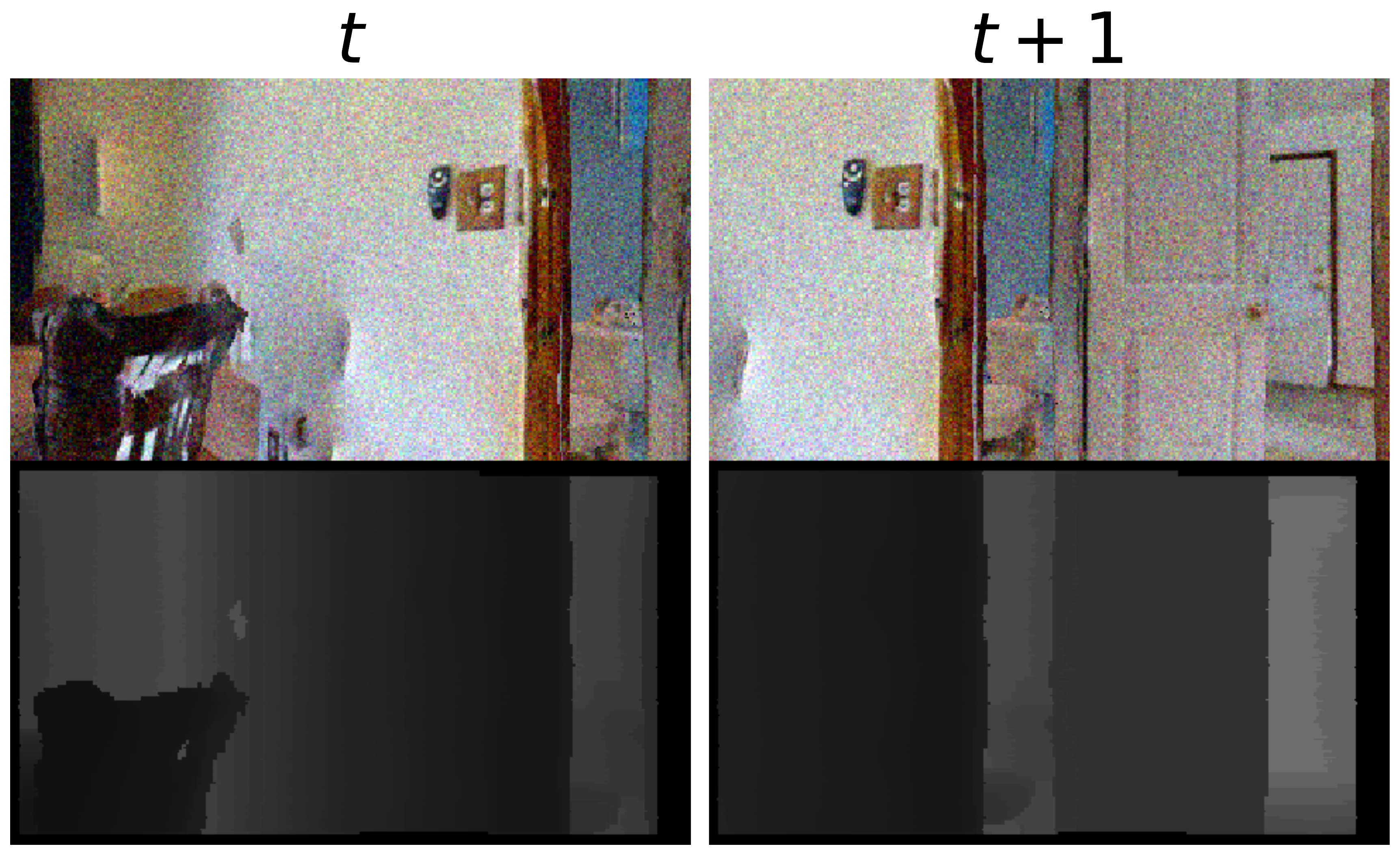}
    \captionsetup{width=\linewidth}
    \caption{\texttt{turn\_right}, with collision, $(0.06, 0.00, -29.4)$.}
    \label{supp fig: vo qualitative right collision}
\end{subfigure}%
\vspace{0.1cm}
\caption{Qualitative examples of translation and rotation changes caused by each action. The changes are presented in the order of $(\xi^x_{\mathcal{C}_t \rightarrow \mathcal{C}_{t+1}}, \xi^z_{\mathcal{C}_t \rightarrow \mathcal{C}_{t+1}}, \theta_{\mathcal{C}_t \rightarrow \mathcal{C}_{t+1}})$.
}
\label{supp fig: vo qualitative}
\end{figure*}

\figref{supp fig: vo qualitative} shows qualitative examples of translation and rotation changes resulting from each action.
Apart from the noisy egocentric observations, the complexity of estimating the $SE(2)$ transformation also stems from similar translation and rotation changes across different actions. For example, $\xi^x_{\mathcal{C}_t \rightarrow \mathcal{C}_{t+1}}$ in all six figures is extremely similar.

\subsection{VO Model Evaluation}\label{supp: sec: vo train: vo eval}

\begin{figure*}[t]
\centering
\begin{subfigure}{\textwidth}
\centering
    \includegraphics[width=\linewidth]{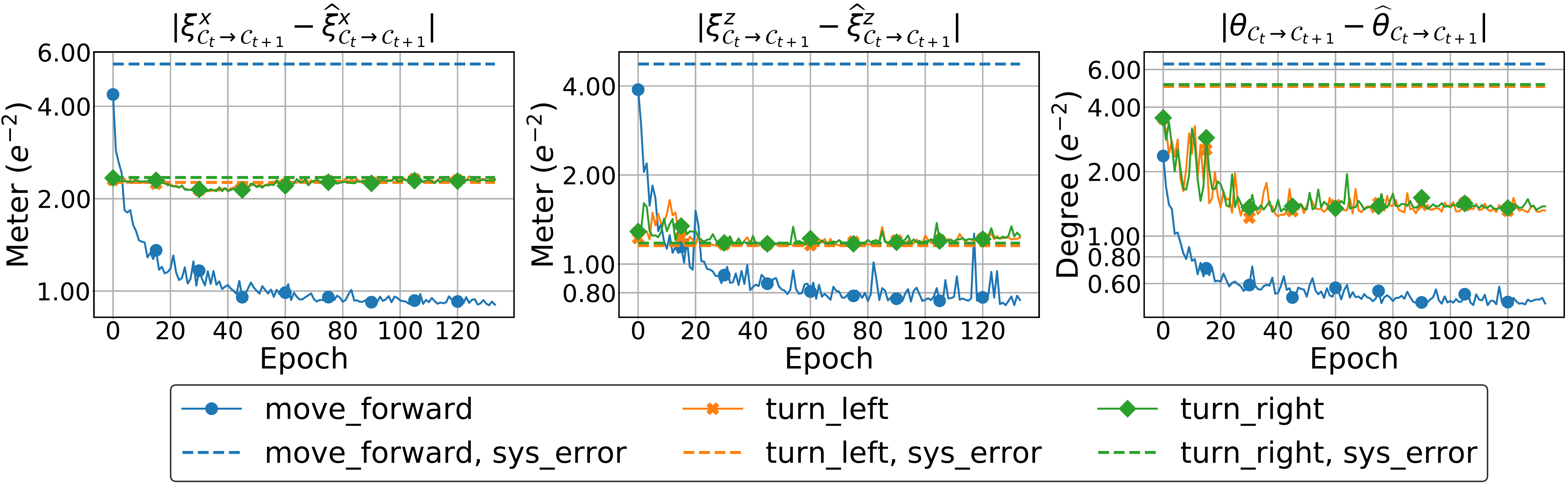}
    \captionsetup{width=0.9\linewidth}
    \caption{Unified model.}
    \label{supp fig: vo eval unified dropout}
\end{subfigure}%
\hfill
\begin{subfigure}{\textwidth}
\centering
    \includegraphics[width=\linewidth]{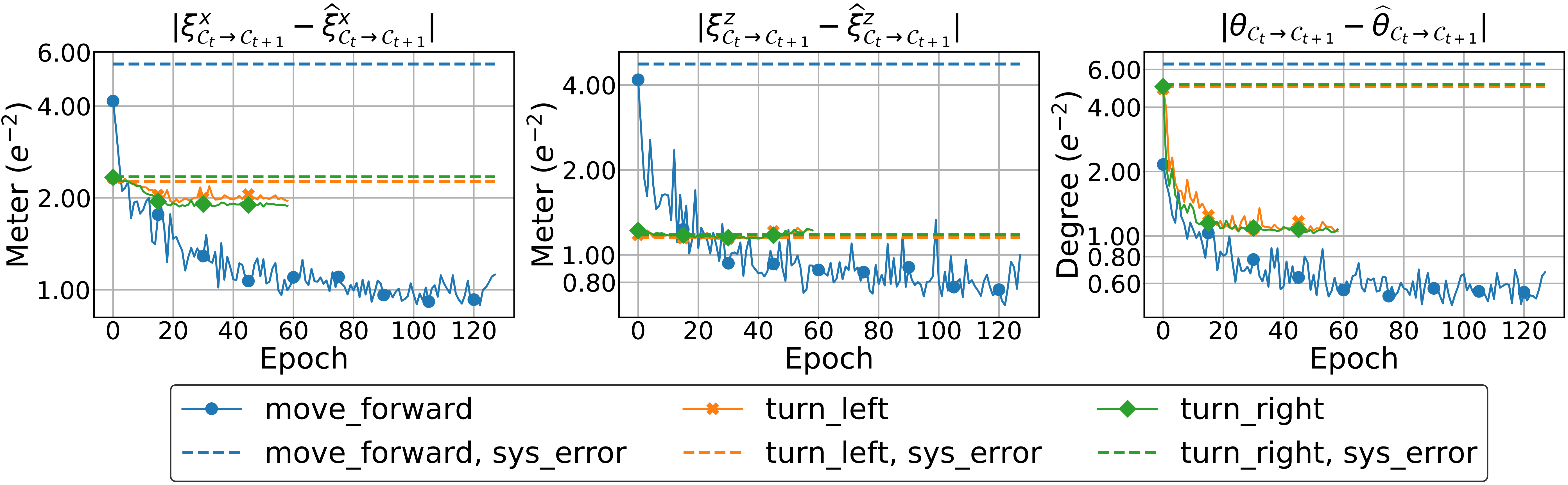}
    \captionsetup{width=0.9\linewidth}
    \caption{SepAct model.}
    \label{supp fig: vo eval sep act dropout}
\end{subfigure}%
\vspace{0.1cm}
\caption{Evaluation of VO models on generated validation dataset $\mathcal{D}_{\text{val}}$. We show the average absolute difference between ground-truth value and prediction from VO models. The $y$-axis uses $\log$-scale. The \texttt{sys\_error} is defined in \secref{supp: sec: vo train: vo eval}.}
\label{supp fig: vo eval}
\end{figure*}

\figref{supp fig: vo eval} shows the evaluation curve on $\mathcal{D}_{\text{val}}$ for the \textit{Unified} and \textit{SepAct} models, namely the VO model of Row 6 and Row 8 in \tabref{tab: offline eval combine}. We define \texttt{sys\_error} as the average absolute difference between ground-truth and estimated values if the VO model always predicts the mean of the training data in \figref{supp fig: vo act dist}. For example, if we let $\mathcal{D}^{\text{forward}} \subset \mathcal{D}$ and $\mathcal{D}_{\text{val}}^{\text{forward}} \subset \mathcal{D}_{\text{val}}$ be datasets whose  data points are generated by the  \texttt{move\_forward} action, we compute the \texttt{sys\_error} of \texttt{move\_forward} on $\xi^x_{\mathcal{C}_{t} \rightarrow \mathcal{C}_{t+1}}$  as:
\begin{align}
    \texttt{sys\_error} &= \frac{1}{\vert \mathcal{D}_{\text{val}}^{\text{forward}} \vert} \sum\limits_{d_{\mathcal{C}_t \rightarrow \mathcal{C}_{t+1}} \in \mathcal{D}_{\text{val}}^{\text{forward}}} \vert \xi^x_{\mathcal{C}_{t} \rightarrow \mathcal{C}_{t+1}} - \mu \vert, \nonumber \\
    \;\text{where}\; & \mu = \frac{1}{\vert \mathcal{D}^{\text{forward}} \vert} \sum\limits_{d_{\mathcal{C}_t \rightarrow \mathcal{C}_{t+1}} \in \mathcal{D}^{\text{forward}}} \xi^x_{\mathcal{C}_{t} \rightarrow \mathcal{C}_{t+1}}.
\end{align}
Here $d_{\mathcal{C}_t \rightarrow \mathcal{C}_{t+1}} = \left( (I_t, I_{t+1}), \bm{\xi}_{\mathcal{C}_t \rightarrow \mathcal{C}_{t+1}}, \theta_{\mathcal{C}_t \rightarrow \mathcal{C}_{t+1}} \right)$ and $\mu = 0.018$ from the first histogram of \figref{supp fig: vo act dist forward}. Note, \texttt{sys\_error} is computed equivalently for $\xi^x_{\mathcal{C}_t \rightarrow \mathcal{C}_{t+1}}$, $\xi^z_{\mathcal{C}_t \rightarrow \mathcal{C}_{t+1}}$, and $\theta_{\mathcal{C}_t \rightarrow \mathcal{C}_{t+1}}$ of all three actions. The closer the evaluation curve is to \texttt{sys\_error}, the less useful the information that the VO model learns. Apparently, the SepAct model learns more helpful information as its curve is further away from the \texttt{sys\_error} line. 

Meanwhile, as discussed in \secref{supp: sec: vo train: vo data stats},
the training dataset, which represents the actual path's action distribution, is imbalanced. A unified model may encounter overfitting on one action while yielding unsatisfactory prediction on another. 
Specifically, in the first plot of \figref{supp fig: vo eval unified dropout}, the performances on \texttt{turn\_left} and \texttt{turn\_right} encounters overfitting at around the $30^\text{th}$ epoch, while the performance on \texttt{move\_forward} improves even at the $120^\text{th}$ epoch. This issue does not arise in SepAct's evaluation curve in \figref{supp fig: vo eval sep act dropout}, verifying the efficacy of SepAct. 

\section{DeepVO and KITTI}\label{supp: sec: kitti}

\begin{table*}[h]
\renewcommand{\arraystretch}{1.1}
\begin{adjustwidth}{0.0cm}{}
\captionsetup{width=0.9\textwidth}
\caption{Results on KITTI. Values are $r_\text{rel}(^\circ)/t_\text{rel}(\%)$.}
\vspace{-0.4cm}
\renewcommand\theadfont{}
\centering
{%
\begin{tabular}{crrrr} 
\toprule
&  1.DeepVO$^\dagger$ & 2.DeepVO$^\ddagger$ & 3.DeepVO$^\mathsection$ & 4.RGB-D-DD-S-Proj \\
\midrule
Seq09 & N/A & 33.37 / 92.97 & 4.016 / 11.14 & 7.062 / 19.22 \\
Seq10 & 8.83 / 8.11 & 38.68 / 90.22 & 4.498 / 11.24 & 9.298 / 15.80 \\
\toprule
\end{tabular}
}
\label{supp: tab: rebuttal kitti}
\end{adjustwidth}
\end{table*}

In this section we discuss  implementation details of DeepVO as well as our model's performance on KITTI.

\noindent\textbf{DeepVO implementation.} There isn't an official code of DeepVO and the most-starred public one yields incorrect results (\tabref{supp: tab: rebuttal kitti}'s Col.~2)\footnote{\url{https://github.com/ChiWeiHsiao/DeepVO-pytorch}}. 
Our re-implementation of DeepVO (Col.~3 in~\tabref{supp: tab: rebuttal kitti}) matches the  numbers reported in the original DeepVO paper (Col.~1)\footnote{Differences are due to the rare train/test split in the DeepVO paper while we train on Seq00-08 and evaluate on Seq09/10 as Tab.~1 in~\cite{Chen2020ASO}.}.
Therefore, we apply our implemented DeepVO in the PointGoal navigation task.

\noindent\textbf{Our VO module on KITTI~\cite{Geiger2012AreWR}.} In order to run our VO module on KITTI, we need depth information. We use one of the best entries (DeepPruner~\cite{Duggal2019DeepPrunerLE}) in Tab.~3 from~\cite{Laga2020ASO} to obtain a depth estimate.
As can be inferred from \tabref{supp: tab: rebuttal kitti}'s Col.~3 \vs 4 and \tabref{tab: offline eval combine}'s Row 0 \vs Row 18, outdoor and indoor tasks have their own challenges.

\section{Estimate Relative Pose from Depth}\label{supp: sec: vo from depth}

Because depth is noisy as mentioned in~\secref{sec: approach}, it prevents reliable estimation of relative pose. To verify, we experiment with the following pipeline.

\begin{figure}[t!]
\centering
\captionsetup[subfigure]{aboveskip=1pt,belowskip=1pt}
\begin{subfigure}{0.93\linewidth}
\centering
    \includegraphics[width=\linewidth]{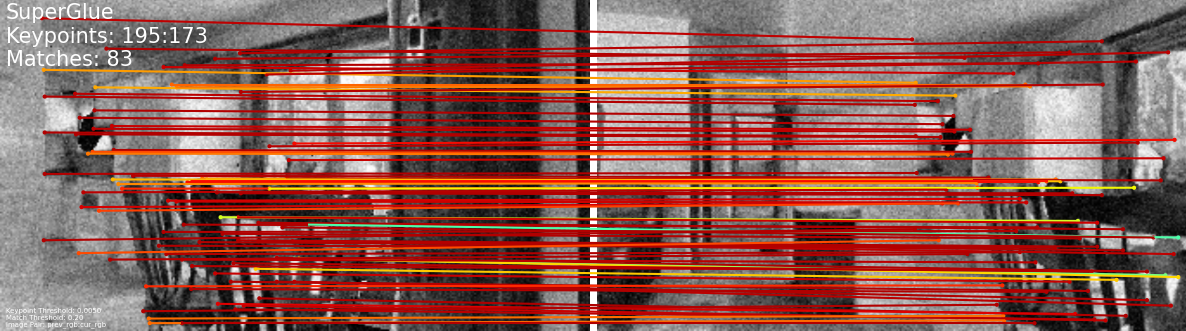}
    \captionsetup{width=\linewidth}
    \caption{\label{fig: feat match spsg} Matched points from SPSG.}
    
\end{subfigure}%
\hspace{0.0\linewidth}
\begin{subfigure}{0.95\linewidth}
\centering
    \includegraphics[width=\linewidth]{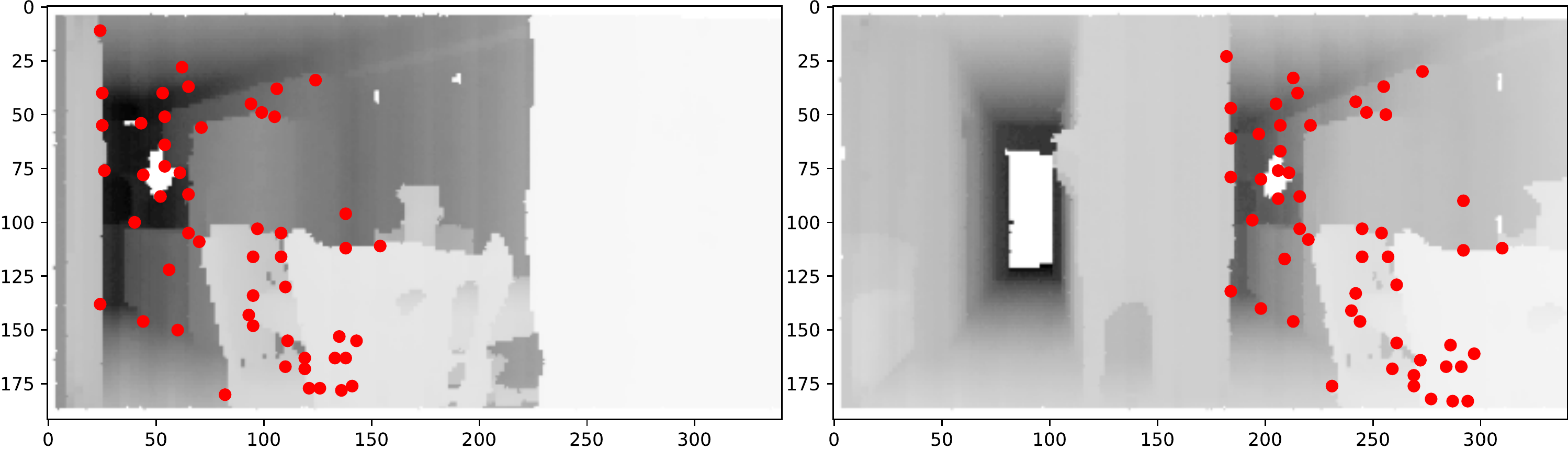}
    \captionsetup{width=\linewidth}
    \caption{Inliers from \texttt{OpenCV}.}
    \label{fig: feat match inlier}
\end{subfigure}%
\vspace{-0.3cm}
\caption{\label{fig: feat match} Qualitative examples for feature matching.}
\end{figure}

\noindent\textbf{1) Find matching points.} To extract and match point descriptors in adjacent RGB frames,  we use the recent SuperPoint-SuperGlue (SPSG)~\cite{DeTone2018SuperPointSI,Sarlin2020SuperGlueLF} which was shown to improve over traditional hand-engineered methods. Qualitatively,~\figref{fig: feat match spsg} verifies high-quality matches.

\noindent\textbf{2) Compute relative pose.} We use \texttt{findEssentialMat} and \texttt{recoverPose} from \texttt{OpenCV}  to recover rotation $\widehat{\theta}_{\mathcal{C}_{t} \rightarrow \mathcal{C}_{t+1}}$ and \textit{direction} of translation.~\figref{fig: feat match inlier} shows  inliers for~\figref{fig: feat match spsg} found by \texttt{OpenCV}. High-quality inliers ease the analysis %
as the final VO prediction error unlikely stems from mismatched points.

\noindent\textbf{3) Resolve scale ambiguity.}
\textbf{3).a} With depth, we compute 3D coordinates of inliers in two camera coordinate systems.
\textbf{3).b} We rotate 3D coordinates in one frame with $\widehat{\theta}_{\mathcal{C}_{t} \rightarrow \mathcal{C}_{t+1}}$.
\textbf{3).c} We compute the scale as the averaged norm between the rotated coordinates and the coordinates in the other frame. To obtain the final translation $(\hat{\xi}_{\mathcal{C}_{t} \rightarrow \mathcal{C}_{t+1}}^x, \hat{\xi}_{\mathcal{C}_{t} \rightarrow \mathcal{C}_{t+1}}^z)$, we rescale the \textit{direction} produced by \texttt{OpenCV}. The obtained VO  error (\texttt{E1}) is much larger than ours (\texttt{E3}) (\tabref{tab: supp: vo err on val}) and prevents successful navigation.

\noindent\textbf{4) Additional oracle experiment.} We conduct an oracle experiment  using \textit{ground-truth} rotation $\theta_{\mathcal{C}_{t} \rightarrow \mathcal{C}_{t+1}}$ in \textbf{3).b}.
From \texttt{E2}~\vs~\texttt{E3} (ours):  directly estimating relative pose from depth is inferior. Note, the validation set scenes
are not used for training our VO model (\secref{sec: experimental setup}).

\begin{table}[t]
\renewcommand{\arraystretch}{0.8}
\centering
\newcommand{\tmpDel}{0.4em}
{
\begin{tabular}{c @{\hskip \tmpDel}|@{\hskip \tmpDel} c @{\hskip \tmpDel}|@{\hskip \tmpDel} c} 
\toprule
 \texttt{E1} ($e^{-2}$) & \texttt{E2} ($e^{-2}$) & \texttt{E3} ($e^{-2}$)  \\
\midrule
(15.9, 21.3, 8.51) & (3.97, 10.6, 0.00) & (1.22, 0.86, 0.66) \\
\toprule
\end{tabular}
}
\vspace{-0.3cm}
\caption{
\textbf{VO prediction error on $\mathcal{D}_\text{val}$} (50000 entries, see~\secref{supp: sec: vo train: env}).
Lower is better.
Following~\tabref{tab: offline eval combine}, we report %
$(\hat{\xi}_{\mathcal{C}_{t} \rightarrow \mathcal{C}_{t+1}}^x, \hat{\xi}_{\mathcal{C}_{t} \rightarrow \mathcal{C}_{t+1}}^z, \widehat{\theta}_{\mathcal{C}_{t} \rightarrow \mathcal{C}_{t+1}})$.
\texttt{E1}: Feature Matching;
\texttt{E2}: Feature Matching Oracle;
\texttt{E3}: our result.
}
\label{tab: supp: vo err on val}
\vspace{-0.3cm}
\end{table}

\section{More Qualitative Results}\label{supp: sec: more qualitative}

\begin{figure*}[t]
\centering
\begin{subfigure}{0.25\textwidth}
\centering
    \includegraphics[width=\linewidth]{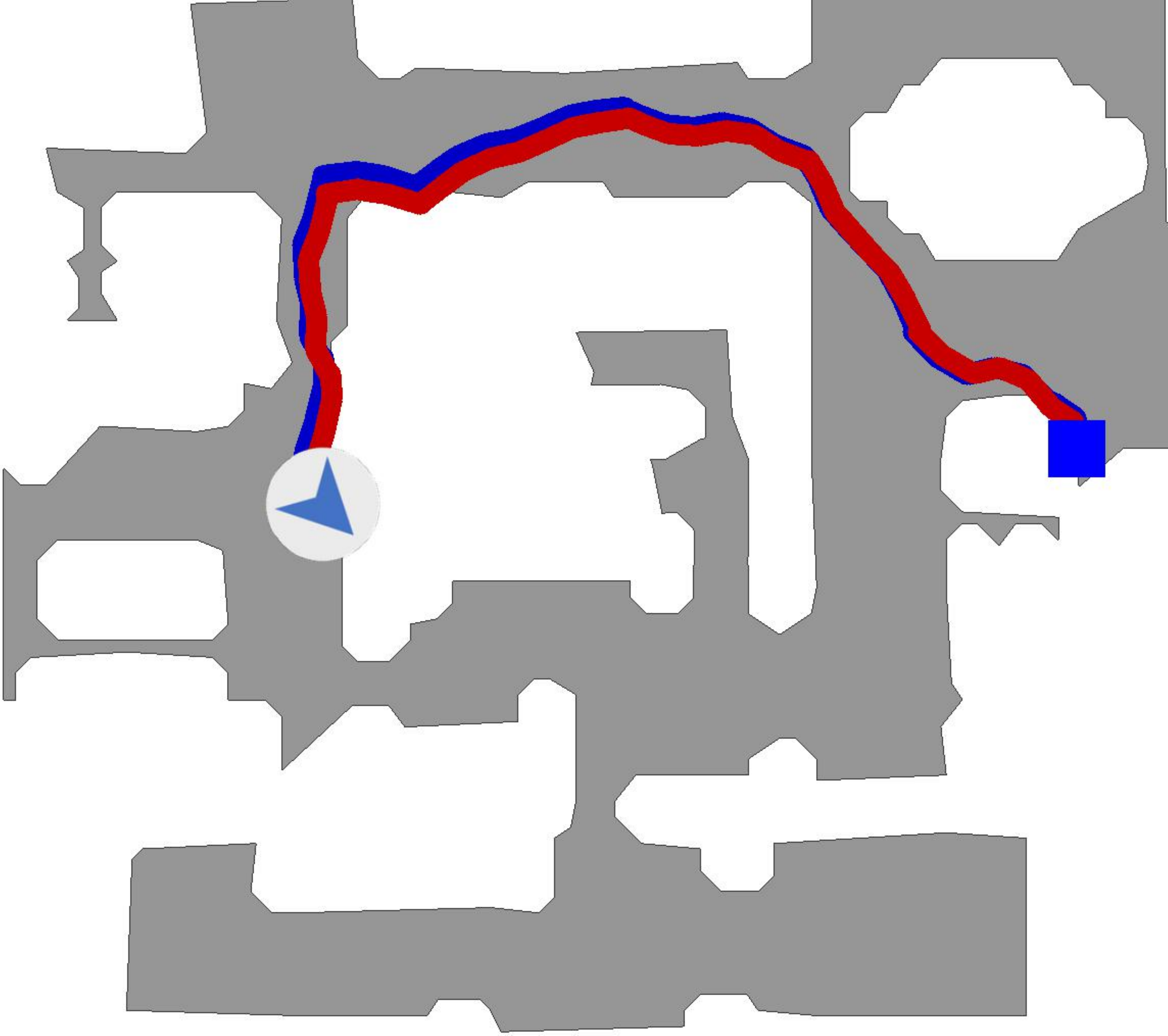}
    \captionsetup{width=0.9\linewidth}
    \caption{Scioto, SPL $87\%$.}
    \label{supp fig: nav qualitative, Scioto}
\end{subfigure}%
\hfill
\begin{subfigure}{0.3\textwidth}
\centering
    \includegraphics[width=\linewidth]{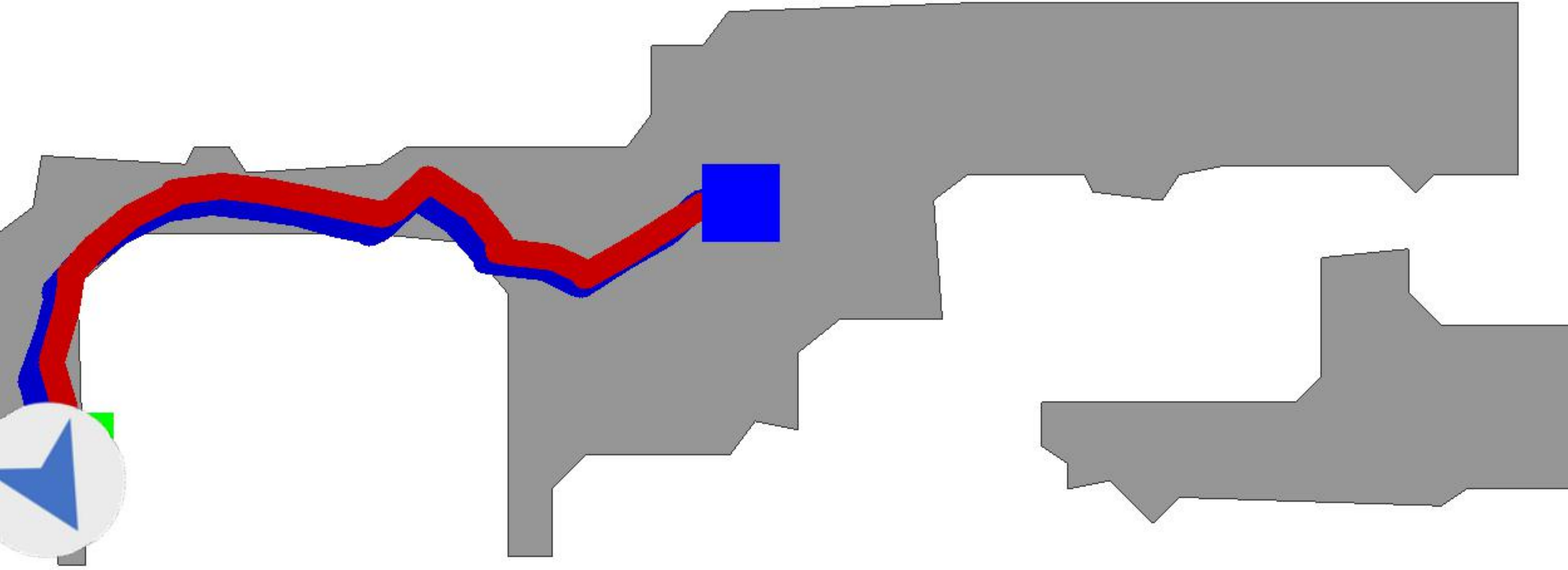}
    \captionsetup{width=0.9\linewidth}
    \caption{Pablo, SPL $84\%$.}
    \label{supp fig: nav qualitative, Pablo}
\end{subfigure}%
\hfill
\begin{subfigure}{0.4\textwidth}
\centering
    \includegraphics[width=\linewidth]{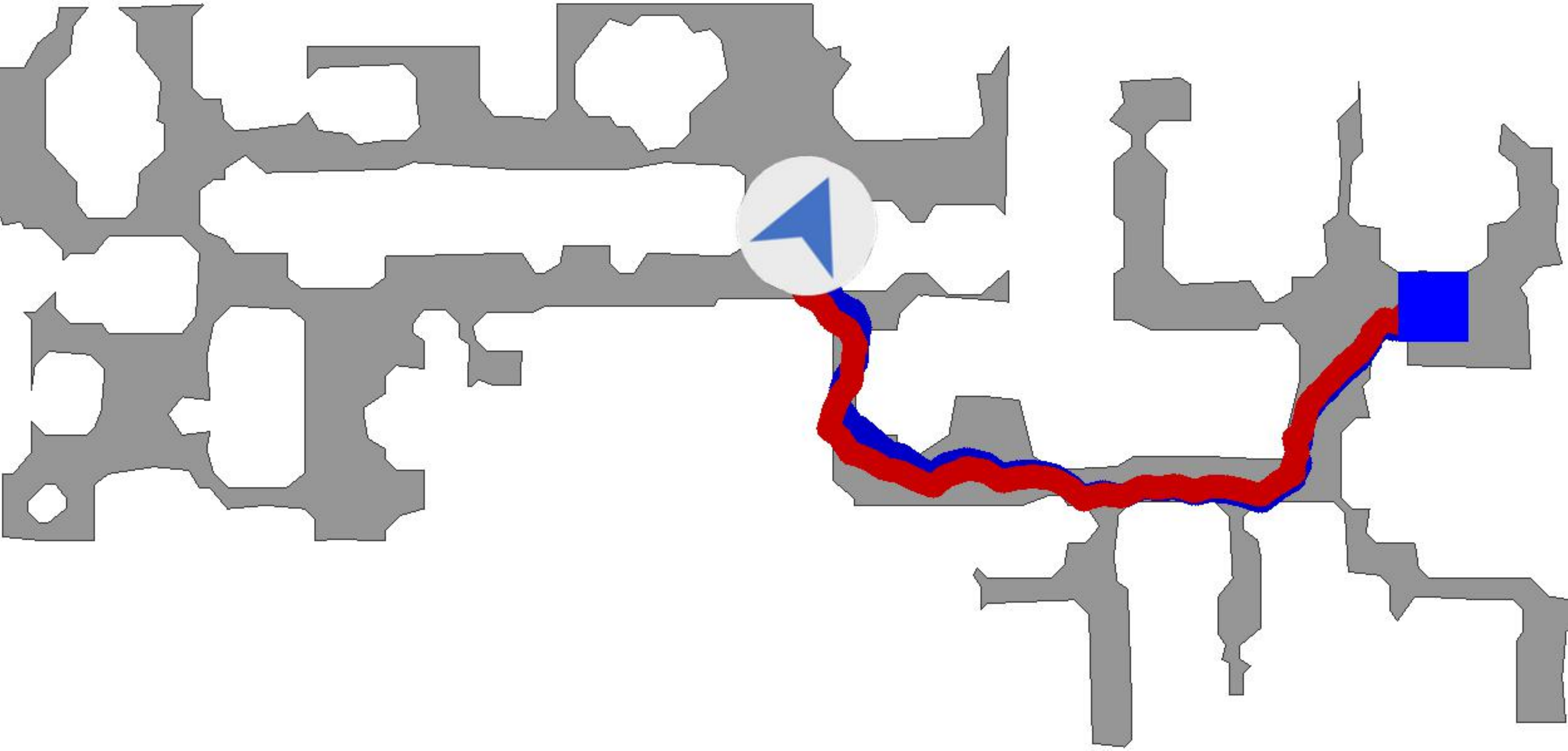}
    \captionsetup{width=0.9\linewidth}
    \caption{Mosquito, SPL $83\%$.}
    \label{supp fig: nav qualitative, Mosquito}
\end{subfigure}%
\hfill
\begin{subfigure}{0.3\textwidth}
\centering
    \includegraphics[width=\linewidth]{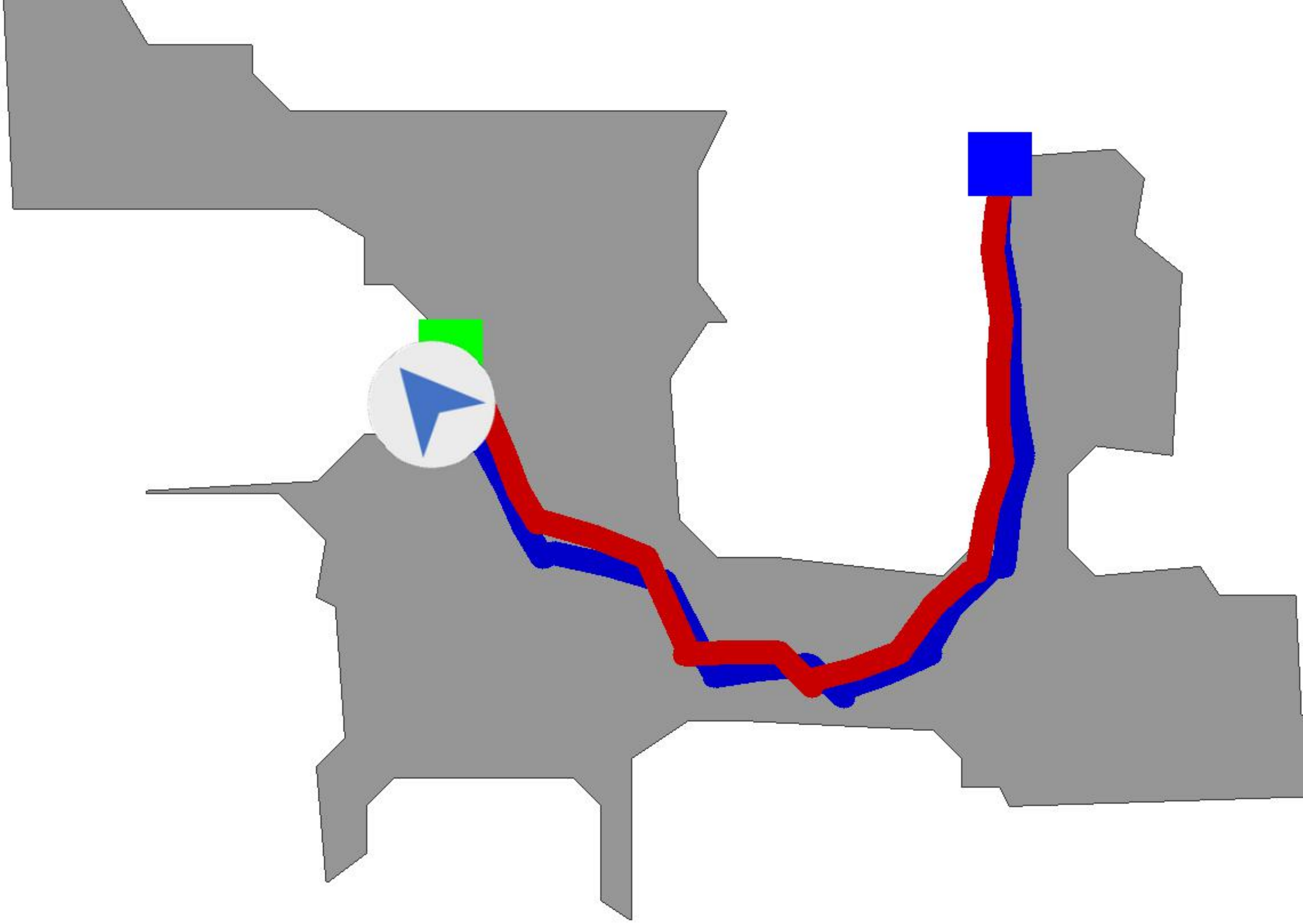}
    \captionsetup{width=0.9\linewidth}
    \caption{Denmark, SPL $82\%$.}
    \label{supp fig: nav qualitative, Denmark}
\end{subfigure}%
\hfill
\begin{subfigure}{0.3\textwidth}
\centering
    \includegraphics[width=\linewidth]{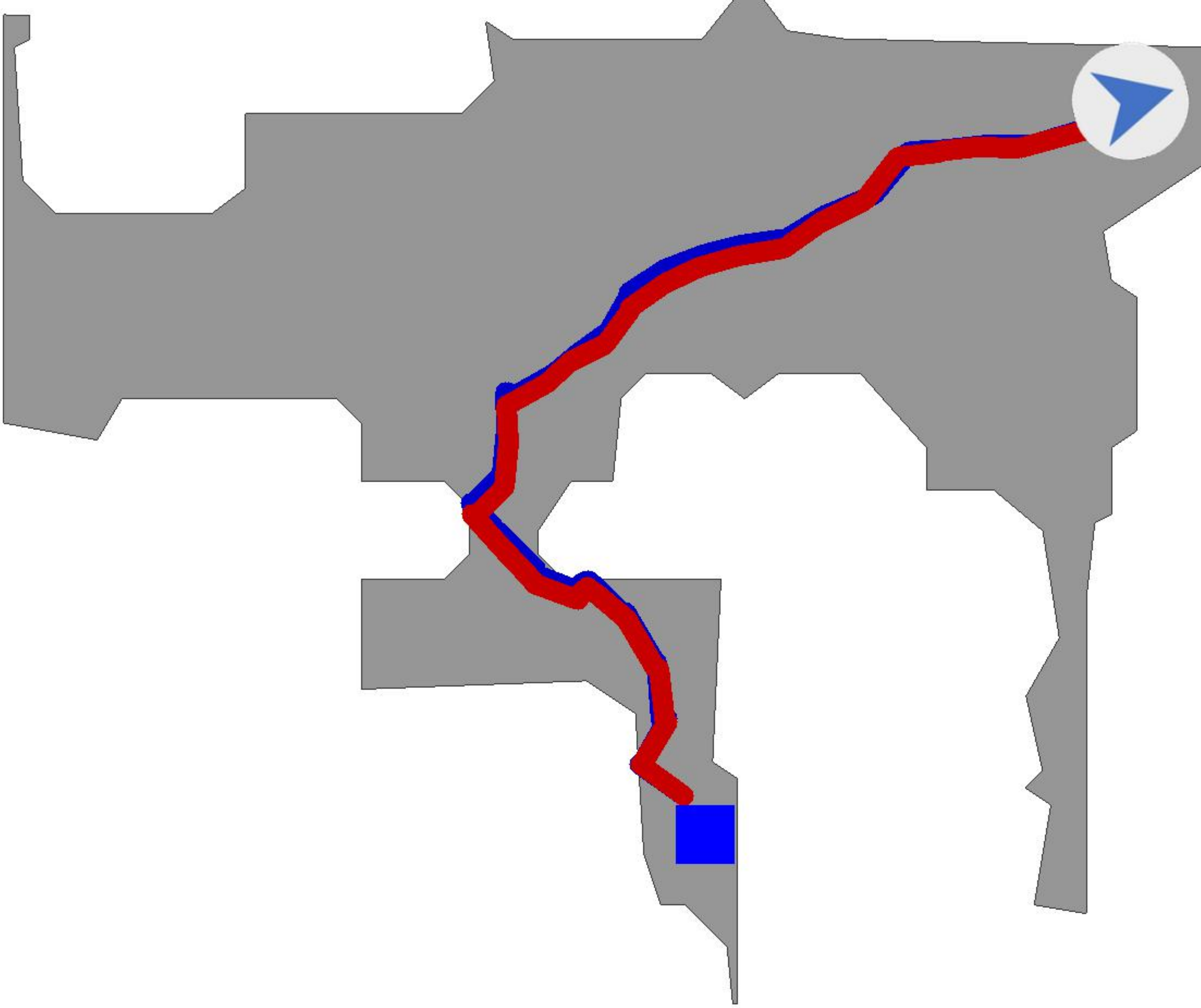}
    \captionsetup{width=0.9\linewidth}
    \caption{Greigsville, SPL $81\%$.}
    \label{supp fig: nav qualitative, Greigsville}
\end{subfigure}%
\hfill
\begin{subfigure}{0.3\textwidth}
\centering
    \includegraphics[width=\linewidth]{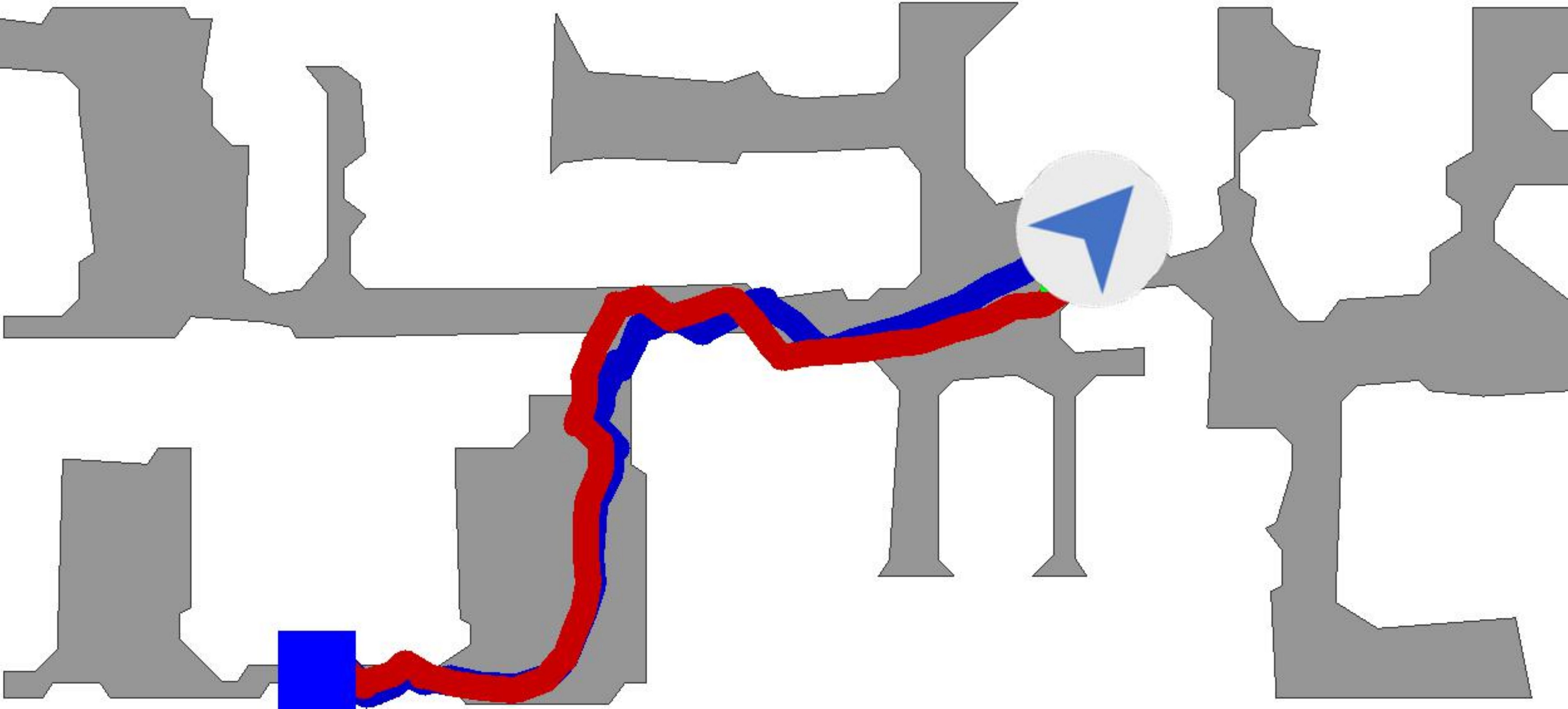}
    \captionsetup{width=0.9\linewidth}
    \caption{Cantwell, SPL $78\%$.}
    \label{supp fig: nav qualitative, Cantwell}
\end{subfigure}%
\hfill
\begin{subfigure}{0.3\textwidth}
\centering
    \includegraphics[width=\linewidth]{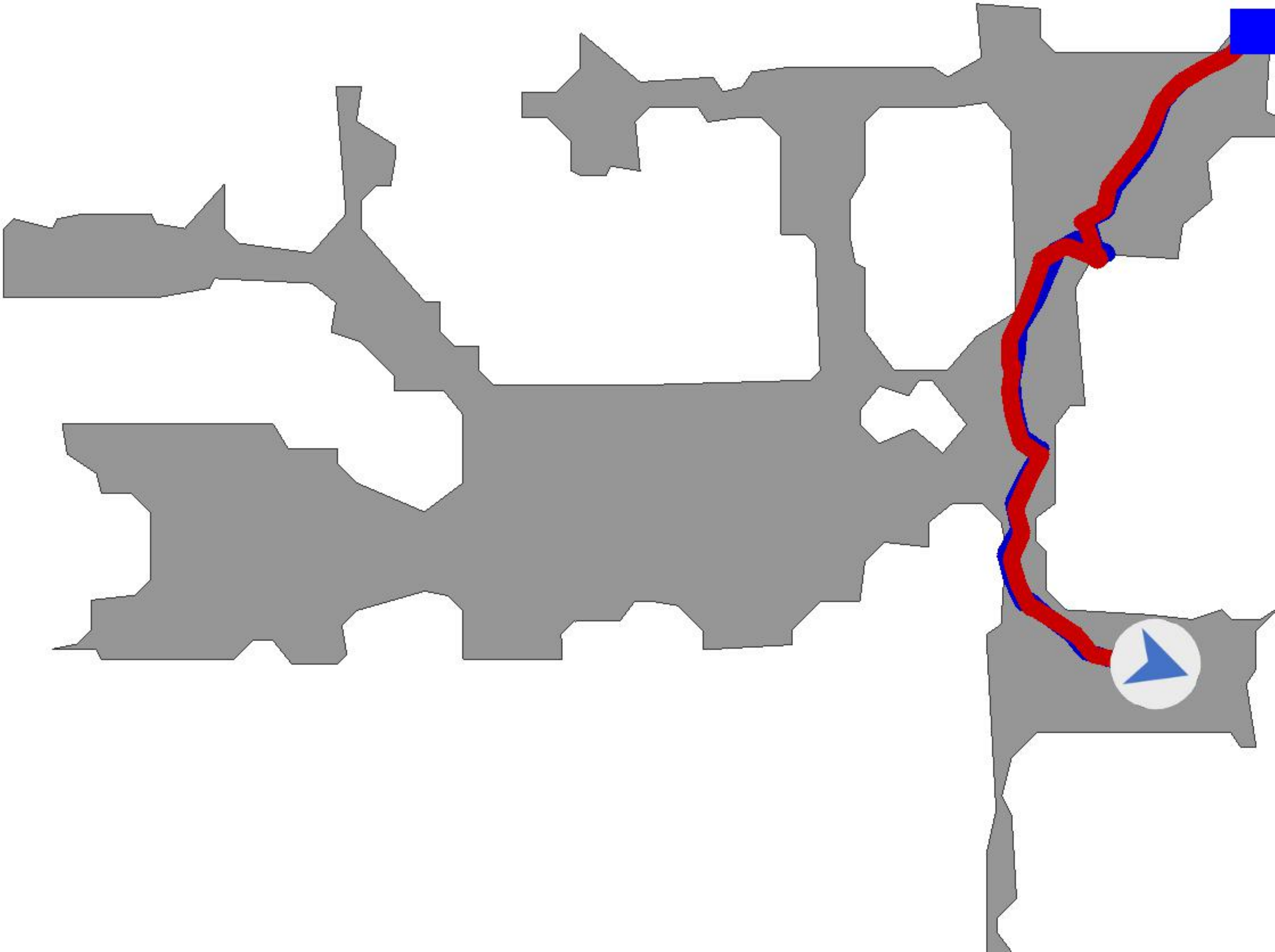}
    \captionsetup{width=0.9\linewidth}
    \caption{Eastville, SPL $78\%$.}
    \label{supp fig: nav qualitative, Eastville}
\end{subfigure}%
\hfill
\begin{subfigure}{0.35\textwidth}
\centering
    \includegraphics[width=\linewidth]{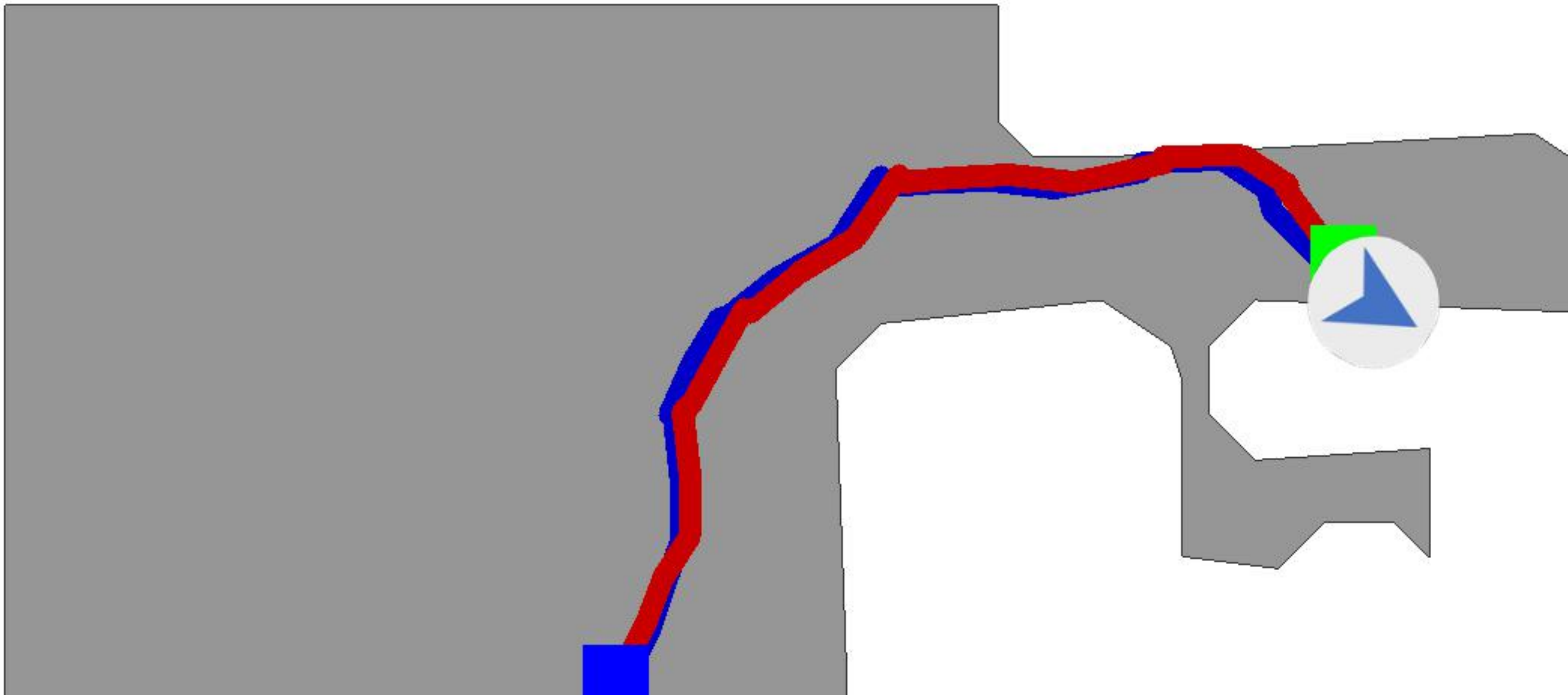}
    \captionsetup{width=0.9\linewidth}
    \caption{Edgemere, SPL $75\%$.}
    \label{supp fig: nav qualitative, Edgemere}
\end{subfigure}%
\hfill
\begin{subfigure}{0.23\textwidth}
\centering
    \includegraphics[width=\linewidth]{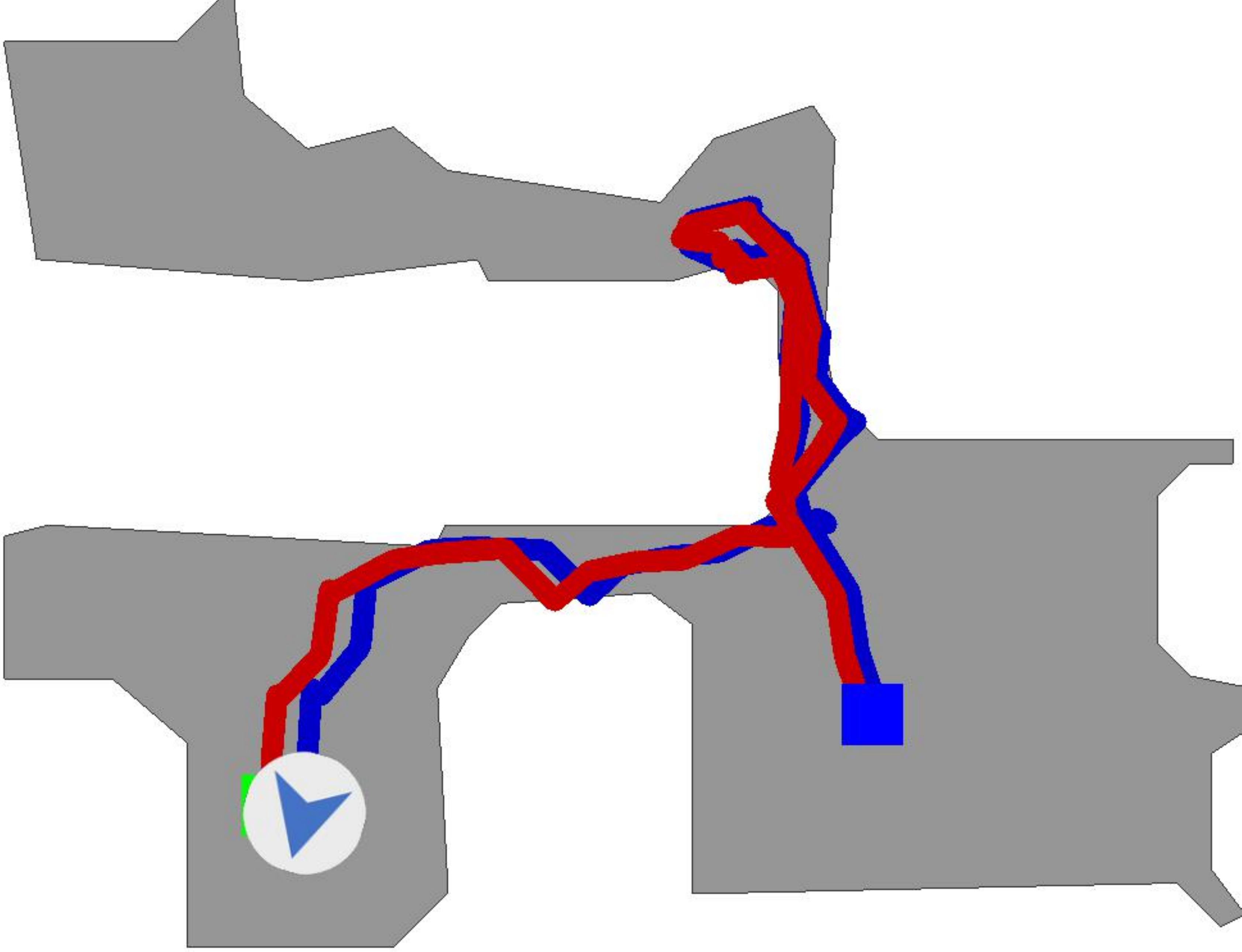}
    \captionsetup{width=0.9\linewidth}
    \caption{Eudora, SPL $35\%$.}
    \label{supp fig: nav qualitative, Eudora}
\end{subfigure}%
\hfill
\begin{subfigure}{0.3\textwidth}
\centering
    \includegraphics[width=\linewidth]{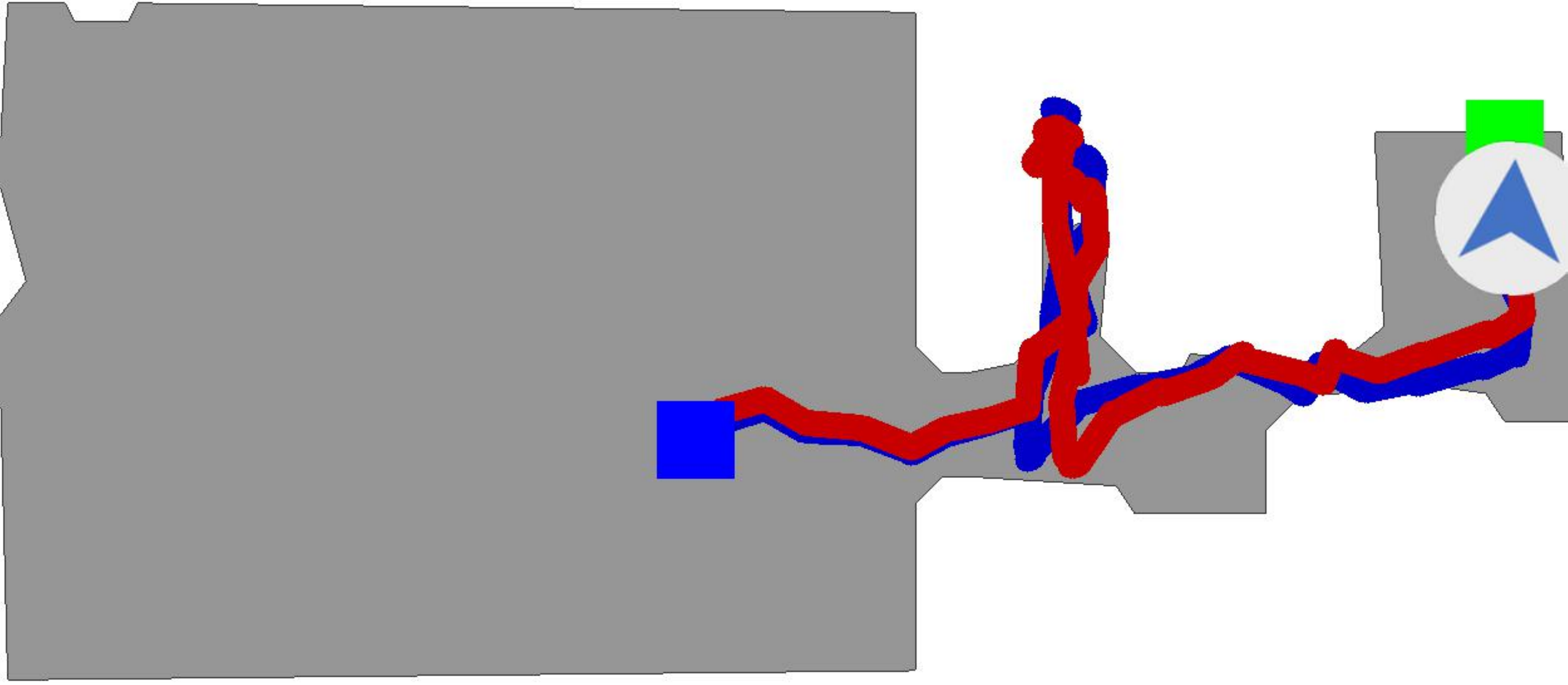}
    \captionsetup{width=0.9\linewidth}
    \caption{Sisters, fail.}
    \label{supp fig: nav qualitative, Sisters}
\end{subfigure}%
\vspace{0.1cm}
\caption{Qualitative results (best viewed in color). Agent is asked to navigate from {\color{blue}blue square} to {\color{ForestGreen}green square}. {\color{blue}Blue curve} is the actual path the agent takes while {\color{red}red curve} is based on the agent's estimate of its location from the VO model by integrating over $SE(2)$ estimation of each step.}
\label{supp fig: nav qualitative}
\end{figure*}

In \figref{supp fig: nav qualitative}, we provide additional qualitative results when integrating the navigation policy with our VO model.

\end{document}